\documentclass{bmcart}

\usepackage{amsthm,amsmath}
\usepackage[utf8]{inputenc} %unicode support

\usepackage{amssymb}
\usepackage{booktabs}
\usepackage{bbm}
\usepackage{xurl}
\usepackage{subfig}
\usepackage{pythonhighlight}
\usepackage{graphicx}
\usepackage{tikz}
\usepackage{adjustbox}

\startlocaldefs
\endlocaldefs

\begin{document}

%%% Start of article front matter
\begin{frontmatter}

\begin{fmbox}
\dochead{Research}


\title{Standardized Evaluation of Machine Learning Methods for Evolving Data Streams}

%%%%%%%%%%%%%%%%%%%%%%%%%%%%%%%%%%%%%%%%%%%%%%
%%                                          %%
%% Enter the authors here                   %%
%%                                          %%
%% Specify information, if available,       %%
%% in the form:                             %%
%%   <key>={<id1>,<id2>}                    %%
%%   <key>=                                 %%
%% Comment or delete the keys which are     %%
%% not used. Repeat \author command as much %%
%% as required.                             %%
%%                                          %%
%%%%%%%%%%%%%%%%%%%%%%%%%%%%%%%%%%%%%%%%%%%%%%

\author[
  addressref={aff1},                   % id's of addresses, e.g. {aff1,aff2}
  corref={aff1},                       % id of corresponding address, if any
% noteref={n1},                        % id's of article notes, if any
  email={johannes-christian.haug@uni-tuebingen.de}   % email address
]{\inits{J.H.}\fnm{Johannes} \snm{Haug}}
\author[
  addressref={aff1},
  email={efstratia.tramountani@student.uni-tuebingen.de}
]{\inits{E.T.}\fnm{Effi} \snm{Tramountani}}
\author[
  addressref={aff1},
  email={gjergji.kasneci@uni-tuebingen.de}
]{\inits{G.K.}\fnm{Gjergji} \snm{Kasneci}}

%%%%%%%%%%%%%%%%%%%%%%%%%%%%%%%%%%%%%%%%%%%%%%
%%                                          %%
%% Enter the authors' addresses here        %%
%%                                          %%
%% Repeat \address commands as much as      %%
%% required.                                %%
%%                                          %%
%%%%%%%%%%%%%%%%%%%%%%%%%%%%%%%%%%%%%%%%%%%%%%

\address[id=aff1]{%                           % unique id
  \orgdiv{Data Science and Analytics},             % department, if any
  \orgname{University of Tuebingen},          % university, etc
  \city{Tuebingen},                              % city
  \cny{Germany}                                    % country
}
%\address[id=aff2]{%
%  \orgdiv{Institute of Biology},
%  \orgname{National University of Sciences},
%  %\street{},
%  %\postcode{}
%  \city{Kiel},
%  \cny{Germany}
%}

%%%%%%%%%%%%%%%%%%%%%%%%%%%%%%%%%%%%%%%%%%%%%%
%%                                          %%
%% Enter short notes here                   %%
%%                                          %%
%% Short notes will be after addresses      %%
%% on first page.                           %%
%%                                          %%
%%%%%%%%%%%%%%%%%%%%%%%%%%%%%%%%%%%%%%%%%%%%%%

%%\note{Sample of title note}     % note to the article
%\note[id=n1]{Equal contributor} % note, connected to author

\end{fmbox}% comment this for two column layout

%%%%%%%%%%%%%%%%%%%%%%%%%%%%%%%%%%%%%%%%%%%%%%%
%%                                           %%
%% The Abstract begins here                  %%
%%                                           %%
%% Please refer to the Instructions for      %%
%% authors on https://www.biomedcentral.com/ %%
%% and include the section headings          %%
%% accordingly for your article type.        %%
%%                                           %%
%%%%%%%%%%%%%%%%%%%%%%%%%%%%%%%%%%%%%%%%%%%%%%%

\begin{abstractbox}

\begin{abstract} % abstract
Due to the unspecified and dynamic nature of data streams, online machine learning requires powerful and flexible solutions. However, evaluating online machine learning methods under realistic conditions is difficult. Existing work therefore often draws on different heuristics and simulations that do not necessarily produce meaningful and reliable results. Indeed, in the absence of common evaluation standards, it often remains unclear how online learning methods will perform in practice or in comparison to similar work. In this paper, we propose a comprehensive set of properties for high-quality machine learning in evolving data streams. In particular, we discuss sensible performance measures and evaluation strategies for online predictive modelling, online feature selection and concept drift detection. As one of the first works, we also look at the interpretability of online learning methods. The proposed evaluation standards are provided in a new Python framework called \textit{float}. Float is completely modular and allows the simultaneous integration of common libraries, such as scikit-multiflow or river, with custom code. Float is open-sourced and can be accessed at \url{https://github.com/haugjo/float}. In this sense, we hope that our work will contribute to more standardized, reliable and realistic testing and comparison of online machine learning methods.
\end{abstract}

%%%%%%%%%%%%%%%%%%%%%%%%%%%%%%%%%%%%%%%%%%%%%%
%%                                          %%
%% The keywords begin here                  %%
%%                                          %%
%% Put each keyword in separate \kwd{}.     %%
%%                                          %%
%%%%%%%%%%%%%%%%%%%%%%%%%%%%%%%%%%%%%%%%%%%%%%


% MSC classifications codes, if any
%\begin{keyword}[class=AMS]
%\kwd[Primary ]{}
%\kwd{}
%\kwd[; secondary ]{}
%\end{keyword}

\end{abstractbox}
%
%\end{fmbox}% uncomment this for two column layout

\end{frontmatter}

%%%%%%%%%%%%%%%%%%%%%%%%%%%%%%%%%%%%%%%%%%%%%%%%
%%                                            %%
%% The Main Body begins here                  %%
%%                                            %%
%% Please refer to the instructions for       %%
%% authors on:                                %%
%% https://www.biomedcentral.com/getpublished %%
%% and include the section headings           %%
%% accordingly for your article type.         %%
%%                                            %%
%% See the Results and Discussion section     %%
%% for details on how to create sub-sections  %%
%%                                            %%
%% use \cite{...} to cite references          %%
%%  \cite{koon} and                           %%
%%  \cite{oreg,khar,zvai,xjon,schn,pond}      %%
%%                                            %%
%%%%%%%%%%%%%%%%%%%%%%%%%%%%%%%%%%%%%%%%%%%%%%%%

%%%%%%%%%%%%%%%%%%%%%%%%% start of article main body
% <put your article body there>

\section{Introduction}
Data-driven or web-based applications like social media, e-commerce, and trading systems often generate and operate on large-scale evolving data streams. Unlike traditional (offline) batch machine learning, online learning methods must be able to process a potentially unlimited stream of observations and adjust to changes in the data generating process \cite{Lu2018,ditzler2015learning,gama2012survey}. Therefore, it can be crucial to gain a solid understanding of a model's strengths and weaknesses before it is deployed -- in particular in critical applications like online banking, autonomous driving or fraud detection. However, due to the unspecified behaviour of data streams, it is often unclear how online learning methods can be evaluated under realistic conditions.

% ###################################### Introductory Example #########################################
\begin{table}[t]
    \caption{\textbf{Inconsistency of Evaluations in Online Machine Learning.} Due to non-standardized evaluation strategies, test results reported in the data stream literature can diverge by a considerable margin. For example, below \textit{we show the accuracy for the standard Hoeffding Tree \cite{domingos2000mining} on four data sets, as reported in five different papers}. In the extreme case, the reported accuracy scores differ up to 13.51 percent points. Incidentally, the Covertype and Poker data sets are strongly imbalanced. Hence, accuracy is not an ideal evaluation measure to begin with. Such inconsistencies in experiment design and reported measures can be highly misleading -- especially for beginners in the field of online machine learning.}
    \label{tab:introductory_example}
    \centering
    \begin{adjustbox}{max width=\textwidth}
        \begin{tabular}{lllllll}
        \toprule
        \textbf{Dataset} & \textbf{\cite{holmes2005stress}} & \textbf{\cite{bifet2010fast}} & \textbf{\cite{read2012batch}} & \textbf{\cite{barddal2019learning}} & \textbf{\cite{barddal2020regularized}} & \textbf{Max. Diff.} \\
        \cmidrule(lr){1-1} \cmidrule(lr){2-6} \cmidrule(lr){7-7}
        Spambase & - & - & - & 80.35 & 85.47 & 5.12\\
        Poker & - & 73.62 & 76.07 & - & - & 2.45\\
        Electricity & - & 75.35 & 79.20 & - & 77.62 & 3.85\\
        Covertype & 66.83 & 68.30 & 80.31 & 80.34 & 73.71 & 13.51\\
        \bottomrule
        \end{tabular}
    \end{adjustbox}
\end{table}
% ###################################### Introductory Example #########################################

Unlike traditional batch-trained machine learning methods, online learning models can only access a fraction of the data at every given time step. Accordingly, common evaluation strategies that require the data to be available in main memory (e.g., cross-validation) are not applicable out of the box. Instead, we usually resort to sequential and simulation-based evaluation schemes. Although some evaluation strategies have emerged in the past \cite{Gama2009,Gama2012,Bifet2013}, we still lack a comprehensive and well-defined standard for the benchmarking of online learning methods \cite{Krawczyk2017}. Indeed, aside from online predictive modelling, there are hardly any evaluation standards for concept drift detection and online feature selection. 

As a consequence, there can be considerable differences in the evaluation methods used in existing work. For example, Table \ref{tab:introductory_example} depicts test results from five different research papers. Specifically, we compare the accuracy of a Very Fast Decision Tree \cite{Domingos2001} for four data sets as reported in the respective papers. Due to different evaluation strategies and unspecified behaviour, the results vary considerably. Without common evaluation standards and frameworks, empirical results are often only valid in a very restricted environment and are generally not comparable. 

In this work, we discuss good practices and standards for the evaluation of online learning methods. We summarize and extend popular evaluation strategies for data stream learning and introduce a comprehensive catalogue of requirements and performance measures. In particular, we propose important properties for online predictive models, online feature selection and concept drift detection. Although our focus is on online classification, most of the proposed properties also apply directly to regression or unsupervised tasks. Additionally, we briefly discuss the selection of adequate data sets, which remains an open issue.

To make the proposed evaluation standards more accessible, we introduce a new Python framework for \textit{Frictionless Online Analysis and Testing (float)}. Float is a lightweight and high-level framework that automates and standardizes inherent tasks of online evaluations. Float's modular architecture and evaluation pipeline simplify the integration of custom code with common online learning libraries such as scikit-multiflow \cite{montiel2018scikit} or river \cite{montiel2021river}. Moreover, float provides a variety of useful visualizations. The proposed framework is distributed under the MIT license via Github and the Python packaging index PyPI.

In summary, we contribute a comprehensive set of properties and performance measures that allow for a more standardized and realistic evaluation of online learning methods. The new Python framework, float, provides high-level access to the proposed standards and enables a quicker, more comparable and more reliable benchmarking in research and practice.

In Section \ref{sec:notation}, we introduce basic online learning concepts. In Section \ref{sec:related_work}, we briefly review previous studies on the evaluation of data stream learning. Afterwards, we discuss general evaluation strategies in Section \ref{sec:strategies} and propose a comprehensive set of properties for online predictive models, concept drift detection and online feature selection in Section \ref{sec:properties}. In this context, we also discuss the interpretability of models in the presence of incremental updates and concept drift. We provide relevant open-source resources for streaming data in Section \ref{sec:datasets}. Finally, we introduce the float framework and illustrate its use in Section \ref{sec:float}.

\section{Online Learning Preliminaries}\label{sec:notation}
A data stream can be represented by a (potentially infinite) series of time steps. At each time step $t$, the data stream produces observations $x_t \in \mathbb{R}^{n_t \times m_t}$ and corresponding labels $y_t \in \mathbb{R}^{n_t}$, where $n_t$ is the number of observations and $m_t$ is the number of features. The joint probability distribution $P_t(X, Y)$ denotes the active data concept at time step $t$, where $X$ and $Y$ are random variables corresponding to the observations and labels.

The active concept may evolve over time. Specifically, \textit{concept drift} describes a change in the joint probability distribution between two time steps, i.e., $P_{t_1}(X,Y) \neq P_{t_2}(X,Y)$. In general, we distinguish between real concept drift, i.e., $P_{t_1}(Y|X) \neq P_{t_2}(Y|X)$, and virtual concept drift, i.e., $P_{t_1}(X) \neq P_{t_2}(X)$. Unlike virtual concept drift, real concept drift affects the decision boundary. Note that there is a broader categorization of concept drift in the literature, e.g., based on its magnitude, length, or recurrences \cite{webb2016characterizing,Lu2018}.

In general, online machine learning deals with the same tasks as its offline counterpart. This includes supervised tasks like classification and regression or unsupervised tasks like clustering. 
%In this work, we focus on online classification, as it is the most prevalent use case in research. However, note that most of our considerations can also be directly applied to regression and clustering methods.

\section{Related Work}\label{sec:related_work}
Data streams are subject to external influences and temporal change. Therefore, it is often difficult to set up a testing environment that allows for meaningful evaluation of online learning methods. Various evaluation strategies and best practices have been developed in the past. While we discuss these strategies in more detail in Section \ref{sec:strategies} and \ref{sec:properties}, we provide a brief overview of existing work below.

The work of \cite{Domingos2001} was among the first to talk about important criteria for data stream mining. Much later, the authors in \cite{Gama2009} discussed issues that arise in the evaluation of online learning methods. This work was later extended \cite{Gama2012}. More recently, the work of \cite{Bifet2015} summarized the advantages and disadvantages of different evaluation strategies in the context of data stream classification. 

Although the criteria proposed in the above works are broadly applicable, additional challenges can arise in more specific contexts. For example, the papers of \cite{Goncalves2014} and \cite{Lu2018} discussed evaluation strategies, measures and data sets that can be used to evaluate concept drift detection and adaptation techniques. Likewise, the authors in \cite{RamirezGallego2017} proposed important criteria regarding the preprocessing of streaming data, including online feature selection. Besides, the works of \cite{Krawczyk2017} and \cite{Gomes2017} investigated the reliable evaluation of ensemble methods, which are a prominent group of high-performing online predictive models. Finally, the survey of \cite{Kolajo2019} provided a summary of popular data streaming tools and benchmarks in the context of big data mining.

Although various evaluation standards have been proposed over the years, most of them are either outdated, have a narrow focus, or remain superficial. In this paper, we provide a concise summary of the most popular evaluation practices in online machine learning. As one of the first works, we discuss and propose evaluation properties and measures for predictive modelling, feature selection and concept drift detection in non-stationary data streams. This work may thus guide beginners and experts alike in the conception of more standardized benchmarks and experiments.

\section{Evaluation Strategies}\label{sec:strategies}
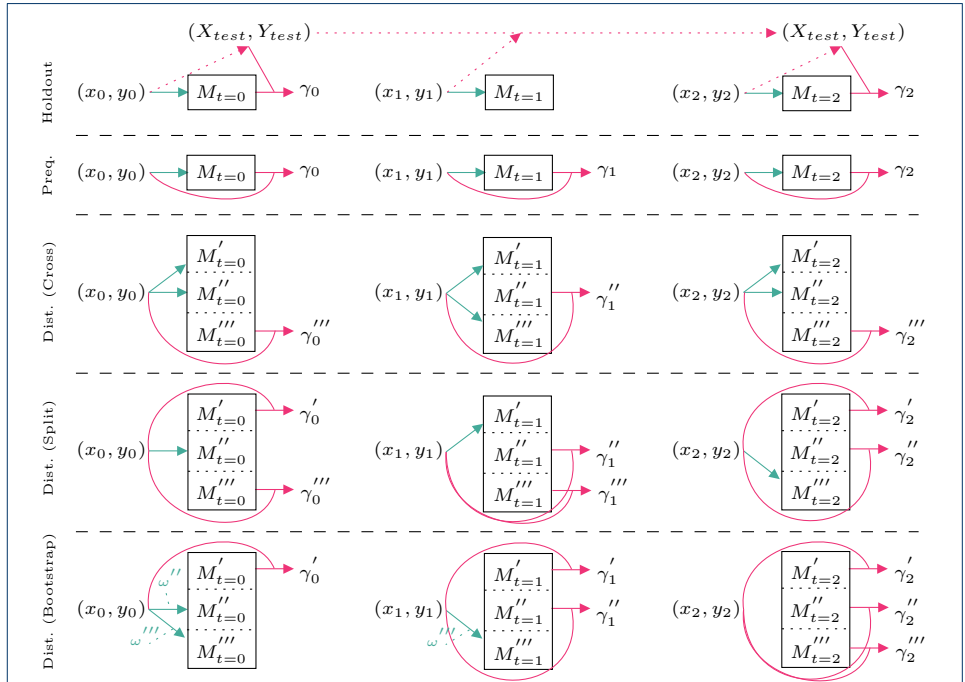
\begin{figure}
\centering
\begin{tikzpicture}[x=0.75pt,y=0.75pt,yscale=-1,xscale=1]

%Straight Lines [id:da05441254622743097] 
\draw [color={rgb, 255:red, 68; green, 170; blue, 153 }  ,draw opacity=1 ]   (127.23,133.87) -- (143.9,133.87) ;
\draw [shift={(146.9,133.87)}, rotate = 180] [fill={rgb, 255:red, 68; green, 170; blue, 153 }  ,fill opacity=1 ][line width=0.08]  [draw opacity=0] (5.36,-2.57) -- (0,0) -- (5.36,2.57) -- cycle    ;
%Straight Lines [id:da16123628948470703] 
\draw [color={rgb, 255:red, 238; green, 51; blue, 119 }  ,draw opacity=1 ]   (180.3,133.87) -- (196.97,133.87) ;
\draw [shift={(199.97,133.87)}, rotate = 180] [fill={rgb, 255:red, 238; green, 51; blue, 119 }  ,fill opacity=1 ][line width=0.08]  [draw opacity=0] (5.36,-2.57) -- (0,0) -- (5.36,2.57) -- cycle    ;
%Curve Lines [id:da8360540669027154] 
\draw [color={rgb, 255:red, 238; green, 51; blue, 119 }  ,draw opacity=1 ]   (127.23,133.87) .. controls (136.5,146.67) and (186,156.17) .. (190,133.67) ;
%Straight Lines [id:da9662375999423225] 
\draw [color={rgb, 255:red, 68; green, 170; blue, 153 }  ,draw opacity=1 ]   (277.23,133.87) -- (293.9,133.87) ;
\draw [shift={(296.9,133.87)}, rotate = 180] [fill={rgb, 255:red, 68; green, 170; blue, 153 }  ,fill opacity=1 ][line width=0.08]  [draw opacity=0] (5.36,-2.57) -- (0,0) -- (5.36,2.57) -- cycle    ;
%Straight Lines [id:da10031998960195465] 
\draw [color={rgb, 255:red, 238; green, 51; blue, 119 }  ,draw opacity=1 ]   (330.3,133.87) -- (346.97,133.87) ;
\draw [shift={(349.97,133.87)}, rotate = 180] [fill={rgb, 255:red, 238; green, 51; blue, 119 }  ,fill opacity=1 ][line width=0.08]  [draw opacity=0] (5.36,-2.57) -- (0,0) -- (5.36,2.57) -- cycle    ;
%Straight Lines [id:da19650023496562974] 
\draw [color={rgb, 255:red, 68; green, 170; blue, 153 }  ,draw opacity=1 ]   (427.23,133.87) -- (443.9,133.87) ;
\draw [shift={(446.9,133.87)}, rotate = 180] [fill={rgb, 255:red, 68; green, 170; blue, 153 }  ,fill opacity=1 ][line width=0.08]  [draw opacity=0] (5.36,-2.57) -- (0,0) -- (5.36,2.57) -- cycle    ;
%Straight Lines [id:da2290358802624486] 
\draw [color={rgb, 255:red, 238; green, 51; blue, 119 }  ,draw opacity=1 ]   (480.3,133.87) -- (496.97,133.87) ;
\draw [shift={(499.97,133.87)}, rotate = 180] [fill={rgb, 255:red, 238; green, 51; blue, 119 }  ,fill opacity=1 ][line width=0.08]  [draw opacity=0] (5.36,-2.57) -- (0,0) -- (5.36,2.57) -- cycle    ;
%Straight Lines [id:da010012354148875247] 
\draw [color={rgb, 255:red, 68; green, 170; blue, 153 }  ,draw opacity=1 ]   (127.23,93.37) -- (143.9,93.37) ;
\draw [shift={(146.9,93.37)}, rotate = 180] [fill={rgb, 255:red, 68; green, 170; blue, 153 }  ,fill opacity=1 ][line width=0.08]  [draw opacity=0] (5.36,-2.57) -- (0,0) -- (5.36,2.57) -- cycle    ;
%Straight Lines [id:da9828934377591694] 
\draw [color={rgb, 255:red, 238; green, 51; blue, 119 }  ,draw opacity=1 ]   (180.3,93.37) -- (196.97,93.37) ;
\draw [shift={(199.97,93.37)}, rotate = 180] [fill={rgb, 255:red, 238; green, 51; blue, 119 }  ,fill opacity=1 ][line width=0.08]  [draw opacity=0] (5.36,-2.57) -- (0,0) -- (5.36,2.57) -- cycle    ;
%Straight Lines [id:da8617705764339885] 
\draw [color={rgb, 255:red, 68; green, 170; blue, 153 }  ,draw opacity=1 ]   (277.23,93.37) -- (293.9,93.37) ;
\draw [shift={(296.9,93.37)}, rotate = 180] [fill={rgb, 255:red, 68; green, 170; blue, 153 }  ,fill opacity=1 ][line width=0.08]  [draw opacity=0] (5.36,-2.57) -- (0,0) -- (5.36,2.57) -- cycle    ;
%Straight Lines [id:da1235067023353047] 
\draw [color={rgb, 255:red, 238; green, 51; blue, 119 }  ,draw opacity=1 ] [dash pattern={on 0.84pt off 2.51pt}]  (127.23,93.37) -- (174.28,71.44) ;
\draw [shift={(177,70.17)}, rotate = 155.01] [fill={rgb, 255:red, 238; green, 51; blue, 119 }  ,fill opacity=1 ][line width=0.08]  [draw opacity=0] (5.36,-2.57) -- (0,0) -- (5.36,2.57) -- cycle    ;
%Straight Lines [id:da7673467412843704] 
\draw [color={rgb, 255:red, 238; green, 51; blue, 119 }  ,draw opacity=1 ]   (177,70.17) -- (190,93.17) ;
%Straight Lines [id:da8875425759587823] 
\draw [color={rgb, 255:red, 68; green, 170; blue, 153 }  ,draw opacity=1 ]   (427.23,93.87) -- (443.9,93.87) ;
\draw [shift={(446.9,93.87)}, rotate = 180] [fill={rgb, 255:red, 68; green, 170; blue, 153 }  ,fill opacity=1 ][line width=0.08]  [draw opacity=0] (5.36,-2.57) -- (0,0) -- (5.36,2.57) -- cycle    ;
%Straight Lines [id:da5826777470632967] 
\draw [color={rgb, 255:red, 238; green, 51; blue, 119 }  ,draw opacity=1 ]   (480.3,93.87) -- (496.97,93.87) ;
\draw [shift={(499.97,93.87)}, rotate = 180] [fill={rgb, 255:red, 238; green, 51; blue, 119 }  ,fill opacity=1 ][line width=0.08]  [draw opacity=0] (5.36,-2.57) -- (0,0) -- (5.36,2.57) -- cycle    ;
%Straight Lines [id:da13592425958715926] 
\draw [color={rgb, 255:red, 238; green, 51; blue, 119 }  ,draw opacity=1 ] [dash pattern={on 0.84pt off 2.51pt}]  (427.23,93.87) -- (473.3,71.48) ;
\draw [shift={(476,70.17)}, rotate = 154.08] [fill={rgb, 255:red, 238; green, 51; blue, 119 }  ,fill opacity=1 ][line width=0.08]  [draw opacity=0] (5.36,-2.57) -- (0,0) -- (5.36,2.57) -- cycle    ;
%Straight Lines [id:da011720383950776014] 
\draw [color={rgb, 255:red, 238; green, 51; blue, 119 }  ,draw opacity=1 ]   (476,70.17) -- (490.5,93.67) ;
%Straight Lines [id:da6184167695436757] 
\draw [color={rgb, 255:red, 238; green, 51; blue, 119 }  ,draw opacity=1 ] [dash pattern={on 0.84pt off 2.51pt}]  (211.5,63.42) -- (440.5,63.42) ;
\draw [shift={(443.5,63.42)}, rotate = 180] [fill={rgb, 255:red, 238; green, 51; blue, 119 }  ,fill opacity=1 ][line width=0.08]  [draw opacity=0] (5.36,-2.57) -- (0,0) -- (5.36,2.57) -- cycle    ;
%Straight Lines [id:da7496580712035084] 
\draw [color={rgb, 255:red, 238; green, 51; blue, 119 }  ,draw opacity=1 ] [dash pattern={on 0.84pt off 2.51pt}]  (277.23,93.37) -- (312.16,65.3) ;
\draw [shift={(314.5,63.42)}, rotate = 141.21] [fill={rgb, 255:red, 238; green, 51; blue, 119 }  ,fill opacity=1 ][line width=0.08]  [draw opacity=0] (5.36,-2.57) -- (0,0) -- (5.36,2.57) -- cycle    ;
%Straight Lines [id:da8767074024568511] 
\draw [color={rgb, 255:red, 68; green, 170; blue, 153 }  ,draw opacity=1 ]   (126.73,194.37) -- (143.4,194.37) ;
\draw [shift={(146.4,194.37)}, rotate = 180] [fill={rgb, 255:red, 68; green, 170; blue, 153 }  ,fill opacity=1 ][line width=0.08]  [draw opacity=0] (5.36,-2.57) -- (0,0) -- (5.36,2.57) -- cycle    ;
%Straight Lines [id:da8302641635131853] 
\draw [color={rgb, 255:red, 68; green, 170; blue, 153 }  ,draw opacity=1 ]   (126.73,194.37) -- (143.11,181.98) ;
\draw [shift={(145.5,180.17)}, rotate = 142.89] [fill={rgb, 255:red, 68; green, 170; blue, 153 }  ,fill opacity=1 ][line width=0.08]  [draw opacity=0] (5.36,-2.57) -- (0,0) -- (5.36,2.57) -- cycle    ;
%Curve Lines [id:da7281845444143209] 
\draw [color={rgb, 255:red, 238; green, 51; blue, 119 }  ,draw opacity=1 ]   (126.73,194.37) .. controls (123,232.67) and (184.5,242.67) .. (190.5,213.67) ;
%Straight Lines [id:da02511518298014659] 
\draw [color={rgb, 255:red, 238; green, 51; blue, 119 }  ,draw opacity=1 ]   (180.3,213.87) -- (196.97,213.87) ;
\draw [shift={(199.97,213.87)}, rotate = 180] [fill={rgb, 255:red, 238; green, 51; blue, 119 }  ,fill opacity=1 ][line width=0.08]  [draw opacity=0] (5.36,-2.57) -- (0,0) -- (5.36,2.57) -- cycle    ;
%Straight Lines [id:da4021103491585998] 
\draw [color={rgb, 255:red, 68; green, 170; blue, 153 }  ,draw opacity=1 ]   (276.73,194.87) -- (292.64,207.32) ;
\draw [shift={(295,209.17)}, rotate = 218.06] [fill={rgb, 255:red, 68; green, 170; blue, 153 }  ,fill opacity=1 ][line width=0.08]  [draw opacity=0] (5.36,-2.57) -- (0,0) -- (5.36,2.57) -- cycle    ;
%Straight Lines [id:da05792216383778759] 
\draw [color={rgb, 255:red, 238; green, 51; blue, 119 }  ,draw opacity=1 ]   (329.8,194.37) -- (346.47,194.37) ;
\draw [shift={(349.47,194.37)}, rotate = 180] [fill={rgb, 255:red, 238; green, 51; blue, 119 }  ,fill opacity=1 ][line width=0.08]  [draw opacity=0] (5.36,-2.57) -- (0,0) -- (5.36,2.57) -- cycle    ;
%Straight Lines [id:da44281349076348686] 
\draw [color={rgb, 255:red, 68; green, 170; blue, 153 }  ,draw opacity=1 ]   (426.73,194.37) -- (443.4,194.37) ;
\draw [shift={(446.4,194.37)}, rotate = 180] [fill={rgb, 255:red, 68; green, 170; blue, 153 }  ,fill opacity=1 ][line width=0.08]  [draw opacity=0] (5.36,-2.57) -- (0,0) -- (5.36,2.57) -- cycle    ;
%Straight Lines [id:da7929249480578928] 
\draw [color={rgb, 255:red, 68; green, 170; blue, 153 }  ,draw opacity=1 ]   (126.73,274.37) -- (143.4,274.37) ;
\draw [shift={(146.4,274.37)}, rotate = 180] [fill={rgb, 255:red, 68; green, 170; blue, 153 }  ,fill opacity=1 ][line width=0.08]  [draw opacity=0] (5.36,-2.57) -- (0,0) -- (5.36,2.57) -- cycle    ;
%Straight Lines [id:da8300430326065109] 
\draw [color={rgb, 255:red, 238; green, 51; blue, 119 }  ,draw opacity=1 ]   (180.3,293.87) -- (196.97,293.87) ;
\draw [shift={(199.97,293.87)}, rotate = 180] [fill={rgb, 255:red, 238; green, 51; blue, 119 }  ,fill opacity=1 ][line width=0.08]  [draw opacity=0] (5.36,-2.57) -- (0,0) -- (5.36,2.57) -- cycle    ;
%Straight Lines [id:da3803847187223992] 
\draw [color={rgb, 255:red, 238; green, 51; blue, 119 }  ,draw opacity=1 ]   (329.8,273.87) -- (346.47,273.87) ;
\draw [shift={(349.47,273.87)}, rotate = 180] [fill={rgb, 255:red, 238; green, 51; blue, 119 }  ,fill opacity=1 ][line width=0.08]  [draw opacity=0] (5.36,-2.57) -- (0,0) -- (5.36,2.57) -- cycle    ;
%Curve Lines [id:da27206117651689077] 
\draw [color={rgb, 255:red, 238; green, 51; blue, 119 }  ,draw opacity=1 ]   (126.73,274.37) .. controls (122,239.67) and (179.5,229.17) .. (190,253.67) ;
%Straight Lines [id:da7897518868053235] 
\draw [color={rgb, 255:red, 238; green, 51; blue, 119 }  ,draw opacity=1 ]   (180.3,253.87) -- (196.97,253.87) ;
\draw [shift={(199.97,253.87)}, rotate = 180] [fill={rgb, 255:red, 238; green, 51; blue, 119 }  ,fill opacity=1 ][line width=0.08]  [draw opacity=0] (5.36,-2.57) -- (0,0) -- (5.36,2.57) -- cycle    ;
%Straight Lines [id:da2350877022307376] 
\draw [color={rgb, 255:red, 238; green, 51; blue, 119 }  ,draw opacity=1 ]   (330.3,294.37) -- (346.97,294.37) ;
\draw [shift={(349.97,294.37)}, rotate = 180] [fill={rgb, 255:red, 238; green, 51; blue, 119 }  ,fill opacity=1 ][line width=0.08]  [draw opacity=0] (5.36,-2.57) -- (0,0) -- (5.36,2.57) -- cycle    ;
%Straight Lines [id:da11813493119575758] 
\draw [color={rgb, 255:red, 238; green, 51; blue, 119 }  ,draw opacity=1 ]   (479.8,253.87) -- (496.47,253.87) ;
\draw [shift={(499.47,253.87)}, rotate = 180] [fill={rgb, 255:red, 238; green, 51; blue, 119 }  ,fill opacity=1 ][line width=0.08]  [draw opacity=0] (5.36,-2.57) -- (0,0) -- (5.36,2.57) -- cycle    ;
%Straight Lines [id:da7317305779748473] 
\draw [color={rgb, 255:red, 238; green, 51; blue, 119 }  ,draw opacity=1 ]   (480.3,273.37) -- (496.97,273.37) ;
\draw [shift={(499.97,273.37)}, rotate = 180] [fill={rgb, 255:red, 238; green, 51; blue, 119 }  ,fill opacity=1 ][line width=0.08]  [draw opacity=0] (5.36,-2.57) -- (0,0) -- (5.36,2.57) -- cycle    ;
%Shape: Rectangle [id:dp2815317844547103] 
\draw   (146,125.17) -- (180.3,125.17) -- (180.3,142.67) -- (146,142.67) -- cycle ;
%Shape: Rectangle [id:dp6212475116305147] 
\draw   (296,125.17) -- (330.3,125.17) -- (330.3,142.67) -- (296,142.67) -- cycle ;
%Shape: Rectangle [id:dp2015588807003712] 
\draw   (446.5,125.17) -- (480.8,125.17) -- (480.8,142.67) -- (446.5,142.67) -- cycle ;
%Shape: Rectangle [id:dp44764845691287625] 
\draw   (146.5,84.67) -- (180.8,84.67) -- (180.8,102.17) -- (146.5,102.17) -- cycle ;
%Shape: Rectangle [id:dp7603288399959627] 
\draw   (296.5,84.67) -- (330.8,84.67) -- (330.8,102.17) -- (296.5,102.17) -- cycle ;
%Shape: Rectangle [id:dp05980952931935812] 
\draw   (446.5,85.17) -- (480.8,85.17) -- (480.8,102.67) -- (446.5,102.67) -- cycle ;
%Straight Lines [id:da8056330484483887] 
\draw  [dash pattern={on 4.5pt off 4.5pt}]  (90,115.17) -- (515,115.17) ;
%Straight Lines [id:da22862062887886503] 
\draw  [dash pattern={on 4.5pt off 4.5pt}]  (90,155.17) -- (515,155.17) ;
%Straight Lines [id:da3553970286037631] 
\draw  [dash pattern={on 4.5pt off 4.5pt}]  (90,235.17) -- (515,235.17) ;
%Straight Lines [id:da4431369823917657] 
\draw  [dash pattern={on 4.5pt off 4.5pt}]  (90,315.17) -- (515,315.17) ;
%Straight Lines [id:da7595777350129311] 
\draw [color={rgb, 255:red, 68; green, 170; blue, 153 }  ,draw opacity=1 ]   (126.73,354.37) -- (143.4,354.37) ;
\draw [shift={(146.4,354.37)}, rotate = 180] [fill={rgb, 255:red, 68; green, 170; blue, 153 }  ,fill opacity=1 ][line width=0.08]  [draw opacity=0] (5.36,-2.57) -- (0,0) -- (5.36,2.57) -- cycle    ;
%Straight Lines [id:da9686764759498705] 
\draw [color={rgb, 255:red, 238; green, 51; blue, 119 }  ,draw opacity=1 ]   (329.8,353.87) -- (346.47,353.87) ;
\draw [shift={(349.47,353.87)}, rotate = 180] [fill={rgb, 255:red, 238; green, 51; blue, 119 }  ,fill opacity=1 ][line width=0.08]  [draw opacity=0] (5.36,-2.57) -- (0,0) -- (5.36,2.57) -- cycle    ;
%Straight Lines [id:da9034678126076277] 
\draw [color={rgb, 255:red, 238; green, 51; blue, 119 }  ,draw opacity=1 ]   (180.3,333.87) -- (196.97,333.87) ;
\draw [shift={(199.97,333.87)}, rotate = 180] [fill={rgb, 255:red, 238; green, 51; blue, 119 }  ,fill opacity=1 ][line width=0.08]  [draw opacity=0] (5.36,-2.57) -- (0,0) -- (5.36,2.57) -- cycle    ;
%Straight Lines [id:da5780010144949692] 
\draw [color={rgb, 255:red, 238; green, 51; blue, 119 }  ,draw opacity=1 ]   (479.8,333.87) -- (496.47,333.87) ;
\draw [shift={(499.47,333.87)}, rotate = 180] [fill={rgb, 255:red, 238; green, 51; blue, 119 }  ,fill opacity=1 ][line width=0.08]  [draw opacity=0] (5.36,-2.57) -- (0,0) -- (5.36,2.57) -- cycle    ;
%Straight Lines [id:da2852992447363758] 
\draw [color={rgb, 255:red, 238; green, 51; blue, 119 }  ,draw opacity=1 ]   (480.3,353.37) -- (496.97,353.37) ;
\draw [shift={(499.97,353.37)}, rotate = 180] [fill={rgb, 255:red, 238; green, 51; blue, 119 }  ,fill opacity=1 ][line width=0.08]  [draw opacity=0] (5.36,-2.57) -- (0,0) -- (5.36,2.57) -- cycle    ;
%Straight Lines [id:da6601946274415871] 
\draw [color={rgb, 255:red, 68; green, 170; blue, 153 }  ,draw opacity=1 ] [dash pattern={on 0.84pt off 2.51pt}]  (135.5,346.17) -- (137.5,354.67) ;
%Straight Lines [id:da915567866709855] 
\draw [color={rgb, 255:red, 68; green, 170; blue, 153 }  ,draw opacity=1 ] [dash pattern={on 0.84pt off 2.51pt}]  (139,364.17) -- (127,371.67) ;
%Shape: Rectangle [id:dp15589344650697523] 
\draw   (146,165.67) -- (180.3,165.67) -- (180.3,224.17) -- (146,224.17) -- cycle ;
%Straight Lines [id:da14375686318471348] 
\draw  [dash pattern={on 0.84pt off 2.51pt}]  (146,184.17) -- (180.5,184.17) ;
%Straight Lines [id:da9216396158846123] 
\draw  [dash pattern={on 0.84pt off 2.51pt}]  (146,204.67) -- (180.5,204.67) ;
%Curve Lines [id:da13728500050148162] 
\draw [color={rgb, 255:red, 238; green, 51; blue, 119 }  ,draw opacity=1 ]   (277.23,133.87) .. controls (286.5,146.67) and (336,156.17) .. (340,133.67) ;
%Curve Lines [id:da14163152263043255] 
\draw [color={rgb, 255:red, 238; green, 51; blue, 119 }  ,draw opacity=1 ]   (427.23,133.87) .. controls (436.5,146.67) and (486,156.17) .. (490,133.67) ;
%Shape: Rectangle [id:dp11858003864274425] 
\draw   (295.5,166.67) -- (329.8,166.67) -- (329.8,225.17) -- (295.5,225.17) -- cycle ;
%Straight Lines [id:da6158239085128172] 
\draw  [dash pattern={on 0.84pt off 2.51pt}]  (295.5,185.17) -- (330,185.17) ;
%Straight Lines [id:da06383664331628336] 
\draw  [dash pattern={on 0.84pt off 2.51pt}]  (295.5,205.67) -- (330,205.67) ;
%Straight Lines [id:da5363527008269005] 
\draw [color={rgb, 255:red, 68; green, 170; blue, 153 }  ,draw opacity=1 ]   (276.73,194.87) -- (293.11,182.48) ;
\draw [shift={(295.5,180.67)}, rotate = 142.89] [fill={rgb, 255:red, 68; green, 170; blue, 153 }  ,fill opacity=1 ][line width=0.08]  [draw opacity=0] (5.36,-2.57) -- (0,0) -- (5.36,2.57) -- cycle    ;
%Shape: Rectangle [id:dp2185743282644863] 
\draw   (446.5,165.67) -- (480.8,165.67) -- (480.8,224.17) -- (446.5,224.17) -- cycle ;
%Straight Lines [id:da018408685142940806] 
\draw  [dash pattern={on 0.84pt off 2.51pt}]  (446.5,184.17) -- (481,184.17) ;
%Straight Lines [id:da654744217578541] 
\draw  [dash pattern={on 0.84pt off 2.51pt}]  (446.5,204.67) -- (481,204.67) ;
%Straight Lines [id:da8858973093444968] 
\draw [color={rgb, 255:red, 68; green, 170; blue, 153 }  ,draw opacity=1 ]   (426.73,194.37) -- (443.11,181.98) ;
\draw [shift={(445.5,180.17)}, rotate = 142.89] [fill={rgb, 255:red, 68; green, 170; blue, 153 }  ,fill opacity=1 ][line width=0.08]  [draw opacity=0] (5.36,-2.57) -- (0,0) -- (5.36,2.57) -- cycle    ;
%Curve Lines [id:da01458901756313491] 
\draw [color={rgb, 255:red, 238; green, 51; blue, 119 }  ,draw opacity=1 ]   (427.23,194.37) .. controls (423.5,232.67) and (485,242.67) .. (491,213.67) ;
%Straight Lines [id:da9638767081220299] 
\draw [color={rgb, 255:red, 238; green, 51; blue, 119 }  ,draw opacity=1 ]   (480.8,213.87) -- (497.47,213.87) ;
\draw [shift={(500.47,213.87)}, rotate = 180] [fill={rgb, 255:red, 238; green, 51; blue, 119 }  ,fill opacity=1 ][line width=0.08]  [draw opacity=0] (5.36,-2.57) -- (0,0) -- (5.36,2.57) -- cycle    ;
%Shape: Rectangle [id:dp3526578009243768] 
\draw   (146.5,245.67) -- (180.8,245.67) -- (180.8,304.17) -- (146.5,304.17) -- cycle ;
%Straight Lines [id:da28775944659771246] 
\draw  [dash pattern={on 0.84pt off 2.51pt}]  (146.5,264.17) -- (181,264.17) ;
%Straight Lines [id:da7992213812915319] 
\draw  [dash pattern={on 0.84pt off 2.51pt}]  (146.5,284.67) -- (181,284.67) ;
%Curve Lines [id:da6690107647446977] 
\draw [color={rgb, 255:red, 238; green, 51; blue, 119 }  ,draw opacity=1 ]   (126.73,274.37) .. controls (123,312.67) and (184.5,322.67) .. (190.5,293.67) ;
%Shape: Rectangle [id:dp37582619811091655] 
\draw   (295.5,246.67) -- (329.8,246.67) -- (329.8,305.17) -- (295.5,305.17) -- cycle ;
%Straight Lines [id:da9211601659875399] 
\draw  [dash pattern={on 0.84pt off 2.51pt}]  (295.5,265.17) -- (330,265.17) ;
%Straight Lines [id:da34356783147683356] 
\draw  [dash pattern={on 0.84pt off 2.51pt}]  (295.5,285.67) -- (330,285.67) ;
%Curve Lines [id:da792992187101383] 
\draw [color={rgb, 255:red, 238; green, 51; blue, 119 }  ,draw opacity=1 ]   (276.73,274.87) .. controls (273,313.17) and (334.5,323.17) .. (340.5,294.17) ;
%Curve Lines [id:da8697830976265508] 
\draw [color={rgb, 255:red, 238; green, 51; blue, 119 }  ,draw opacity=1 ]   (276.73,274.87) .. controls (277,324.17) and (349,319.67) .. (340,274.17) ;
%Straight Lines [id:da07413341703837051] 
\draw [color={rgb, 255:red, 68; green, 170; blue, 153 }  ,draw opacity=1 ]   (276.73,274.87) -- (293.11,262.48) ;
\draw [shift={(295.5,260.67)}, rotate = 142.89] [fill={rgb, 255:red, 68; green, 170; blue, 153 }  ,fill opacity=1 ][line width=0.08]  [draw opacity=0] (5.36,-2.57) -- (0,0) -- (5.36,2.57) -- cycle    ;
%Shape: Rectangle [id:dp7294514032694115] 
\draw   (446,245.67) -- (480.3,245.67) -- (480.3,304.17) -- (446,304.17) -- cycle ;
%Straight Lines [id:da031246457517178028] 
\draw  [dash pattern={on 0.84pt off 2.51pt}]  (446,264.17) -- (480.5,264.17) ;
%Straight Lines [id:da6626306869455039] 
\draw  [dash pattern={on 0.84pt off 2.51pt}]  (446,284.67) -- (480.5,284.67) ;
%Curve Lines [id:da30309927639899525] 
\draw [color={rgb, 255:red, 238; green, 51; blue, 119 }  ,draw opacity=1 ]   (426.73,274.37) .. controls (422,239.67) and (479.5,229.17) .. (490,253.67) ;
%Straight Lines [id:da7366681999912015] 
\draw [color={rgb, 255:red, 68; green, 170; blue, 153 }  ,draw opacity=1 ]   (426.73,274.37) -- (442.64,286.82) ;
\draw [shift={(445,288.67)}, rotate = 218.06] [fill={rgb, 255:red, 68; green, 170; blue, 153 }  ,fill opacity=1 ][line width=0.08]  [draw opacity=0] (5.36,-2.57) -- (0,0) -- (5.36,2.57) -- cycle    ;
%Shape: Rectangle [id:dp9549670683670552] 
\draw   (146.5,325.67) -- (180.8,325.67) -- (180.8,384.17) -- (146.5,384.17) -- cycle ;
%Straight Lines [id:da20161054162742742] 
\draw  [dash pattern={on 0.84pt off 2.51pt}]  (146.5,344.17) -- (181,344.17) ;
%Straight Lines [id:da9856163892657601] 
\draw  [dash pattern={on 0.84pt off 2.51pt}]  (146,364.67) -- (180.5,364.67) ;
%Straight Lines [id:da9309553384109088] 
\draw [color={rgb, 255:red, 68; green, 170; blue, 153 }  ,draw opacity=1 ]   (126.73,354.37) -- (142.64,366.82) ;
\draw [shift={(145,368.67)}, rotate = 218.06] [fill={rgb, 255:red, 68; green, 170; blue, 153 }  ,fill opacity=1 ][line width=0.08]  [draw opacity=0] (5.36,-2.57) -- (0,0) -- (5.36,2.57) -- cycle    ;
%Curve Lines [id:da4751253857211828] 
\draw [color={rgb, 255:red, 238; green, 51; blue, 119 }  ,draw opacity=1 ]   (126.73,354.37) .. controls (122,319.67) and (179.5,309.17) .. (190,333.67) ;
%Shape: Rectangle [id:dp0661534832509687] 
\draw   (296,326.17) -- (330.3,326.17) -- (330.3,384.67) -- (296,384.67) -- cycle ;
%Straight Lines [id:da6082569601371319] 
\draw  [dash pattern={on 0.84pt off 2.51pt}]  (296,344.67) -- (330.5,344.67) ;
%Straight Lines [id:da571418654671892] 
\draw  [dash pattern={on 0.84pt off 2.51pt}]  (296,365.17) -- (330.5,365.17) ;
%Straight Lines [id:da19646973708154447] 
\draw [color={rgb, 255:red, 68; green, 170; blue, 153 }  ,draw opacity=1 ]   (276.73,354.87) -- (292.64,367.32) ;
\draw [shift={(295,369.17)}, rotate = 218.06] [fill={rgb, 255:red, 68; green, 170; blue, 153 }  ,fill opacity=1 ][line width=0.08]  [draw opacity=0] (5.36,-2.57) -- (0,0) -- (5.36,2.57) -- cycle    ;
%Straight Lines [id:da9321093560190346] 
\draw [color={rgb, 255:red, 238; green, 51; blue, 119 }  ,draw opacity=1 ]   (329.8,334.37) -- (346.47,334.37) ;
\draw [shift={(349.47,334.37)}, rotate = 180] [fill={rgb, 255:red, 238; green, 51; blue, 119 }  ,fill opacity=1 ][line width=0.08]  [draw opacity=0] (5.36,-2.57) -- (0,0) -- (5.36,2.57) -- cycle    ;
%Curve Lines [id:da8369059740803126] 
\draw [color={rgb, 255:red, 238; green, 51; blue, 119 }  ,draw opacity=1 ]   (276.73,354.87) .. controls (272,320.17) and (329.5,309.67) .. (340,334.17) ;
%Straight Lines [id:da7072831909654225] 
\draw [color={rgb, 255:red, 68; green, 170; blue, 153 }  ,draw opacity=1 ] [dash pattern={on 0.84pt off 2.51pt}]  (288.5,364.67) -- (276.5,372.17) ;
%Shape: Rectangle [id:dp07075138743714171] 
\draw   (446,325.17) -- (480.3,325.17) -- (480.3,383.67) -- (446,383.67) -- cycle ;
%Straight Lines [id:da05490849874809389] 
\draw  [dash pattern={on 0.84pt off 2.51pt}]  (446,343.67) -- (480.5,343.67) ;
%Straight Lines [id:da599869790260811] 
\draw  [dash pattern={on 0.84pt off 2.51pt}]  (446,364.17) -- (480.5,364.17) ;
%Curve Lines [id:da9913482581800517] 
\draw [color={rgb, 255:red, 238; green, 51; blue, 119 }  ,draw opacity=1 ]   (426.73,354.37) .. controls (422,319.67) and (479.5,309.17) .. (490,333.67) ;
%Curve Lines [id:da8039781161398114] 
\draw [color={rgb, 255:red, 238; green, 51; blue, 119 }  ,draw opacity=1 ]   (426.73,354.37) .. controls (427,403.67) and (499,399.17) .. (490,353.67) ;
%Curve Lines [id:da9135506018921373] 
\draw [color={rgb, 255:red, 238; green, 51; blue, 119 }  ,draw opacity=1 ]   (426.73,354.37) .. controls (423,392.67) and (484.5,402.67) .. (490.5,373.67) ;
%Straight Lines [id:da18177182915773415] 
\draw [color={rgb, 255:red, 238; green, 51; blue, 119 }  ,draw opacity=1 ]   (480.3,373.87) -- (496.97,373.87) ;
\draw [shift={(499.97,373.87)}, rotate = 180] [fill={rgb, 255:red, 238; green, 51; blue, 119 }  ,fill opacity=1 ][line width=0.08]  [draw opacity=0] (5.36,-2.57) -- (0,0) -- (5.36,2.57) -- cycle    ;
%Curve Lines [id:da8347298817778355] 
\draw [color={rgb, 255:red, 238; green, 51; blue, 119 }  ,draw opacity=1 ]   (276.73,354.87) .. controls (277,404.17) and (349,399.67) .. (340,354.17) ;
%Curve Lines [id:da8671953812339777] 
\draw [color={rgb, 255:red, 238; green, 51; blue, 119 }  ,draw opacity=1 ]   (426.73,274.37) .. controls (427,323.67) and (499,319.17) .. (490,273.67) ;
%Curve Lines [id:da05440266182199127] 
\draw [color={rgb, 255:red, 238; green, 51; blue, 119 }  ,draw opacity=1 ]   (276.73,194.87) .. controls (277,244.17) and (349,239.67) .. (340,194.17) ;

% Text Node
\draw (149.17,129.13) node [anchor=north west][inner sep=0.75pt]  [font=\scriptsize]  {$M_{t=0}$};
% Text Node
\draw (89.03,127.87) node [anchor=north west][inner sep=0.75pt]  [font=\scriptsize]  {$( x_{0} ,y_{0})$};
% Text Node
\draw (201.57,129.2) node [anchor=north west][inner sep=0.75pt]  [font=\scriptsize]  {$\gamma _{0}$};
% Text Node
\draw (72.03,147.03) node [anchor=north west][inner sep=0.75pt]  [font=\tiny,rotate=-270] [align=left] {Preq.};
% Text Node
\draw (299.17,129.13) node [anchor=north west][inner sep=0.75pt]  [font=\scriptsize]  {$M_{t=1}$};
% Text Node
\draw (239.03,127.87) node [anchor=north west][inner sep=0.75pt]  [font=\scriptsize]  {$( x_{1} ,y_{1})$};
% Text Node
\draw (351.57,129.2) node [anchor=north west][inner sep=0.75pt]  [font=\scriptsize]  {$\gamma _{1}$};
% Text Node
\draw (449.17,129.13) node [anchor=north west][inner sep=0.75pt]  [font=\scriptsize]  {$M_{t=2}$};
% Text Node
\draw (389.03,127.87) node [anchor=north west][inner sep=0.75pt]  [font=\scriptsize]  {$( x_{2} ,y_{2})$};
% Text Node
\draw (501.57,129.2) node [anchor=north west][inner sep=0.75pt]  [font=\scriptsize]  {$\gamma _{2}$};
% Text Node
\draw (149.17,88.63) node [anchor=north west][inner sep=0.75pt]  [font=\scriptsize]  {$M_{t=0}$};
% Text Node
\draw (89.03,87.37) node [anchor=north west][inner sep=0.75pt]  [font=\scriptsize]  {$( x_{0} ,y_{0})$};
% Text Node
\draw (201.57,88.7) node [anchor=north west][inner sep=0.75pt]  [font=\scriptsize]  {$\gamma _{0}$};
% Text Node
\draw (72.03,110.53) node [anchor=north west][inner sep=0.75pt]  [font=\tiny,rotate=-270] [align=left] {Holdout};
% Text Node
\draw (299.17,88.63) node [anchor=north west][inner sep=0.75pt]  [font=\scriptsize]  {$M_{t=1}$};
% Text Node
\draw (239.03,87.37) node [anchor=north west][inner sep=0.75pt]  [font=\scriptsize]  {$( x_{1} ,y_{1})$};
% Text Node
\draw (145.03,57.37) node [anchor=north west][inner sep=0.75pt]  [font=\scriptsize]  {$( X_{test} ,Y_{test})$};
% Text Node
\draw (449.17,89.13) node [anchor=north west][inner sep=0.75pt]  [font=\scriptsize]  {$M_{t=2}$};
% Text Node
\draw (389.03,87.87) node [anchor=north west][inner sep=0.75pt]  [font=\scriptsize]  {$( x_{2} ,y_{2})$};
% Text Node
\draw (501.57,89.2) node [anchor=north west][inner sep=0.75pt]  [font=\scriptsize]  {$\gamma _{2}$};
% Text Node
\draw (444.53,57.37) node [anchor=north west][inner sep=0.75pt]  [font=\scriptsize]  {$( X_{test} ,Y_{test})$};
% Text Node
\draw (149.17,167.13) node [anchor=north west][inner sep=0.75pt]  [font=\scriptsize]  {$M_{t=0}^{'}$};
% Text Node
\draw (89.03,188.37) node [anchor=north west][inner sep=0.75pt]  [font=\scriptsize]  {$( x_{0} ,y_{0})$};
% Text Node
\draw (202.07,206.7) node [anchor=north west][inner sep=0.75pt]  [font=\scriptsize]  {$\gamma _{0}^{'''}$};
% Text Node
\draw (72.03,221.03) node [anchor=north west][inner sep=0.75pt]  [font=\tiny,rotate=-270] [align=left] {Dist. (Cross)};
% Text Node
\draw (149.17,186.63) node [anchor=north west][inner sep=0.75pt]  [font=\scriptsize]  {$M_{t=0}^{''}$};
% Text Node
\draw (149.17,207.13) node [anchor=north west][inner sep=0.75pt]  [font=\scriptsize]  {$M_{t=0}^{'''}$};
% Text Node
\draw (72.03,298.53) node [anchor=north west][inner sep=0.75pt]  [font=\tiny,rotate=-270] [align=left] {Dist. (Split)};
% Text Node
\draw (299.17,168.13) node [anchor=north west][inner sep=0.75pt]  [font=\scriptsize]  {$M_{t=1}^{'}$};
% Text Node
\draw (239.03,188.37) node [anchor=north west][inner sep=0.75pt]  [font=\scriptsize]  {$( x_{1} ,y_{1})$};
% Text Node
\draw (299.17,187.63) node [anchor=north west][inner sep=0.75pt]  [font=\scriptsize]  {$M_{t=1}^{''}$};
% Text Node
\draw (299.17,207.63) node [anchor=north west][inner sep=0.75pt]  [font=\scriptsize]  {$M_{t=1}^{'''}$};
% Text Node
\draw (351.07,188.7) node [anchor=north west][inner sep=0.75pt]  [font=\scriptsize]  {$\gamma _{1}^{''}$};
% Text Node
\draw (449.17,167.13) node [anchor=north west][inner sep=0.75pt]  [font=\scriptsize]  {$M_{t=2}^{'}$};
% Text Node
\draw (389.03,188.37) node [anchor=north west][inner sep=0.75pt]  [font=\scriptsize]  {$( x_{2} ,y_{2})$};
% Text Node
\draw (502.07,206.7) node [anchor=north west][inner sep=0.75pt]  [font=\scriptsize]  {$\gamma _{2}^{'''}$};
% Text Node
\draw (448.67,186.63) node [anchor=north west][inner sep=0.75pt]  [font=\scriptsize]  {$M_{t=2}^{''}$};
% Text Node
\draw (449.17,207.13) node [anchor=north west][inner sep=0.75pt]  [font=\scriptsize]  {$M_{t=2}^{'''}$};
% Text Node
\draw (149.17,247.13) node [anchor=north west][inner sep=0.75pt]  [font=\scriptsize]  {$M_{t=0}^{'}$};
% Text Node
\draw (89.03,268.37) node [anchor=north west][inner sep=0.75pt]  [font=\scriptsize]  {$( x_{0} ,y_{0})$};
% Text Node
\draw (202.07,286.7) node [anchor=north west][inner sep=0.75pt]  [font=\scriptsize]  {$\gamma _{0}^{'''}$};
% Text Node
\draw (149.17,266.63) node [anchor=north west][inner sep=0.75pt]  [font=\scriptsize]  {$M_{t=0}^{''}$};
% Text Node
\draw (149.17,287.13) node [anchor=north west][inner sep=0.75pt]  [font=\scriptsize]  {$M_{t=0}^{'''}$};
% Text Node
\draw (299.17,248.13) node [anchor=north west][inner sep=0.75pt]  [font=\scriptsize]  {$M_{t=1}^{'}$};
% Text Node
\draw (239.03,268.37) node [anchor=north west][inner sep=0.75pt]  [font=\scriptsize]  {$( x_{1} ,y_{1})$};
% Text Node
\draw (299.17,267.63) node [anchor=north west][inner sep=0.75pt]  [font=\scriptsize]  {$M_{t=1}^{''}$};
% Text Node
\draw (299.17,288.13) node [anchor=north west][inner sep=0.75pt]  [font=\scriptsize]  {$M_{t=1}^{'''}$};
% Text Node
\draw (351.07,268.2) node [anchor=north west][inner sep=0.75pt]  [font=\scriptsize]  {$\gamma _{1}^{''}$};
% Text Node
\draw (449.17,247.13) node [anchor=north west][inner sep=0.75pt]  [font=\scriptsize]  {$M_{t=2}^{'}$};
% Text Node
\draw (389.03,268.37) node [anchor=north west][inner sep=0.75pt]  [font=\scriptsize]  {$( x_{2} ,y_{2})$};
% Text Node
\draw (448.67,266.63) node [anchor=north west][inner sep=0.75pt]  [font=\scriptsize]  {$M_{t=2}^{''}$};
% Text Node
\draw (449.17,287.13) node [anchor=north west][inner sep=0.75pt]  [font=\scriptsize]  {$M_{t=2}^{'''}$};
% Text Node
\draw (201.57,246.7) node [anchor=north west][inner sep=0.75pt]  [font=\scriptsize]  {$\gamma _{0}^{'}$};
% Text Node
\draw (352.07,287.2) node [anchor=north west][inner sep=0.75pt]  [font=\scriptsize]  {$\gamma _{1}^{'''}$};
% Text Node
\draw (501.57,246.7) node [anchor=north west][inner sep=0.75pt]  [font=\scriptsize]  {$\gamma _{2}^{'}$};
% Text Node
\draw (501.57,267.7) node [anchor=north west][inner sep=0.75pt]  [font=\scriptsize]  {$\gamma _{2}^{''}$};
% Text Node
\draw (72.03,386.03) node [anchor=north west][inner sep=0.75pt]  [font=\tiny,rotate=-270] [align=left] {Dist. (Bootstrap)};
% Text Node
\draw (149.17,327.13) node [anchor=north west][inner sep=0.75pt]  [font=\scriptsize]  {$M_{t=0}^{'}$};
% Text Node
\draw (89.03,348.37) node [anchor=north west][inner sep=0.75pt]  [font=\scriptsize]  {$( x_{0} ,y_{0})$};
% Text Node
\draw (149.17,346.63) node [anchor=north west][inner sep=0.75pt]  [font=\scriptsize]  {$M_{t=0}^{''}$};
% Text Node
\draw (149.17,367.13) node [anchor=north west][inner sep=0.75pt]  [font=\scriptsize]  {$M_{t=0}^{'''}$};
% Text Node
\draw (299.17,328.13) node [anchor=north west][inner sep=0.75pt]  [font=\scriptsize]  {$M_{t=1}^{'}$};
% Text Node
\draw (239.03,348.37) node [anchor=north west][inner sep=0.75pt]  [font=\scriptsize]  {$( x_{1} ,y_{1})$};
% Text Node
\draw (299.17,347.63) node [anchor=north west][inner sep=0.75pt]  [font=\scriptsize]  {$M_{t=1}^{''}$};
% Text Node
\draw (299.17,368.13) node [anchor=north west][inner sep=0.75pt]  [font=\scriptsize]  {$M_{t=1}^{'''}$};
% Text Node
\draw (351.07,348.2) node [anchor=north west][inner sep=0.75pt]  [font=\scriptsize]  {$\gamma _{1}^{''}$};
% Text Node
\draw (449.17,327.13) node [anchor=north west][inner sep=0.75pt]  [font=\scriptsize]  {$M_{t=2}^{'}$};
% Text Node
\draw (389.03,348.37) node [anchor=north west][inner sep=0.75pt]  [font=\scriptsize]  {$( x_{2} ,y_{2})$};
% Text Node
\draw (448.67,346.63) node [anchor=north west][inner sep=0.75pt]  [font=\scriptsize]  {$M_{t=2}^{''}$};
% Text Node
\draw (449.17,367.13) node [anchor=north west][inner sep=0.75pt]  [font=\scriptsize]  {$M_{t=2}^{'''}$};
% Text Node
\draw (201.57,326.7) node [anchor=north west][inner sep=0.75pt]  [font=\scriptsize]  {$\gamma _{0}^{'}$};
% Text Node
\draw (501.57,326.7) node [anchor=north west][inner sep=0.75pt]  [font=\scriptsize]  {$\gamma _{2}^{'}$};
% Text Node
\draw (501.57,347.7) node [anchor=north west][inner sep=0.75pt]  [font=\scriptsize]  {$\gamma _{2}^{''}$};
% Text Node
\draw (130,332.32) node [anchor=north west][inner sep=0.75pt]  [font=\tiny,color={rgb, 255:red, 68; green, 170; blue, 153 }  ,opacity=1 ]  {$\omega ^{''}$};
% Text Node
\draw (116,361.82) node [anchor=north west][inner sep=0.75pt]  [font=\tiny,color={rgb, 255:red, 68; green, 170; blue, 153 }  ,opacity=1 ]  {$\omega ^{'''}$};
% Text Node
\draw (351.57,327.2) node [anchor=north west][inner sep=0.75pt]  [font=\scriptsize]  {$\gamma _{1}^{'}$};
% Text Node
\draw (265.5,362.32) node [anchor=north west][inner sep=0.75pt]  [font=\tiny,color={rgb, 255:red, 68; green, 170; blue, 153 }  ,opacity=1 ]  {$\omega ^{'''}$};
% Text Node
\draw (502.07,366.7) node [anchor=north west][inner sep=0.75pt]  [font=\scriptsize]  {$\gamma _{2}^{'''}$};

\end{tikzpicture}
\caption{\textbf{Evaluation Strategies for Online Machine Learning.} Since online predictive models are trained incrementally, traditional (e.g., sampling-based) evaluation strategies are not applicable. Instead, we can use periodic holdout, prequential, or distributed $k$-fold evaluation, which we describe in Section \ref{sec:strategies}. For each of the strategies, at each time step we first test the model (pink), which yields a performance measure $\gamma \in \mathbb{R}$, and then update the model (green). In a periodic holdout evaluation, we maintain a test set $(X_{test},Y_{test})$ that may be updated over time (indicated by dashed arrows). In a distributed $k$-fold evaluation, we train $k$ instances of the model in parallel (indicated by quotes). The instances to be trained and tested are randomly selected according to one of three schemes (cross-/split-/bootstrap-validation). In the bootstrap validation scheme, $\omega$ specifies a training weight drawn from the Poisson distribution.
}
\label{fig:strategies}
\end{figure}

To train and evaluate a machine learning model, we require data. Traditionally, we have access to a training data set during development. This data set is split in a meaningful way (e.g., once in a holdout validation or $k$ times in a $k$-fold cross-validation). We then train the model on the training set and evaluate it on the test set. In data streams, however, we do not have access to a complete data set at any time. In addition, the data generating distribution can change. As a result, online learning models need to be continuously updated over time. Therefore, it would not be sensible to evaluate an online learning model at a specific point in time, i.e., at a single training stage using a static train/test split. Selecting an adequate time step would also be difficult in practice, since a data stream is potentially infinite. Rather, we need to evaluate online learning models periodically. On this basis, several evaluation strategies have been proposed, which we illustrate in Figure \ref{fig:strategies} and discuss below.

%----------------- TIKZ Evaluation Strategies -----------%
%\input{tikz_strategies}
%----------------- TIKZ Evaluation Strategies -----------%

\subsection{Periodical Holdout Evaluation}
In a periodical holdout evaluation, we test the online predictive model at frequent intervals. The holdout set comprises test observations that represent the active concept. Since the active concept evolves, the holdout set should be updated over time. For example, we may replace old instances in the holdout set as we progress. Between the periodic evaluations, we continue updating the model, but do not test its performance. The test frequency is usually controlled via a hyperparameter. Depending on the test frequency, a holdout evaluation might miss particularly short-term data concepts. Moreover, holdout evaluations yield only snapshots of the system at particular time steps. That is, there is no guarantee that the holdout evaluation at a time step $t$ is also representative for $t-1$ and $t+1$. Therefore, we believe that a holdout evaluation should generally only be used when both the data distribution and the predictive model are known to be stable.

\subsection{Prequential Evaluation}
Unlike a periodical holdout evaluation, a prequential strategy (also known as test-then-train) evaluates the predictive model at each time step. Specifically, new observations are first used to test and then update the model. This is the most common evaluation strategy in practice. A prequential evaluation gives a more pessimistic performance estimate than the holdout evaluation \cite{Gama2009}. In particular, the prequential evaluation also takes into account all early performance measurements, which are often poor due to the initial training of the model. 

To mitigate this effect, one may introduce a forgetting mechanism via sliding windows or fading factors \cite{Gama2012}. In the sliding window approach, we aggregate the measurements obtained in a time window. Any measurement that falls outside the specified window no longer contributes to the performance estimate. In a fading factor approach (i.e., exponentially weighted averaging), the influence of old measurements decreases over time according to the specified fading (or decay) factor. Unlike a sliding window, a fading factor approach is memory-less, meaning that old measurements do not need to be saved.

\subsection{Distributed k-Fold Evaluation}
Cross-validation is one of the most popular evaluation techniques in offline learning environments because it provides more robust results than a traditional holdout evaluation. However, cross-validation and other sampling techniques are generally not applicable in a data stream, since we cannot access the entire data set and the data distribution is subject to change \cite{Gama2009}. Therefore, the periodical holdout and prequential strategies described above are commonly used even though they do not provide statistical significance. Indeed, hypothesis testing is rather uncommon in the data stream literature.

As an alternative, the authors in \cite{Bifet2015} proposed a $k$-fold distributed evaluation strategy that allows us to perform hypothesis testing for online learning models. To this end, we need to run $k$ instances of the same predictive model in parallel. Each streaming observation is distributed to one or multiple model instances according to one of three validation schemes. In the \textit{cross-validation} scheme, at every time step, we randomly pick one classifier instance for testing and use the remaining ones for training. Conversely, the \textit{split-validation} scheme trains one randomly selected model instance per time step and tests all others. Finally, in the \textit{bootstrap-validation} scheme, we sample weights from a Poisson distribution for each classifier instance. The weights indicate whether a model instance is used for testing (zero weight) or training (weight greater than zero). Note that the Poisson weight also controls the influence of an observation during training. For example, with a sample weight of 3, we would use an observation three times to update the corresponding model instance. In this way, the bootstrap validation scheme simulates random sampling with replacement \cite{Bifet2015}.

The experiments of \cite{Bifet2015} showed that the $k$-fold distributed cross-validation strategy gives reliable results for different hypothesis tests. However, the distributed $k$-fold evaluation is usually much more costly than a periodical holdout or prequential evaluation.

% ###################################### Table of Properties #########################################
\begin{table*}[t]
    \caption{\textbf{Evaluation Properties for Data Stream Methods.} Below we summarize all properties and corresponding performance measures introduced in Section \ref{sec:properties} for the evaluation of machine learning methods in evolving data streams. In general, we recommend to use a prequential evaluation strategy or, if statistical significance is required, a more costly $k$-fold distributed cross validation \cite{Bifet2015} (Section \ref{sec:strategies}). With the novel Python framework \textit{float} (Section \ref{sec:float}), we can run standardized experiments and evaluate these properties in a few lines of code.}
    \label{tab:properties}
    \centering
    \begin{adjustbox}{max width=\columnwidth}
        \begin{tabular}{ll}
        \toprule
        \textbf{Property} & \textbf{Meaningful Evaluation Measures} \\
        \cmidrule(lr){1-1} \cmidrule(lr){2-2}
        Predictive Performance & Generalization error; \\
        & F1 measure or $\kappa$-statistic \cite{Bifet2015} (for classification)\\
        Computational Efficiency & Computation time (for training and testing);\\ 
        & RAM-Hours \cite{bifet2010fast}\\
        Algorithmic Stability & Noise Variability (Eq. \eqref{eq:noise_variability})\\
        Concept Drift Adaptability & Drift Performance Deterioration (Eq. \eqref{eq:drift_performance_det}); \\
        & Drift Restoration Time (Eq. \eqref{eq:drift_restoration}) \\
        Interpretability & Complexity over time, e.g. no. of parameters/splits \cite{haug2022dynamic}\\
        \cmidrule(lr){1-1}\cmidrule(lr){2-2}
        \textit{Concept Drift Detection} & \\
        \cmidrule(lr){1-1}\cmidrule(lr){2-2}
        Detection Truthfulness & Detected Change Rate (Eq. \eqref{eq:det_change_rate}); \\
        & False Discovery Rate (Eq. \eqref{eq:fals_disc_rate});\\
        & Mean Time Between False Alarms \cite{Bifet2013}\\
        Detection Timeliness & Delay (no. of time steps); Mean Time Ratio (Eq. \eqref{eq:mean_time_ratio}) \cite{Bifet2013} \\
        \cmidrule(lr){1-1}\cmidrule(lr){2-2}
        \textit{Online Feature Selection} & \\
        \cmidrule(lr){1-1}\cmidrule(lr){2-2}
        Feature Set Stability & Adjusted Nogueira stability (Eq. \eqref{eq:stability}) \cite{nogueira2017stability,Haug2020}\\
        Feature Selectivity & Reduction rate (in \% of the original feature dimensionality)\\
        \bottomrule
        \end{tabular}
    \end{adjustbox}
\end{table*}
% ###################################### Table of Properties #########################################

\section{Properties for Online Machine Learning Methods}\label{sec:properties}
As described above, evolving data streams bring several challenges. Next, we translate these challenges into a set of fundamental properties and requirements for online machine learning. In addition, we discuss meaningful performance measures to evaluate each property. We begin with general properties and then address specific aspects of concept drift detection and online feature selection. Unless stated otherwise, we do not impose any restrictions on the functional form of the online predictive model. That is, we assume that the predictive model can be queried but otherwise remains a black-box. Consequently, the proposed properties are directly applicable to any online learning framework. In combination, these properties enable online learning methods that can be effectively used under real-world streaming conditions.

\subsection{General Properties}
In the following, we specify general properties for high-quality online learning methods. The importance of each property may depend on the application at hand. We list all properties along with recommended performance measures in Table \ref{tab:properties}.

\subsubsection{Predictive Performance and Generalization}
High predictive performance, i.e., low generalization error, is a central goal of any predictive machine learning method. In general, this property describes the ability of a model to correctly predict the target of previously unobserved test data. As discussed in Section \ref{sec:strategies}, we need to choose between different strategies to periodically evaluate an online predictive model. In this context, we can apply the same or slightly adjusted performance measures as for offline learning. For example, commonly used measures for online classification are the 0-1 loss and the accuracy (which only gives valid results for balanced target distributions). For more robust results on imbalanced data, one may use a combination of precision and recall (e.g., F1) \cite{Lu2018}, variations of the $\kappa$-statistic \cite{Bifet2015} or an adaptation of the Area Under the Curve of the Receiver Operating characteristic (AUC) \cite{Brzezinski2017}.

\subsubsection{Computational Efficiency}
To model data streams in theory, we usually apply a logical notion of time, i.e., we treat time as a sequence of discrete steps (see Section \ref{sec:notation}). However, in practice, streaming observations can arrive in very quick succession. For example, large scale web-based applications such as credit card transactions may produce many new observations per second. In order to avoid a backlog and update the model in real-time, incoming observations should be processed at the rate they arrive \cite{Gama2009}. Likewise, online applications often have limited access to hardware capacities, e.g., in a distributed or embedded system. Therefore, it can be crucial that online learning models are efficient and use few resources.

Measuring the computation time is relatively simple with standard packages (e.g., the \textit{time}-package in Python). The computation times for model training and testing should be monitored independently \cite{RamirezGallego2017}. To quantify the memory usage, the authors in \cite{bifet2010fast} proposed to use RAM-hours. One RAM-hour corresponds to one Gigabyte of RAM used for one hour. However, during development, it can be difficult to accurately monitor the RAM usage of a particular model or process. Moreover, existing software packages for monitoring memory usage are often inaccurate or extremely slow. Indeed, both the estimated computation time and memory usage depend heavily on the implementation and the given hardware specifications. Hence, efficiency estimates are generally only comparable if they were obtained under the same conditions. A theoretical analysis of model complexity often provides a more reliable indication of computational efficiency.

\subsubsection{Algorithmic Stability}
Real-world data is subject to noise, e.g., due to faulty network connections or (human) errors in data collection. If a model is susceptible to noise, its performance usually suffers. Unlike batch learning methods, online learning models must be able to distinguish noise and outliers from concept drift. This can be difficult because both noise and concept drift manifest as previously unobserved behaviour. Surprisingly, the stability of online learning methods has received little attention in the past.

Stability is often estimated by calculating the variability of the model output for perturbed inputs. That is, we can manipulate an input observation with artificial noise (e.g., sampled from some probability distribution) and monitor the change in a performance measure. To the best of our knowledge, there is no common estimate of stability for data stream methods. Therefore, we define the \textit{noise variability} at time step $t$:
\begin{equation}\label{eq:noise_variability}
    NV_t = \frac{1}{N} \sum^N_{n=1} L\big(y_t, f_t(x_t + z_n)\big) - L\big(y_t, f_t(x_t)\big),
\end{equation}
where $N$ is the number of times we sample noise $z_n \sim \mathcal{Z}$ (e.g., $\mathcal{Z} = \mathcal{N}(0,1)$), $L$ is a performance measure function and $f_t$ is the model at time step $t$. If the noise variability is small for most time steps, i.e., a large number of perturbed observations, we can assume that the online learning model is stable. Nevertheless, it can be difficult to find a meaningful noise distribution $\mathcal{Z}$, so this measure should be used with caution. In general, more attention should be paid to evaluating the stability of machine learning models in data streams. 

\subsubsection{Concept Drift Adaptability}\label{sec:adaptability_drift}
In general, we can deal with concept drift either by passive model adaptation or by active drift detection. In the former approach, the model is adjusted over time, e.g., through continuous updates or the use of sliding windows. Conversely, in the active approach, we aim to identify the exact time of concept drift using a dedicated drift detection method \cite{Haug2021}. Each time we actively detect a concept drift, we re-train the model (or parts of it). Accordingly, active drift detection can, but need not, be part of the drift adaptation process of the online learning model. Next, we discuss the general evaluation of model performance in the presence of concept drift. Note that active drift detection methods should be evaluated in terms of additional properties, which we present in Section \ref{sec:drift_properties}.

Online learning models should be robust enough to suffer only minor performance deterioration after concept drift, and flexible enough to quickly recover previous performance. To the best of our knowledge, we currently lack sensible definitions of corresponding measures. Accordingly, given a concept drift at time step $t_d$, we define \textit{drift performance deterioration} as:
\begin{equation}\label{eq:drift_performance_det}
    DPD_{t_d} = \frac{1}{W} \left(\sum^{W-1}_{w=0} L\big(y_{t_d+w}, \hat{y}_{t_d+w}\big) - \sum^W_{w=1} L\big(y_{t_d-w}, \hat{y}_{t_d-w}\big) \right),
\end{equation}
where $W$ is the size of a time window, $L$ is a performance measure and $\hat{y}_t = f_t(x_t)$ is the prediction of the model $f_t$ for the observation $x_t$. The $DTD_{t_d}$ corresponds to the difference between the mean performance in a window of size $W$ before and after a known concept drift at $t_d$. Similarly, we define the \textit{drift restoration time}:
\begin{align}\label{eq:drift_restoration}
    DRT_{t_d} &= t_{res} - t_d, \text{with}\\
    t_{res} &= \min\left\{t ~|~ t \geq t_d \text{ and } L\big(y_t, \hat{y}_t\big) \leq \frac{1}{W}\sum^W_{w=1} L\big(y_{t_d-w}, \hat{y}_{t_d-w}\big)\right\} \nonumber
\end{align}
The $DRT_{t_d}$ corresponds to the first time step after a known drift $t_d$, at which the average predictive performance before the drift has been restored. In the definition of $t_{res}$ in Eq. \eqref{eq:drift_restoration}, we have assumed that the measure $L$ is decremental, i.e., small values returned by $L$ correspond to a good performance. If $L$ were instead incremental (i.e., a higher $L$ is better), we would need to replace the ``$\leq$'' condition with ``$\geq$''. 

%Besides, note that both DPD and DRT are computed with respect to a known drift position $t_d$. Thus, unlike earlier performance measures such as accuracy or noise variability (Eq. \eqref{eq:noise_variability}), these measures do not change during time periods of stable data concepts.

To identify and evaluate sensible model adaptations over time, and to compute the above measures, we need ground truth information on known concept drifts. Indeed, without ground truth, it would remain unclear whether model adaptations and performance variations were caused by concept drift, noisy data, unstable model behaviour or random effects. Unfortunately, there are few available real-world data sets with known concept drift. We discuss the selection of adequate data sets in Section \ref{sec:datasets}.

\subsubsection{Interpretability}
Machine learning models are increasingly used in highly sensitive applications like online banking, medical diagnoses or job application systems. With new data protection regulations (e.g., the General Data Protection Regulation of the European Union), the transparency and interpretability of these models has gained considerable attention \cite{miller2019,Guidotti2019,ntoutsi2020bias,pawelczyk2021carla,carvalho2019machine}. Surprisingly, however, the interpretability of online learning methods is still relatively unexplored.

In general, we say that a model is interpretable, if its internal mechanics can be understood by a human. Unfortunately, there is no objective measure of interpretability. Intrinsic interpretability is thus often linked to the complexity of a model \cite{miller2019,bibal2016interpretability}, i.e., the lower the complexity, the higher the interpretability. For example, linear models and shallow decision trees are widely considered to be inherently interpretable. If we consider model complexity as an indicator of interpretability, we can use different evaluation measures such as the depth and the number of splits in a decision tree \cite{moshkovitz2021connecting,haug2022dynamic}, or the number of non-zero parameters in a linear model. Still, such heuristics should be treated with caution, as they usually do not allow for a comparison of different model families.

Moreover, the complexity of an online learning model is generally subject to change over time. Therefore, to achieve good interpretability in the above sense, the changes and updates of model complexity should also be interpretable. For example, the split and prune decisions in an incremental decision tree can be made more interpretable by tying them to shifts in the approximate data concept via meaningful properties \cite{haug2022dynamic}. Likewise, reliable concept drift detection methods could improve the interpretability of online learning \cite{Haug2021}. However, these temporal dynamics make quantifying interpretability even more difficult. In general, the interpretability of online learning models remains an open issue.

\paragraph{Digression: Post-Hoc Explanations in Data Streams}
Post-hoc explanation methods allow us to explain the predictions of complex and black-box models \cite{du2019techniques,Guidotti2019}. Explanation methods like local feature attributions \cite{ribeiro2016should,lundberg2017unified,kasneci2016licon} can generally also be used to explain (non-interpretable) online learning models. Moreover, these explanations may be evaluated with common techniques, e.g., based on feature ablation tests \cite{hooker2019benchmark,Haug2021a}. However, most post-hoc explanation methods, as well as the corresponding evaluation techniques, assume that the complex model has been trained and is therefore stationary. Accordingly, for online learning models, these methods can usually only provide a snapshot for a particular training phase. That is, we can use them to explain the online learning model at a particular time step $t$. However, there is generally no guarantee that the explanation at a given time step will be valid in the future. Indeed, one should be aware of changes in post-hoc explanations caused by incremental model updates and concept drift. Therefore, instead of trying to explain black-box predictions at individual time steps, we should aim for online learning models that are inherently interpretable.

\subsection{Properties For Active Concept Drift Detection}\label{sec:drift_properties}
In addition to predictive modelling, machine learning in data streams includes specific tasks required to preprocess streaming observations or deal with temporal change. While the properties presented above generally also apply to these special tasks, additional challenges and properties arise.

One such special task is active concept drift detection. Concept drift detection methods are used to identify changes in the data generating distribution over time. As mentioned earlier, concept drift detection methods often allow for more effective retraining of obsolete parts of a model. This makes them a powerful tool for avoiding performance degradation in the presence of concept drift. Concept drift detection can also improve overall interpretability by revealing hidden dynamics in the data stream. In general, a concept drift detection model should be able to detect concept drift in time and with few false alarms \cite{Gama2012,Goncalves2014,Lu2018}. As mentioned in Section \ref{sec:adaptability_drift}, we require ground truth information about known drifts to evaluate these properties.

\subsubsection{Detection Truthfulness}
Our goal is to detect every concept drift while minimizing the number of false alarms. To evaluate this property, previous works adopt similar performance measures, albeit using different terminology \cite{Goncalves2014,Bifet2013,Krawczyk2017}. For clarification, we define the most important measures below. 

Let $T_d$ be the set of time steps corresponding to known concept drifts and let $\hat{T}_d$ be the set of detected drifts of a concept drift detection method. We begin by defining the \textit{detected change rate}:
\begin{equation}\label{eq:det_change_rate}
    DCR = \frac{1}{|T_d|} \sum_{t_d \in T_d} \mathbbm{1}\left(\big\{t \in \hat{T}_d ~|~ t_d \leq t \leq t_d + W\big\} \neq \emptyset \right),
\end{equation}
where $|\cdot|$ is the cardinality of a set and $\mathbbm{1}$ is an indicator function that returns 1 if the condition in parentheses is met and 0 otherwise. The window size $W$ can be used to define an interval in which a drift detection is counted as a true positive. This can be useful, as drift detection methods are usually not able to detect a concept drift immediately -- in particular, if the change has small magnitude. The \textit{missed detection rate} introduced in \cite{Bifet2013} is equal to $1-DCR$.

The \textit{false discovery rate} (also known as false positive rate or false alarm rate) is another popular measure:
\begin{equation}\label{eq:fals_disc_rate}
    FDR = 1 - \frac{1}{|\hat{T}_d|} \sum_{t_d \in T_d} \left|\big\{t \in \hat{T}_d ~|~ t_d \leq t \leq t_d + W\big\} \right|
\end{equation}
Additionally, we may compute the \textit{time between false alarms} \cite{Bifet2013}, which is quantified by the number of time steps (or the number of streaming observations) between false drift detections. Good concept drift detection corresponds to a high time between false alarms.

Each above measure can easily be optimised for itself. Specifically, if a method detects concept drift at every time step, it would maximise the detected change rate. Similarly, if a method does not detect a single concept drift, it would minimize the false discovery rate and maximize the time between false alarms. Hence, for a reliable evaluation of active concept drift detection, the measures must be combined. 

\subsubsection{Detection Timeliness} Aside from accurately detecting concept drifts, we also want to reduce the \textit{delay} between known drifts and corresponding detections. A short delay can be crucial in practice to avoid long-term deterioration of predictive performance. The detection delay measures the number of time steps (or alternatively the number of observations) between a known drift and the first corresponding detection \cite{Krawczyk2017}. 

The mean delay (MD), the mean time between false alarms (MTFA) and the detected change rate (DCR, Eq. \eqref{eq:det_change_rate}) can also be aggregated in a \textit{mean time ratio} measure (MTR) \cite{Bifet2013}:
\begin{equation}\label{eq:mean_time_ratio}
    MTR = \frac{MTFA}{MD} \times DCR
\end{equation}
The mean time ratio is a simple approach to combine the two fundamental properties of active concept drift detection. However, this measure should be used with caution, as the time between false alarms and the delay are not normalized, which can lead to very different mean time ratios depending on the data set at hand.

Unlike previous measures, the performance measures for active concept drift detection only need to be calculated once at the end of the evaluation. Hence, these results are not influenced by the evaluation strategy that we apply (see Section \ref{sec:strategies}).

\subsection{Properties For Online Feature Selection}
By reducing the input dimensionality, feature selection methods often allow for more efficient and discriminative online learning. Similar to concept drift detection, online feature selection poses additional requirements, which we discuss in the following.

\subsubsection{Feature Set Stability}
In offline learning scenarios, feature selection is usually performed once before model training. Conversely, online feature selection models need to be updated over time, since the importance of features may shift with concept drift. Yet, large variations of the selected features between time steps can be perceived as unintuitive or non-robust \cite{Haug2020}. Besides, frequent changes to the selected features may entail excessive and costly updates of the predictive model. Consequently, we generally aim for stable feature sets over time -- in particular, if the data concept is known to be stable. In this sense, feature set stability is related to the general \textit{robustness to noise} property introduced above.

The stability of feature sets in offline learning scenarios has attracted attention in the past \cite{bommert2021stabm,kalousis2005stability}. In \cite{Haug2020}, the authors proposed an adaptation of a popular offline stability measure due to \cite{nogueira2017stability} for the evaluation of online feature selection methods. Accordingly, let $a_t \in \{0,1\}^m$ be the active feature vector at time step $t$, where $m$ is the total number of features. In the vector $a_t$, selected features are represented by ones and unselected features are represented by zeros. The \textit{feature selection stability} at time step $t$ for a sliding window of size $w$ is then defined as:
\begin{equation}\label{eq:stability}
    FSS_{t,w} = 1 - \frac{\frac{1}{m} \sum^m_{j=1} s^2_j}{\frac{k}{m}\left(1-\frac{k}{m}\right)},
\end{equation}
where $k$ is the number of selected features and $s^2_j = \frac{w}{w-1}\hat{p}_j(1-\hat{p}_j)$ is the unbiased sample variance of the selection of feature $j$, with $\hat{p}_j = \frac{1}{w}\sum^{w-1}_{i=0} a_{t-i,j}$. To compute the stability between the consecutive feature sets at $t-1$ and $t$, we can set the window size to $w=2$. The feature set stability due to \eqref{eq:stability} decreases, if the total variability of the selected features $\sum^m_{j=1} s^2_j$ increases. Conversely, the stability is maximized if $s^2_j = 0$ for all features $j$, i.e., if the selected feature set remains stationary over the full length of the sliding window.

Offline feature set stability measures do not take into account concept drift, where we would normally tolerate some degree of variability. Hence, stability measures adopted from the offline literature, as proposed above, should be considered with care. In general, measuring the stability of feature sets in the presence of concept drift is an open problem.

\subsubsection{Feature Selectivity}
With most online feature selection methods, the size of the returned feature set must be specified in advance. However, if a feature selection method is able to automatically determine the ideal feature set size, the reduction rate could be another useful evaluation measure \cite{RamirezGallego2017}. Specifically, the reduction rate indicates the percentage of original features that has not been selected. Given similar predictive performance, the smaller feature set, i.e., the larger reduction rate, is generally preferable.

\subsection{Additional Considerations}
In the following, we briefly list additional properties and considerations that should be taken into account when developing and evaluating online machine learning methods. Although each individual point would deserve its own section, this is beyond the scope of this paper.

\begin{itemize}
    \item \textbf{Label Delay}: In practice, labelling information is often costly and only becomes available some time after the corresponding observation. In fraud prevention scenarios, for example, a fraudulent transaction may not be recognized for several days or weeks. For the sake of simplicity, we often assume that the labels for all observations are available immediately. However, powerful online learning methods should be able to handle delayed label information.
    \item \textbf{Normalization}: Normalization can dramatically improve the performance of a predictive model. Unfortunately, it is not possible to normalize streaming data in one go. In fact, feature scales might shift over time due to concept drift. Accordingly, if the feature scales are not known in advance, the information used to normalize incoming observations should be iteratively updated. In this context, one should also take care of outliers, which can considerably impact the normalization (e.g., in the common min-max scaling). In general, the normalization of streaming observations should receive more attention.
    \item \textbf{Imbalanced Targets}: The target class of streaming data might be heavily imbalanced (e.g., in the context of detecting credit card fraud). Indeed, imbalances might be subject to temporal change. For example, credit card transactions have a higher frequency in the evenings and during weekends. Therefore, we require online learning methods to robustly handle imbalanced target distributions. Likewise, we need to select performance measures that are meaningful in the presence of imbalances, e.g. the F1 measure for classification.
    \item \textbf{Feature/Concept Evolution}: Concept drift may alter the set of features and classes that we observe over time. We call these phenomena \textit{feature evolution} and \textit{concept evolution}, respectively \cite{masud2010classification}. For example, new trending topics (i.e., classes) on social media might temporarily produce new hashtags (i.e., features). The occurrence of feature evolution and concept evolution poses additional difficulties for online machine learning methods.
\end{itemize}

\section{Finding Real-World Benchmark Data Sets}\label{sec:datasets}
Many evaluation measures presented above require ground truth information about known concept drift in order to be calculated. However, it is often impossible to obtain ground truth from real-world processes. Although there are approaches to induce artificial concept drift to real-world data sets \cite{sethi2017reliable,Haug2021}, we argue that there is a fundamental shortage of adequate benchmark data sets for the evaluation of online learning methods. 

Indeed, synthetically generated data streams still dominate the literature. Generating synthetic data with various kinds of concept drift is straight-forward through packages like scikit-multiflow \cite{montiel2018scikit} or river \cite{montiel2021river}. A comprehensive collection of synthetic data streams has been proposed by the authors in \cite{manapragada2020eager}.

Yet, there is as series of real-world data sets that are frequently used for the evaluation of online learning methods. The authors in \cite{souza2020challenges} provide an extensive summary and discussion of some of the most popular real-world data sets for online learning. In addition, they propose a collection of insect classification data sets, which comprise natural concept drift. The \textit{Insects} data sets comprise sensor information from flying insect species, obtained in a controlled environment. By adjusting the temperature and humidity, the authors obtained different types of natural concept drift. Another recently published data set with known concept drift is \textit{TüEyeQ} \cite{kasneci2021tueyeq,kasneci2020tueeyeq}. TüEyeQ comprises sociodemographic information about participants in an IQ test. The data contains natural concept drifts by switching between different task blocks and increasing difficulty within each block.

Finally, there are various public sources from which streaming data sets can be obtained: 
\begin{itemize}
    \item \url{https://www.openml.org/search?type=data} (search for "data stream" or "concept drift")
    \item \url{https://sites.google.com/view/uspdsrepository} \cite{souza2020challenges}
    \item \url{https://github.com/ogozuacik/concept-drift-datasets-scikit-multiflow}
    \item \url{https://github.com/vlosing/driftDatasets}
\end{itemize}

\section{The ``float'' Evaluation Package}\label{sec:float}
Along with this paper, we introduce \textit{float}. Float is a modular Python framework for simple and more standardized evaluations of online learning methods. Our framework provides easy and high-level access to popular evaluation strategies and measures, as described above. In this way, float handles large parts of the evaluation and reduces the possibility of human error. Float enables joint integration of popular Python libraries and custom functionality. Accordingly, float is a meaningful extension of comprehensive libraries like scikit-multiflow \cite{montiel2018scikit} or river \cite{montiel2021river}. In this sense, \textit{float is not intended to be another library of state-of-the-art models}. Rather, our goal is to provide tools for creating high-quality experiments and visualisations.

\subsection{Access and Code Quality}
Float is distributed under the MIT license. The framework can currently be accessed via Github (\url{https://github.com/haugjo/float}) or the Python packaging index Pypi (\url{https://pypi.org/project/float-evaluation/}). The source code of float is fully documented according to the Google docstring standard. The documentation can also be accessed at \url{https://haugjo.github.io/float/}. To ensure the quality and readability of our source code, we applied the PEP8 formatting standard. Moreover, we created an extensive set of unit tests to validate all core functionality. The test suite is available on Github.

\subsection{Modularity}
The source code of float is completely modular. We encapsulate related functionality in Python classes. Specifically, there are classes for online prediction, concept drift detection, and online feature selection, as well as corresponding classes for their evaluation. Users can integrate their own models by inheriting from abstract base classes. The evaluation strategies discussed in Section \ref{sec:strategies} are implemented as pipelines (which are also Python classes). The pipelines allow users to specify custom experiments and run any combination of concept drift detection, online feature selection, and predictive models. Indeed, with float it is possible to configure a pipeline that combines custom models and common Python packages. For example, within the same pipeline, we may load a data set via the scikit-multiflow \textit{FileStream} \cite{montiel2018scikit}, implement a custom online classifier, and use scikit-learn metrics for the evaluation. Besides, float provides a number of visualisations that can be used to illustrate the results of the pipeline run. Float also includes various recent and state-of-the-art online learning methods that are not part in any of the major libraries yet. We plan to extend the set of available performance measures, preprocessing techniques, and evaluation strategies in the future and welcome contributions by the community.

%------------- Example Plot 1-------------------------%
\begin{figure}[thbp]
\centering
\subfloat[F1 Performance Measure]{
    \includegraphics[width=.94\columnwidth]{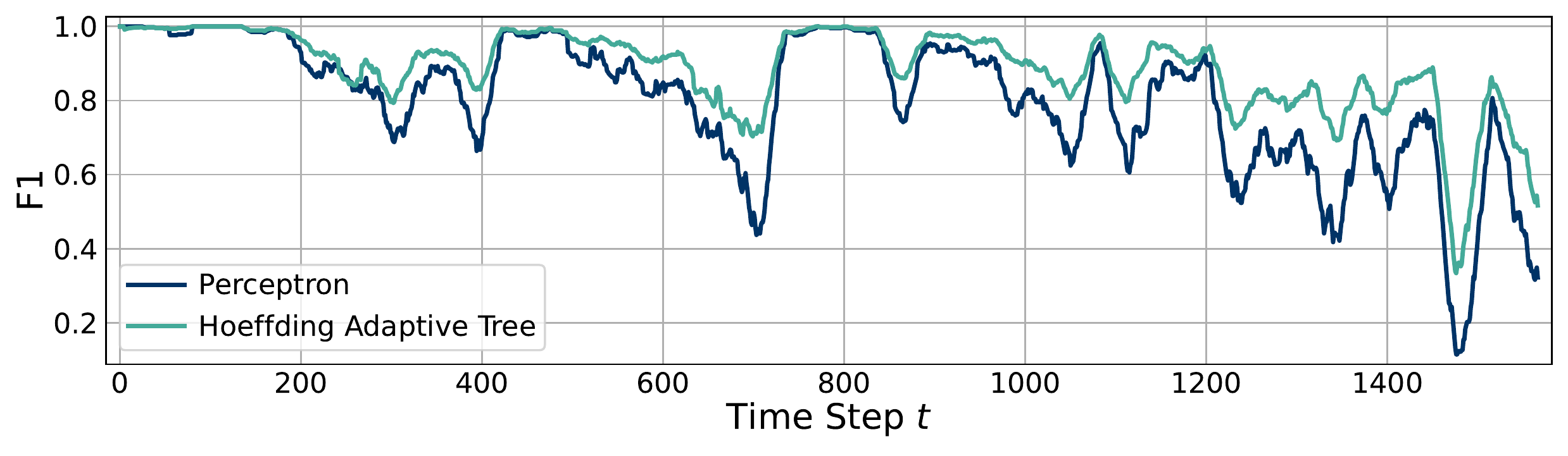}
}
\hfill
\subfloat[Noise Variability (with respect to F1), see Eq. \eqref{eq:noise_variability}.]{
    \includegraphics[width=.94\columnwidth]{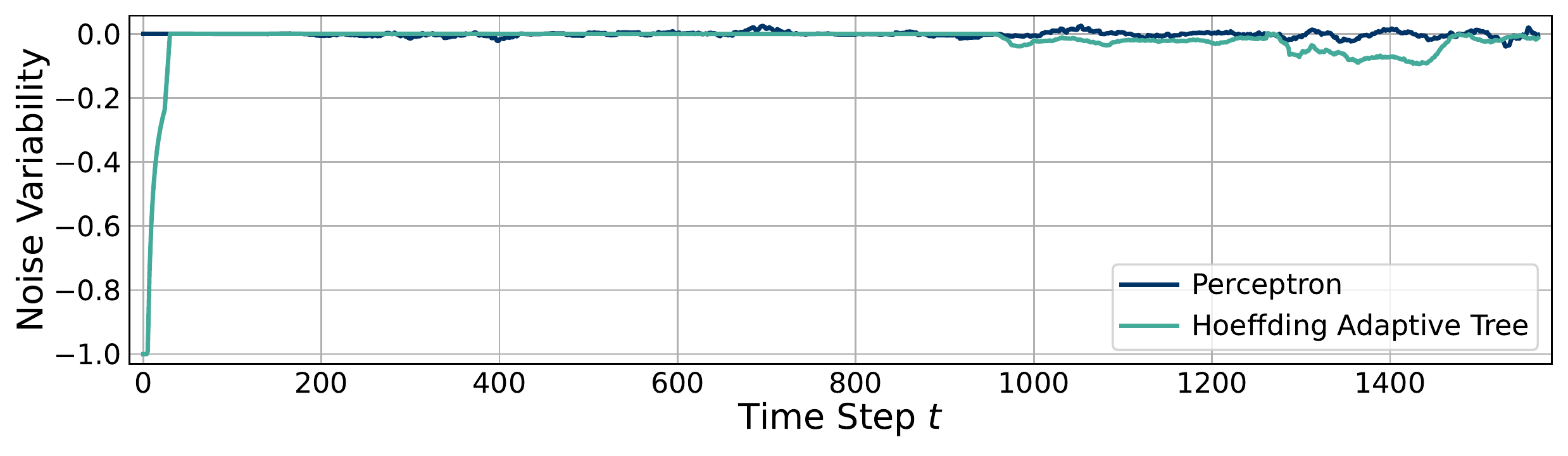}
}
\hfill
\subfloat[Drift Restoration Time (in mean no. of time steps), see Eq. \eqref{eq:drift_restoration}.]{
    \includegraphics[width=.94\columnwidth]{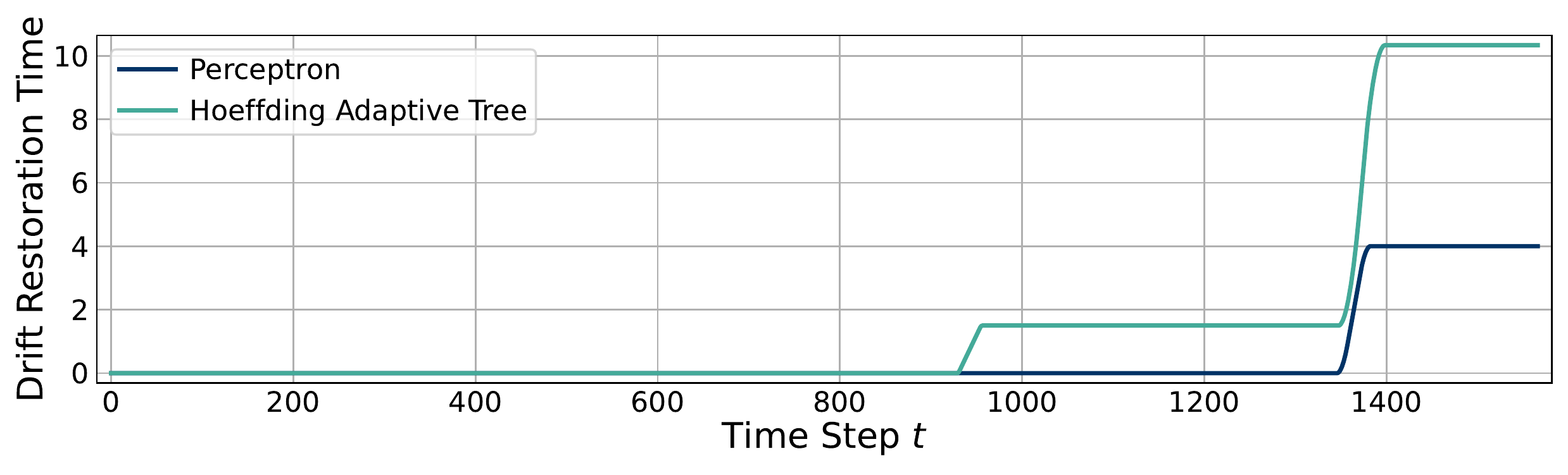}
}
\caption{\textbf{Illustrating Results with Float.} The float framework contains a visualisation module that provides adaptations of common plot types such as line, scatter or bar plots. In this way, float enables a quick and intuitive visualisation of the results stored in an evaluator object. Above, we see the results of the example described in Section \ref{sec:example}.}
\label{fig:initial_test}
\end{figure}
%------------- Example Plot 1-------------------------%

\subsection{Usage}
It is neither meaningful nor feasible to describe all modules and configurations of float in this paper. However, on Github we provide detailed documentation and multiple exemplary experiments in the form of Jupyter notebooks that can help users to familiarize themselves with the float framework and its modules. For illustration, we describe a simple experiment with float below.

\subsubsection{Exemplary Experiment}\label{sec:example}
In this example, we want to train an online predictive model on the TüEyeQ data set \cite{kasneci2021tueyeq}. The classification task is to decide whether a task was passed or failed given a vector of task-specific features and socio-demographic information about the corresponding subject (77 features in total). Note that we will not optimize any hyperparameters of the models involved, as this is only an illustrative experiment. However, for practical applications, float allows us to run multiple configurations of a predictive model in parallel for effective hyperparameter optimization.

We start by comparing a Perceptron model and a Hoeffding Adaptive Tree \cite{bifet2009adaptive}. The source code for this first experiment is provided in Figure \ref{fig:python_code}. We load the data set file with the \textit{DataLoader} module of float. Then we set up the two predictive models. In order to use implementations of scikit-multiflow \cite{montiel2018scikit}, we need to wrap the corresponding objects within a \textit{SkmultiflowClassifier} object. Next, we specify the \textit{PredictionEvaluator} object, which calculates and stores the performance measures of both models. In particular, we instruct the evaluator to compute the F1 measure, the noise variability (Eq. \eqref{eq:noise_variability}), and the drift restoration time (Eq. \eqref{eq:drift_restoration}). For the latter measure, we need information about known concept drifts, which we provide as a hyperparameter. We also specify the batch size that we will use in the prequential evaluation. Note that we can specify any hyperparameter of a measure function directly in the constructor of the float evaluator object. For example, we may set the \textit{zero\_division} parameter of the scikit-learn \textit{f1\_score} function, as well as the \textit{reference\_measure} and \textit{n\_samples} parameters of the noise variability measure. In addition to the raw performance measurements, we also want to obtain the performance aggregated in a sliding window. To this end, we specify a sliding window size of 25. Finally, we set up a prequential pipeline and provide all previously initialised objects. We use 100 observations to pre-train the online learning models before starting the prequential evaluation with a batch size of 10.

%------------- Source Code Sample -------------------------%
\begin{figure}[hbtp]
\centering
\begin{python}
from skmultiflow.trees import HoeffdingAdaptiveTreeClassifier
from skmultiflow.neural_networks import PerceptronMask
from sklearn.metrics import f1_score
from float.data import DataLoader
from float.prediction.evaluation import PredictionEvaluator
from float.prediction.evaluation.measures import noise_variability,
  mean_drift_restoration_time
from float.pipeline import PrequentialPipeline
from float.prediction.skmultiflow import SkmultiflowClassifier

# Load a data set from main memory with the DataLoader module. 
# Alternatively, we can provide a scikit-multiflow FileStream ...
# ... object via the 'stream' attribute.
data_loader = DataLoader(path='./datasets/iq.csv', 
                         target_col=-1)
known_drifts = [4707, 9396, 13570]  # Known drift positions

# Set up online classifiers. 
# Note that we need a wrapper to use scikit-multiflow functionality.
models = [SkmultiflowClassifier(model=PerceptronMask(), 
                                classes=
                                 data_loader.stream.target_values),
          SkmultiflowClassifier(model=
                                 HoeffdingAdaptiveTreeClassifier(), 
                                classes=
                                 data_loader.stream.target_values)]

# Set up an evaluator object for the classifiers:
# Specifically, we want to measure the f1_score, ...
# ... the noise_variability and the drift_restoration_time.
# The arguments of the measure functions can be directly added to ...
# ... the evaluator object constructor, e.g. we may specify ...
# ... the number of samples (n_samples) and the reference_measure ...
# ... used to compute the noise_variability.
evaluator = PredictionEvaluator(measure_funcs=[f1_score, 
                                 noise_variability, 
                                 mean_drift_restoration_time],
                                window_size=25,
                                zero_division=0,
                                reference_measure=f1_score,
                                n_samples=15,
                                batch_size=10,
                                known_drifts=known_drifts)

# Set up a pipeline for a prequential evaluation of the classifiers.
pipeline = PrequentialPipeline(data_loader=data_loader,
                               predictor=models,
                               prediction_evaluator=evaluator,
                               n_max=data_loader.stream.n_samples,
                               batch_size=10, 
                               n_pretrain=100)

# Run the experiment.
pipeline.run()
\end{python}
\caption{\textbf{Conducting Data Stream Experiments in Float.} Here we show the source code for a simple experiment performed with the proposed float evaluation framework. More experiments and a detailed documentation can be found on our Github page.}
\label{fig:python_code}
\end{figure}
%------------- Source Code Sample -------------------------%

To compare the performance of the two classifiers, we show the aggregated F1 score, the noise variability and the drift restoration time in Figure \ref{fig:initial_test} (using the line plot type of float). All displayed measures are stored in the evaluator object. Based on the predictive performance of the classifiers, we can clearly see the concept drift in TüEyeQ. In particular, we observe that the predictive performance of the classifiers within each task block starts to suffer when the IQ-related tasks become more difficult to solve. However, the drift restoration time only increase for the last two concept drifts. In general, the Perceptron algorithm performs worse than the Hoeffding Adaptive Tree in terms of F1, but slightly better in terms of noise variability and drift restoration time. Since the Perceptron model does not actively adapt to concept drift, we might further improve performance by using a drift detection method to trigger active retraining.

For the following experiments, we no longer provide the corresponding source code in order to maintain brevity. However, the general process of specifying float objects remains the same. In addition, the float documentation contains detailed experiment notebooks for each of the modules used.

%------------- Example Plot 2-------------------------%
\begin{figure}[t]
\centering
\subfloat[Concept Drift Detection]{
    \includegraphics[width=.94\columnwidth]{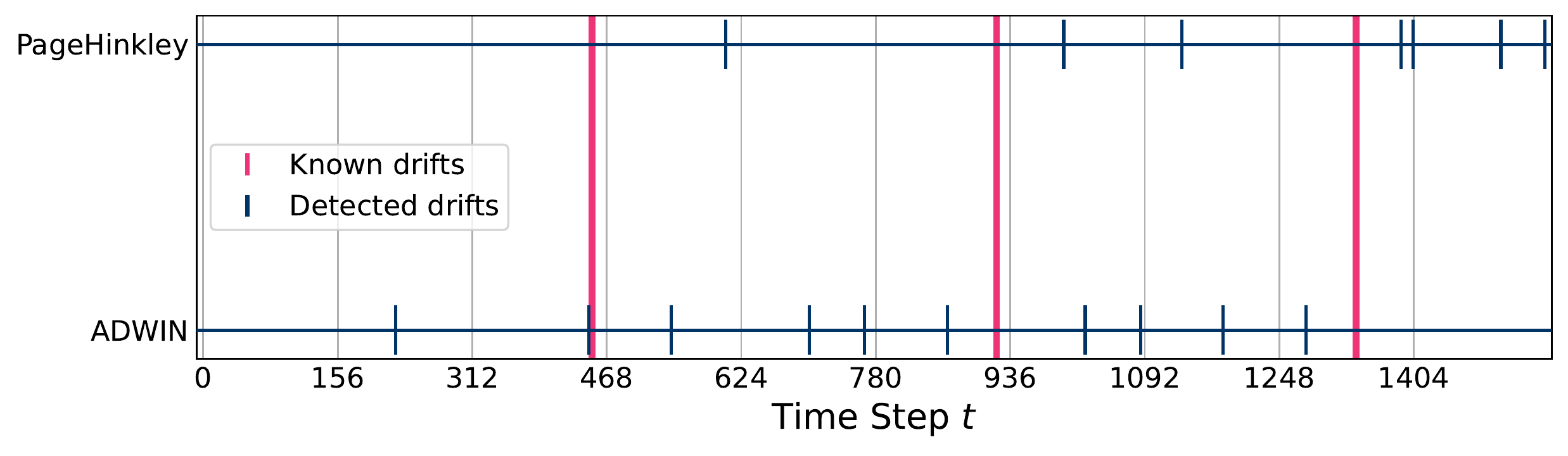}
}
\hfill
\subfloat[Online Feature Selection]{
    \includegraphics[width=.94\columnwidth]{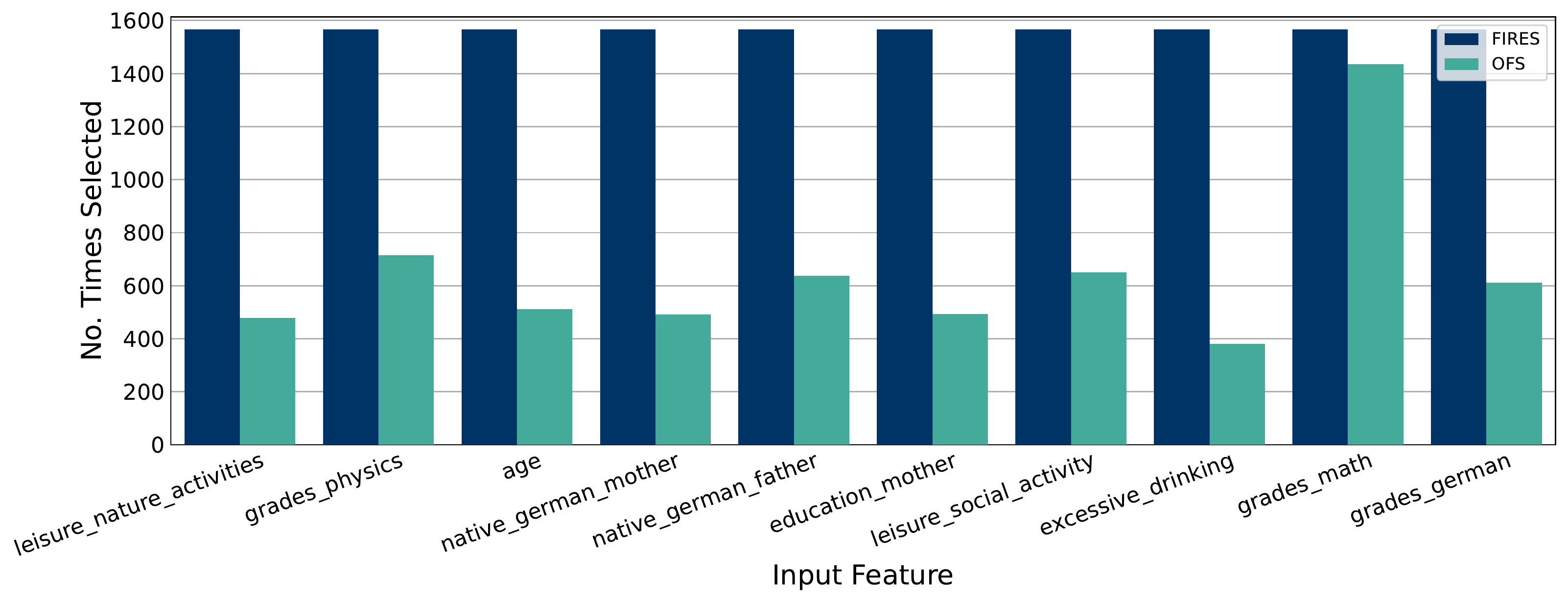}
}
\caption{\textbf{Plotting Concept Drift Detection and Online Feature Selection.} Float provides special plot types for evaluating concept drift detection methods and online feature selection models. Above we compare the detected drifts and the most frequently selected features as described in the simple experiment in Section \ref{sec:example}.}
\label{fig:drift_detection_feature_selection}
\end{figure}
%------------- Example Plot 2-------------------------%

We start by comparing two popular concept drift detection methods: an Adaptive Sliding Window (ADWIN) \cite{bifet2007learning} and a Page-Hinkley test \cite{page1954continuous}. As before, we want to use the corresponding implementation of scikit-multiflow. Float also offers wrapper classes for concept drift detectors from related libraries. We specify a \textit{ChangeDetectionEvaluator} object to compute the detected change rate (Eq. \eqref{eq:det_change_rate}), the false discovery rate (Eq. \eqref{eq:fals_disc_rate}) and the delay. We set a window size of 500 within which we count a drift alarm as a true positive. Float allows the comparison of multiple drift detection models using a special plot type, as shown in Figure \ref{fig:drift_detection_feature_selection}. With a detected change rate of 0.33\%, a false discovery rate of 0.85\% and an average delay of 852 observations, the Page-Hinkley test outperforms ADWIN in the TüEyeQ dataset, although no model performs particularly well. Indeed, if we use Page-Hinkley to reset the Perceptron (which can be done by setting the \textit{reset\_after\_drift} parameter of the \textit{SkmultiflowClassifier} to True), we get no improvement in the average F1 score. Hence, we continue without concept drift detection.

\begin{figure}[t]
\centering
\includegraphics[width=.6\columnwidth]{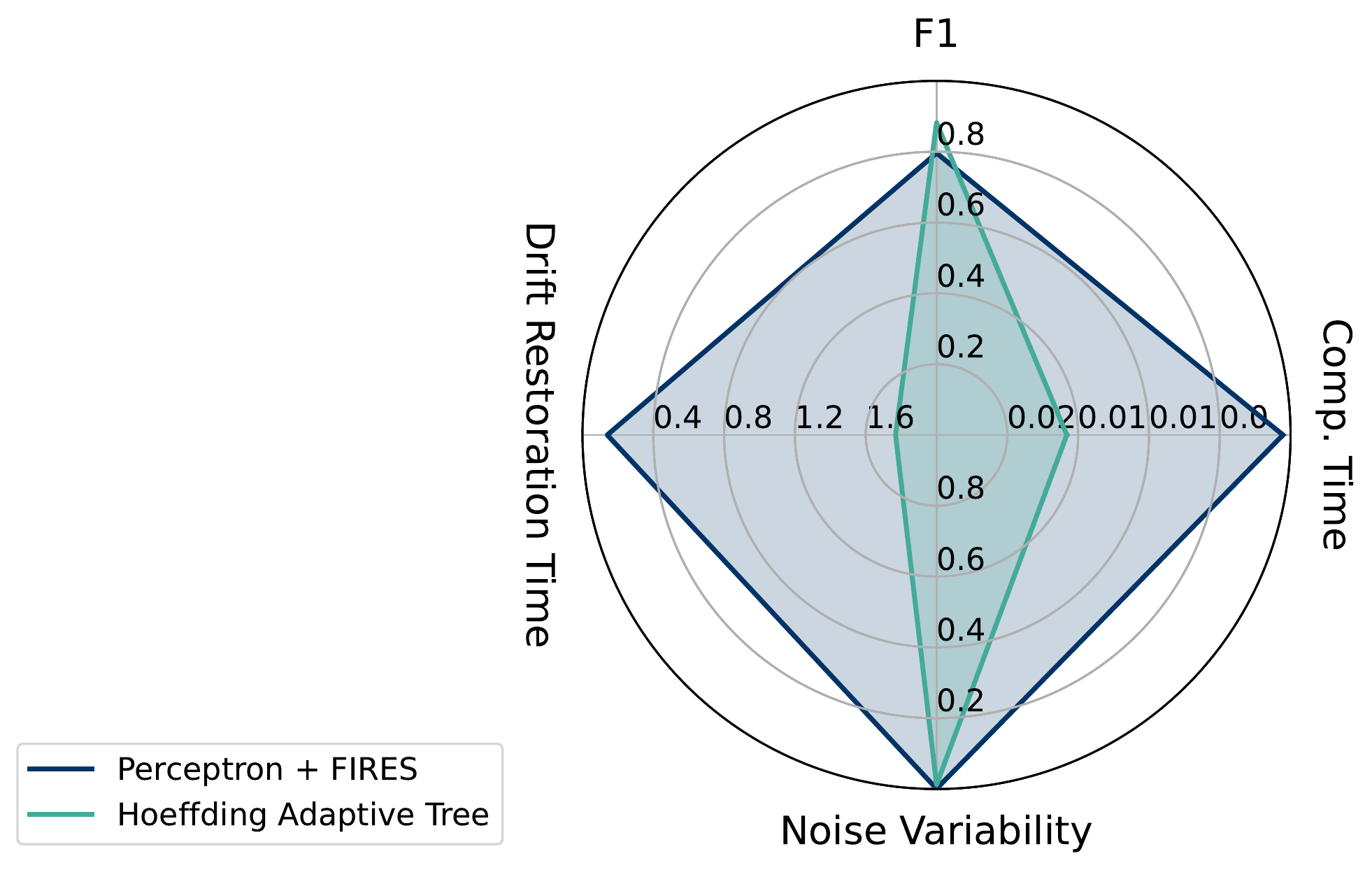}
\caption{\textbf{Comparing Multiple Properties via the Spider Chart.} The float visualisation module contains a spider chart that provides a concise comparison of competing models in terms of various properties. Here we show a summary of the experimental results described in Section \ref{sec:example}.}
\label{fig:spider}
\end{figure}

Finally, we would like to investigate whether we can achieve improvements in any of the performance measures through online feature selection. Accordingly, we compare the online feature selection models FIRES \cite{Haug2020} and OFS \cite{wang2013online}. Both model implementations are provided by float. As before, float includes a dedicated \textit{FeatureSelectionEvaluator}, which we use to compute the feature set stability (Eq. \eqref{eq:stability}) of each approach. In this example, we want to select 25 features. The most frequently selected features can again easily be compared with the float visualization module (see Figure \ref{fig:drift_detection_feature_selection}). The FIRES model outperforms OFS in terms of the feature set stability (0.99 for FIRES and 0.86 for OFS). Moreover, FIRES improves the drift restoration time from 0.5 to 0.4, while OFS worsens the value to 1.08. For the Perceptron, feature selection also leads to a small improvement in the average F1 score. Since the Hoeffding Adaptive Tree performs implicit feature selection, a dedicated feature selection model does not improve performance. Therefore, in our final configuration, we compare the Perceptron in combination with FIRES with the stand-alone Hoeffding Adaptive Tree. We use the spider chart of float to compare both models one last time with regard to various criteria (see Figure \ref{fig:spider}). In our example, the Perceptron with FIRES has advantages in terms of computation time and drift restoration time, but performs worse than the Hoeffding Adaptive Tree regarding the F1 measure. Both models show little variability with noisy inputs. The proposed float framework allowed us to compare these models in a standardized way and with little effort.

%------------- Example Plot 3-------------------------%
%------------- Example Plot 3-------------------------%

\section{Conclusion}
Evolving data streams are found in most large-scale and everyday web applications. In this work, we revisited the challenges of evaluating machine learning methods for dynamic data streams. We proposed a comprehensive set of evaluation properties and performance measures that, unlike previous work, extend to the specific tasks of online feature selection and concept drift detection. To enable a more transparent and standardized comparison of online learning methods, we introduced float. Float is a modular and extensible Python framework that can automate major parts of the simulation and evaluation process and provides a flexible basis for extensive benchmarking. The experiments shown in this paper only give a brief impression of the power and versatility of float. We believe that our work can serve as an important reference for the evaluation of online learning models. With this in mind, we hope that this work will help raise awareness of the importance and practical use of online machine learning and data streams.

%\section*{Appendix}
%Text for this section\ldots

%%%%%%%%%%%%%%%%%%%%%%%%%%%%%%%%%%%%%%%%%%%%%%
%%                                          %%
%% Backmatter begins here                   %%
%%                                          %%
%%%%%%%%%%%%%%%%%%%%%%%%%%%%%%%%%%%%%%%%%%%%%%

\begin{backmatter}

\section*{Abbreviations}%% if any
NV: Noise Variability; DPD: Drift Performance Deterioration; DRT: Drift Restoration Time; DCR: Detected Change Rate; FDR: False Discovery Rate; MD: Mean Delay; MTFA: Mean Time Between False Alarms; MTR: Mean Time Ratio; FSS: Feature Set Stability; ADWIN: Adaptive Windowing; FIRES: Fast, Interpretable and Robust feature Evaluation and Selection; OFS: Online Feature Selection.

\section*{Declarations}

\section*{Ethics approval and consent to participate}%% if any
Not applicable.

\section*{Consent for publication}%% if any
Not applicable.

\section*{Availability of data and materials}%% if any
The TüEyeQ data set analysed during the current study is available in the Harvard Dataverse repository, \url{https://doi.org/10.7910/DVN/JGOCKI} \cite{kasneci2021tueyeq,kasneci2020tueeyeq}. The float framework presented in this paper is available at \url{https://github.com/haugjo/float} and \url{https://pypi.org/project/float-evaluation/}. The documentation of float is available at \url{https://haugjo.github.io/float/}.

\section*{Competing interests}
The authors declare that they have no competing interests.

\section*{Funding}%% if any
Not applicable.

\section*{Authors' contributions}
JH conceived this study and drafted the manuscript. JH and ET implemented the float framework. JH and GK revised the manuscript. All authors read and approved the final manuscript.

\section*{Acknowledgements}%% if any
Not applicable.

%%  After submission of the .TEX file,                     %%
%%  you will be prompted to submit your .BBL file.         %%

% if your bibliography is in bibtex format, use those commands:
%\bibliographystyle{bmc-mathphys} % Style BST file (bmc-mathphys, vancouver, spbasic).
%\bibliography{base}      % Bibliography file (usually '*.bib' )


\begin{thebibliography}{52}
\ifx \bisbn   \undefined \def \bisbn  #1{ISBN #1}\fi
\ifx \binits  \undefined \def \binits#1{#1}\fi
\ifx \bauthor  \undefined \def \bauthor#1{#1}\fi
\ifx \batitle  \undefined \def \batitle#1{#1}\fi
\ifx \bjtitle  \undefined \def \bjtitle#1{#1}\fi
\ifx \bvolume  \undefined \def \bvolume#1{\textbf{#1}}\fi
\ifx \byear  \undefined \def \byear#1{#1}\fi
\ifx \bissue  \undefined \def \bissue#1{#1}\fi
\ifx \bfpage  \undefined \def \bfpage#1{#1}\fi
\ifx \blpage  \undefined \def \blpage #1{#1}\fi
\ifx \burl  \undefined \def \burl#1{\textsf{#1}}\fi
\ifx \doiurl  \undefined \def \doiurl#1{\textsf{#1}}\fi
\ifx \betal  \undefined \def \betal{\textit{et al.}}\fi
\ifx \binstitute  \undefined \def \binstitute#1{#1}\fi
\ifx \binstitutionaled  \undefined \def \binstitutionaled#1{#1}\fi
\ifx \bctitle  \undefined \def \bctitle#1{#1}\fi
\ifx \beditor  \undefined \def \beditor#1{#1}\fi
\ifx \bpublisher  \undefined \def \bpublisher#1{#1}\fi
\ifx \bbtitle  \undefined \def \bbtitle#1{#1}\fi
\ifx \bedition  \undefined \def \bedition#1{#1}\fi
\ifx \bseriesno  \undefined \def \bseriesno#1{#1}\fi
\ifx \blocation  \undefined \def \blocation#1{#1}\fi
\ifx \bsertitle  \undefined \def \bsertitle#1{#1}\fi
\ifx \bsnm \undefined \def \bsnm#1{#1}\fi
\ifx \bsuffix \undefined \def \bsuffix#1{#1}\fi
\ifx \bparticle \undefined \def \bparticle#1{#1}\fi
\ifx \barticle \undefined \def \barticle#1{#1}\fi
\ifx \bconfdate \undefined \def \bconfdate #1{#1}\fi
\ifx \botherref \undefined \def \botherref #1{#1}\fi
\ifx \url \undefined \def \url#1{\textsf{#1}}\fi
\ifx \bchapter \undefined \def \bchapter#1{#1}\fi
\ifx \bbook \undefined \def \bbook#1{#1}\fi
\ifx \bcomment \undefined \def \bcomment#1{#1}\fi
\ifx \oauthor \undefined \def \oauthor#1{#1}\fi
\ifx \citeauthoryear \undefined \def \citeauthoryear#1{#1}\fi
\ifx \endbibitem  \undefined \def \endbibitem {}\fi
\ifx \bconflocation  \undefined \def \bconflocation#1{#1}\fi
\ifx \arxivurl  \undefined \def \arxivurl#1{\textsf{#1}}\fi
\csname PreBibitemsHook\endcsname

%%% 1
\bibitem{Lu2018}
\begin{barticle}
\bauthor{\bsnm{Lu}, \binits{J.}},
\bauthor{\bsnm{Liu}, \binits{A.}},
\bauthor{\bsnm{Dong}, \binits{F.}},
\bauthor{\bsnm{Gu}, \binits{F.}},
\bauthor{\bsnm{Gama}, \binits{J.}},
\bauthor{\bsnm{Zhang}, \binits{G.}}:
\batitle{Learning under concept drift: A review}.
\bjtitle{IEEE Transactions on Knowledge and Data Engineering 31, no. 12 (2018):
  2346-2363}
(\byear{2018}).
doi:\doiurl{10.1109/TKDE.2018.2876857}.
\arxivurl{2004.05785}
\end{barticle}
\endbibitem

%%% 2
\bibitem{ditzler2015learning}
\begin{barticle}
\bauthor{\bsnm{Ditzler}, \binits{G.}},
\bauthor{\bsnm{Roveri}, \binits{M.}},
\bauthor{\bsnm{Alippi}, \binits{C.}},
\bauthor{\bsnm{Polikar}, \binits{R.}}:
\batitle{Learning in nonstationary environments: A survey}.
\bjtitle{IEEE Computational Intelligence Magazine}
\bvolume{10}(\bissue{4}),
\bfpage{12}--\blpage{25}
(\byear{2015})
\end{barticle}
\endbibitem

%%% 3
\bibitem{gama2012survey}
\begin{barticle}
\bauthor{\bsnm{Gama}, \binits{J.}}:
\batitle{A survey on learning from data streams: current and future trends}.
\bjtitle{Progress in Artificial Intelligence}
\bvolume{1}(\bissue{1}),
\bfpage{45}--\blpage{55}
(\byear{2012})
\end{barticle}
\endbibitem

%%% 4
\bibitem{domingos2000mining}
\begin{bchapter}
\bauthor{\bsnm{Domingos}, \binits{P.}},
\bauthor{\bsnm{Hulten}, \binits{G.}}:
\bctitle{Mining high-speed data streams}.
In: \bbtitle{Proceedings of the Sixth ACM SIGKDD International Conference on
  Knowledge Discovery and Data Mining},
pp. \bfpage{71}--\blpage{80}
(\byear{2000})
\end{bchapter}
\endbibitem

%%% 5
\bibitem{holmes2005stress}
\begin{bchapter}
\bauthor{\bsnm{Holmes}, \binits{G.}},
\bauthor{\bsnm{Kirkby}, \binits{R.}},
\bauthor{\bsnm{Pfahringer}, \binits{B.}}:
\bctitle{Stress-testing hoeffding trees}.
In: \bbtitle{European Conference on Principles of Data Mining and Knowledge
  Discovery},
pp. \bfpage{495}--\blpage{502}
(\byear{2005}).
\bcomment{Springer}
\end{bchapter}
\endbibitem

%%% 6
\bibitem{bifet2010fast}
\begin{bchapter}
\bauthor{\bsnm{Bifet}, \binits{A.}},
\bauthor{\bsnm{Holmes}, \binits{G.}},
\bauthor{\bsnm{Pfahringer}, \binits{B.}},
\bauthor{\bsnm{Frank}, \binits{E.}}:
\bctitle{Fast perceptron decision tree learning from evolving data streams}.
In: \bbtitle{Pacific-Asia Conference on Knowledge Discovery and Data Mining},
pp. \bfpage{299}--\blpage{310}
(\byear{2010}).
\bcomment{Springer}
\end{bchapter}
\endbibitem

%%% 7
\bibitem{read2012batch}
\begin{bchapter}
\bauthor{\bsnm{Read}, \binits{J.}},
\bauthor{\bsnm{Bifet}, \binits{A.}},
\bauthor{\bsnm{Pfahringer}, \binits{B.}},
\bauthor{\bsnm{Holmes}, \binits{G.}}:
\bctitle{Batch-incremental versus instance-incremental learning in dynamic and
  evolving data}.
In: \bbtitle{International Symposium on Intelligent Data Analysis},
pp. \bfpage{313}--\blpage{323}
(\byear{2012}).
\bcomment{Springer}
\end{bchapter}
\endbibitem

%%% 8
\bibitem{barddal2019learning}
\begin{bchapter}
\bauthor{\bsnm{Barddal}, \binits{J.P.}},
\bauthor{\bsnm{Enembreck}, \binits{F.}}:
\bctitle{Learning regularized hoeffding trees from data streams}.
In: \bbtitle{Proceedings of the 34th ACM/SIGAPP Symposium on Applied
  Computing},
pp. \bfpage{574}--\blpage{581}
(\byear{2019})
\end{bchapter}
\endbibitem

%%% 9
\bibitem{barddal2020regularized}
\begin{barticle}
\bauthor{\bsnm{Barddal}, \binits{J.P.}},
\bauthor{\bsnm{Enembreck}, \binits{F.}}:
\batitle{Regularized and incremental decision trees for data streams}.
\bjtitle{Annals of Telecommunications}
\bvolume{75}(\bissue{9}),
\bfpage{493}--\blpage{503}
(\byear{2020})
\end{barticle}
\endbibitem

%%% 10
\bibitem{Gama2009}
\begin{bchapter}
\bauthor{\bsnm{Gama}, \binits{J.a.}},
\bauthor{\bsnm{Sebasti\~{a}o}, \binits{R.}},
\bauthor{\bsnm{Rodrigues}, \binits{P.P.}}:
\bctitle{Issues in evaluation of stream learning algorithms}.
\bsertitle{KDD '09},
pp. \bfpage{329}--\blpage{338}.
\bpublisher{Association for Computing Machinery},
\blocation{New York, NY, USA}
(\byear{2009}).
doi:\doiurl{10.1145/1557019.1557060}.
\burl{https://doi.org/10.1145/1557019.1557060}
\end{bchapter}
\endbibitem

%%% 11
\bibitem{Gama2012}
\begin{barticle}
\bauthor{\bsnm{Gama}, \binits{J.}},
\bauthor{\bsnm{Sebasti{\~{a}}o}, \binits{R.}},
\bauthor{\bsnm{Rodrigues}, \binits{P.P.}}:
\batitle{On evaluating stream learning algorithms}.
\bjtitle{Machine Learning}
\bvolume{90}(\bissue{3}),
\bfpage{317}--\blpage{346}
(\byear{2012}).
doi:\doiurl{10.1007/s10994-012-5320-9}
\end{barticle}
\endbibitem

%%% 12
\bibitem{Bifet2013}
\begin{botherref}
\oauthor{\bsnm{Bifet}, \binits{A.}},
\oauthor{\bsnm{Read}, \binits{J.}},
\oauthor{\bsnm{Pfahringer}, \binits{B.}},
\oauthor{\bsnm{Holmes}, \binits{G.}},
\oauthor{\bsnm{{\v{Z}}liobait{\.{e}}}, \binits{I.}}:
{CD}-{MOA}: Change detection framework for massive online analysis,
pp. 92--103.
Springer
(2013).
doi:\doiurl{10.1007/978-3-642-41398-8\_9}
\end{botherref}
\endbibitem

%%% 13
\bibitem{Krawczyk2017}
\begin{barticle}
\bauthor{\bsnm{Krawczyk}, \binits{B.}},
\bauthor{\bsnm{Minku}, \binits{L.L.}},
\bauthor{\bsnm{Gama}, \binits{J.}},
\bauthor{\bsnm{Stefanowski}, \binits{J.}},
\bauthor{\bsnm{Wo{\'{z}}niak}, \binits{M.}}:
\batitle{Ensemble learning for data stream analysis: A survey}.
\bjtitle{Information Fusion}
\bvolume{37},
\bfpage{132}--\blpage{156}
(\byear{2017}).
doi:\doiurl{10.1016/j.inffus.2017.02.004}
\end{barticle}
\endbibitem

%%% 14
\bibitem{Domingos2001}
\begin{bchapter}
\bauthor{\bsnm{Domingos}, \binits{P.M.}},
\bauthor{\bsnm{Hulten}, \binits{G.}}:
\bctitle{Catching up with the data: Research issues in mining data streams}.
In: \bbtitle{DMKD}
(\byear{2001})
\end{bchapter}
\endbibitem

%%% 15
\bibitem{montiel2018scikit}
\begin{barticle}
\bauthor{\bsnm{Montiel}, \binits{J.}},
\bauthor{\bsnm{Read}, \binits{J.}},
\bauthor{\bsnm{Bifet}, \binits{A.}},
\bauthor{\bsnm{Abdessalem}, \binits{T.}}:
\batitle{Scikit-multiflow: A multi-output streaming framework}.
\bjtitle{The Journal of Machine Learning Research}
\bvolume{19}(\bissue{1}),
\bfpage{2915}--\blpage{2914}
(\byear{2018})
\end{barticle}
\endbibitem

%%% 16
\bibitem{montiel2021river}
\begin{botherref}
\oauthor{\bsnm{Montiel}, \binits{J.}},
\oauthor{\bsnm{Halford}, \binits{M.}},
\oauthor{\bsnm{Mastelini}, \binits{S.M.}},
\oauthor{\bsnm{Bolmier}, \binits{G.}},
\oauthor{\bsnm{Sourty}, \binits{R.}},
\oauthor{\bsnm{Vaysse}, \binits{R.}},
\oauthor{\bsnm{Zouitine}, \binits{A.}},
\oauthor{\bsnm{Gomes}, \binits{H.M.}},
\oauthor{\bsnm{Read}, \binits{J.}},
\oauthor{\bsnm{Abdessalem}, \binits{T.}}, et al.:
River: machine learning for streaming data in python
(2021)
\end{botherref}
\endbibitem

%%% 17
\bibitem{webb2016characterizing}
\begin{barticle}
\bauthor{\bsnm{Webb}, \binits{G.I.}},
\bauthor{\bsnm{Hyde}, \binits{R.}},
\bauthor{\bsnm{Cao}, \binits{H.}},
\bauthor{\bsnm{Nguyen}, \binits{H.L.}},
\bauthor{\bsnm{Petitjean}, \binits{F.}}:
\batitle{Characterizing concept drift}.
\bjtitle{Data Mining and Knowledge Discovery}
\bvolume{30}(\bissue{4}),
\bfpage{964}--\blpage{994}
(\byear{2016})
\end{barticle}
\endbibitem

%%% 18
\bibitem{Bifet2015}
\begin{bchapter}
\bauthor{\bsnm{Bifet}, \binits{A.}},
\bauthor{\bparticle{de} \bsnm{Francisci~Morales}, \binits{G.}},
\bauthor{\bsnm{Read}, \binits{J.}},
\bauthor{\bsnm{Holmes}, \binits{G.}},
\bauthor{\bsnm{Pfahringer}, \binits{B.}}:
\bctitle{Efficient online evaluation of big data stream classifiers}.
In: \bbtitle{Proceedings of the 21th {ACM} {SIGKDD} International Conference on
  Knowledge Discovery and Data Mining}.
\bpublisher{{ACM}},
\blocation{New York, NY, USA}
(\byear{2015}).
doi:\doiurl{10.1145/2783258.2783372}
\end{bchapter}
\endbibitem

%%% 19
\bibitem{Goncalves2014}
\begin{barticle}
\bauthor{\bsnm{Gon{\c{c}}alves}, \binits{P.M.}},
\bauthor{\bparticle{de} \bsnm{Carvalho~Santos}, \binits{S.G.T.}},
\bauthor{\bsnm{Barros}, \binits{R.S.M.}},
\bauthor{\bsnm{Vieira}, \binits{D.C.L.}}:
\batitle{A comparative study on concept drift detectors}.
\bjtitle{Expert Systems with Applications}
\bvolume{41}(\bissue{18}),
\bfpage{8144}--\blpage{8156}
(\byear{2014}).
doi:\doiurl{10.1016/j.eswa.2014.07.019}
\end{barticle}
\endbibitem

%%% 20
\bibitem{RamirezGallego2017}
\begin{barticle}
\bauthor{\bsnm{Ram{\'{\i}}rez-Gallego}, \binits{S.}},
\bauthor{\bsnm{Krawczyk}, \binits{B.}},
\bauthor{\bsnm{Garc{\'{\i}}a}, \binits{S.}},
\bauthor{\bsnm{Wo{\'{z}}niak}, \binits{M.}},
\bauthor{\bsnm{Herrera}, \binits{F.}}:
\batitle{A survey on data preprocessing for data stream mining: Current status
  and future directions}.
\bjtitle{Neurocomputing}
\bvolume{239},
\bfpage{39}--\blpage{57}
(\byear{2017}).
doi:\doiurl{10.1016/j.neucom.2017.01.078}
\end{barticle}
\endbibitem

%%% 21
\bibitem{Gomes2017}
\begin{barticle}
\bauthor{\bsnm{Gomes}, \binits{H.M.}},
\bauthor{\bsnm{Barddal}, \binits{J.P.}},
\bauthor{\bsnm{Enembreck}, \binits{F.}},
\bauthor{\bsnm{Bifet}, \binits{A.}}:
\batitle{A survey on ensemble learning for data stream classification}.
\bjtitle{{ACM} Computing Surveys}
\bvolume{50}(\bissue{2}),
\bfpage{1}--\blpage{36}
(\byear{2017}).
doi:\doiurl{10.1145/3054925}
\end{barticle}
\endbibitem

%%% 22
\bibitem{Kolajo2019}
\begin{botherref}
\oauthor{\bsnm{Kolajo}, \binits{T.}},
\oauthor{\bsnm{Daramola}, \binits{O.}},
\oauthor{\bsnm{Adebiyi}, \binits{A.}}:
Big data stream analysis: a systematic literature review.
Journal of Big Data
\textbf{6}(1)
(2019).
doi:\doiurl{10.1186/s40537-019-0210-7}
\end{botherref}
\endbibitem

%%% 23
\bibitem{haug2022dynamic}
\begin{botherref}
\oauthor{\bsnm{Haug}, \binits{J.}},
\oauthor{\bsnm{Broelemann}, \binits{K.}},
\oauthor{\bsnm{Kasneci}, \binits{G.}}:
Dynamic model tree for interpretable data stream learning.
arXiv preprint arXiv:2203.16181
(2022)
\end{botherref}
\endbibitem

%%% 24
\bibitem{nogueira2017stability}
\begin{barticle}
\bauthor{\bsnm{Nogueira}, \binits{S.}},
\bauthor{\bsnm{Sechidis}, \binits{K.}},
\bauthor{\bsnm{Brown}, \binits{G.}}:
\batitle{On the stability of feature selection algorithms.}
\bjtitle{J. Mach. Learn. Res.}
\bvolume{18}(\bissue{1}),
\bfpage{6345}--\blpage{6398}
(\byear{2017})
\end{barticle}
\endbibitem

%%% 25
\bibitem{Haug2020}
\begin{bchapter}
\bauthor{\bsnm{Haug}, \binits{J.}},
\bauthor{\bsnm{Pawelczyk}, \binits{M.}},
\bauthor{\bsnm{Broelemann}, \binits{K.}},
\bauthor{\bsnm{Kasneci}, \binits{G.}}:
\bctitle{Leveraging model inherent variable importance for stable online
  feature selection}.
In: \bbtitle{Proceedings of the 26th {ACM} {SIGKDD} International Conference on
  Knowledge Discovery {\&} Data Mining}.
\bpublisher{{ACM}},
\blocation{New York, NY, USA}
(\byear{2020}).
doi:\doiurl{10.1145/3394486.3403200}
\end{bchapter}
\endbibitem

%%% 26
\bibitem{Brzezinski2017}
\begin{barticle}
\bauthor{\bsnm{Brzezinski}, \binits{D.}},
\bauthor{\bsnm{Stefanowski}, \binits{J.}}:
\batitle{Prequential {AUC}: properties of the area under the {ROC} curve for
  data streams with concept drift}.
\bjtitle{Knowledge and Information Systems}
\bvolume{52}(\bissue{2}),
\bfpage{531}--\blpage{562}
(\byear{2017}).
doi:\doiurl{10.1007/s10115-017-1022-8}
\end{barticle}
\endbibitem

%%% 27
\bibitem{Haug2021}
\begin{bchapter}
\bauthor{\bsnm{Haug}, \binits{J.}},
\bauthor{\bsnm{Kasneci}, \binits{G.}}:
\bctitle{Learning parameter distributions to detect concept drift in data
  streams}.
In: \bbtitle{2020 25th International Conference on Pattern Recognition
  ({ICPR})}.
\bpublisher{{IEEE}},
\blocation{USA}
(\byear{2021}).
doi:\doiurl{10.1109/icpr48806.2021.9412499}
\end{bchapter}
\endbibitem

%%% 28
\bibitem{miller2019}
\begin{barticle}
\bauthor{\bsnm{Miller}, \binits{T.}}:
\batitle{Explanation in artificial intelligence: Insights from the social
  sciences}.
\bjtitle{Artificial Intelligence}
\bvolume{267},
\bfpage{1}--\blpage{38}
(\byear{2019}).
doi:\doiurl{10.1016/j.artint.2018.07.007}
\end{barticle}
\endbibitem

%%% 29
\bibitem{Guidotti2019}
\begin{barticle}
\bauthor{\bsnm{Guidotti}, \binits{R.}},
\bauthor{\bsnm{Monreale}, \binits{A.}},
\bauthor{\bsnm{Ruggieri}, \binits{S.}},
\bauthor{\bsnm{Turini}, \binits{F.}},
\bauthor{\bsnm{Giannotti}, \binits{F.}},
\bauthor{\bsnm{Pedreschi}, \binits{D.}}:
\batitle{A survey of methods for explaining black box models}.
\bjtitle{{ACM} Computing Surveys}
\bvolume{51}(\bissue{5}),
\bfpage{1}--\blpage{42}
(\byear{2019}).
doi:\doiurl{10.1145/3236009}
\end{barticle}
\endbibitem

%%% 30
\bibitem{ntoutsi2020bias}
\begin{barticle}
\bauthor{\bsnm{Ntoutsi}, \binits{E.}},
\bauthor{\bsnm{Fafalios}, \binits{P.}},
\bauthor{\bsnm{Gadiraju}, \binits{U.}},
\bauthor{\bsnm{Iosifidis}, \binits{V.}},
\bauthor{\bsnm{Nejdl}, \binits{W.}},
\bauthor{\bsnm{Vidal}, \binits{M.-E.}},
\bauthor{\bsnm{Ruggieri}, \binits{S.}},
\bauthor{\bsnm{Turini}, \binits{F.}},
\bauthor{\bsnm{Papadopoulos}, \binits{S.}},
\bauthor{\bsnm{Krasanakis}, \binits{E.}}, \betal:
\batitle{Bias in data-driven artificial intelligence systems—an introductory
  survey}.
\bjtitle{Wiley Interdisciplinary Reviews: Data Mining and Knowledge Discovery}
\bvolume{10}(\bissue{3}),
\bfpage{1356}
(\byear{2020})
\end{barticle}
\endbibitem

%%% 31
\bibitem{pawelczyk2021carla}
\begin{botherref}
\oauthor{\bsnm{Pawelczyk}, \binits{M.}},
\oauthor{\bsnm{Bielawski}, \binits{S.}},
\oauthor{\bparticle{Van~den} \bsnm{Heuvel}, \binits{J.}},
\oauthor{\bsnm{Richter}, \binits{T.}},
\oauthor{\bsnm{Kasneci}, \binits{G.}}:
Carla: A python library to benchmark algorithmic recourse and counterfactual
  explanation algorithms
(2021)
\end{botherref}
\endbibitem

%%% 32
\bibitem{carvalho2019machine}
\begin{barticle}
\bauthor{\bsnm{Carvalho}, \binits{D.V.}},
\bauthor{\bsnm{Pereira}, \binits{E.M.}},
\bauthor{\bsnm{Cardoso}, \binits{J.S.}}:
\batitle{Machine learning interpretability: A survey on methods and metrics}.
\bjtitle{Electronics}
\bvolume{8}(\bissue{8}),
\bfpage{832}
(\byear{2019}).
doi:\doiurl{10.3390/electronics8080832}
\end{barticle}
\endbibitem

%%% 33
\bibitem{bibal2016interpretability}
\begin{bchapter}
\bauthor{\bsnm{Bibal}, \binits{A.}},
\bauthor{\bsnm{Fr{\'e}nay}, \binits{B.}}:
\bctitle{Interpretability of machine learning models and representations: an
  introduction}.
In: \bbtitle{ESANN}
(\byear{2016})
\end{bchapter}
\endbibitem

%%% 34
\bibitem{moshkovitz2021connecting}
\begin{bchapter}
\bauthor{\bsnm{Moshkovitz}, \binits{M.}},
\bauthor{\bsnm{Yang}, \binits{Y.-Y.}},
\bauthor{\bsnm{Chaudhuri}, \binits{K.}}:
\bctitle{Connecting interpretability and robustness in decision trees through
  separation}.
In: \bbtitle{International Conference on Machine Learning},
pp. \bfpage{7839}--\blpage{7849}
(\byear{2021}).
\bcomment{PMLR}
\end{bchapter}
\endbibitem

%%% 35
\bibitem{du2019techniques}
\begin{barticle}
\bauthor{\bsnm{Du}, \binits{M.}},
\bauthor{\bsnm{Liu}, \binits{N.}},
\bauthor{\bsnm{Hu}, \binits{X.}}:
\batitle{Techniques for interpretable machine learning}.
\bjtitle{Communications of the {ACM}}
\bvolume{63}(\bissue{1}),
\bfpage{68}--\blpage{77}
(\byear{2019}).
doi:\doiurl{10.1145/3359786}
\end{barticle}
\endbibitem

%%% 36
\bibitem{ribeiro2016should}
\begin{bchapter}
\bauthor{\bsnm{Ribeiro}, \binits{M.T.}},
\bauthor{\bsnm{Singh}, \binits{S.}},
\bauthor{\bsnm{Guestrin}, \binits{C.}}:
\bctitle{"why should i trust you?" explaining the predictions of any
  classifier}.
In: \bbtitle{Proceedings of the 22nd ACM SIGKDD International Conference on
  Knowledge Discovery and Data Mining},
pp. \bfpage{1135}--\blpage{1144}
(\byear{2016})
\end{bchapter}
\endbibitem

%%% 37
\bibitem{lundberg2017unified}
\begin{bchapter}
\bauthor{\bsnm{Lundberg}, \binits{S.M.}},
\bauthor{\bsnm{Lee}, \binits{S.-I.}}:
\bctitle{A unified approach to interpreting model predictions}.
In: \bbtitle{Proceedings of the 31st International Conference on Neural
  Information Processing Systems},
pp. \bfpage{4768}--\blpage{4777}
(\byear{2017})
\end{bchapter}
\endbibitem

%%% 38
\bibitem{kasneci2016licon}
\begin{bchapter}
\bauthor{\bsnm{Kasneci}, \binits{G.}},
\bauthor{\bsnm{Gottron}, \binits{T.}}:
\bctitle{Licon: A linear weighting scheme for the contribution ofinput
  variables in deep artificial neural networks}.
In: \bbtitle{Proceedings of the 25th ACM International on Conference on
  Information and Knowledge Management},
pp. \bfpage{45}--\blpage{54}
(\byear{2016})
\end{bchapter}
\endbibitem

%%% 39
\bibitem{hooker2019benchmark}
\begin{bchapter}
\bauthor{\bsnm{Hooker}, \binits{S.}},
\bauthor{\bsnm{Erhan}, \binits{D.}},
\bauthor{\bsnm{Kindermans}, \binits{P.-J.}},
\bauthor{\bsnm{Kim}, \binits{B.}}:
\bctitle{A benchmark for interpretability methods in deep neural networks}.
In: \beditor{\bsnm{Wallach}, \binits{H.}},
\beditor{\bsnm{Larochelle}, \binits{H.}},
\beditor{\bsnm{Beygelzimer}, \binits{A.}},
\beditor{\bparticle{d\textquotesingle} \bsnm{Alch\'{e}-Buc}, \binits{F.}},
\beditor{\bsnm{Fox}, \binits{E.}},
\beditor{\bsnm{Garnett}, \binits{R.}} (eds.)
\bbtitle{Advances in Neural Information Processing Systems},
vol. \bseriesno{32}.
\bpublisher{Curran Associates, Inc.},
\blocation{Red Hook, NY, USA}
(\byear{2019})
\end{bchapter}
\endbibitem

%%% 40
\bibitem{Haug2021a}
\begin{botherref}
\oauthor{\bsnm{Haug}, \binits{J.}},
\oauthor{\bsnm{Zürn}, \binits{S.}},
\oauthor{\bsnm{El-Jiz}, \binits{P.}},
\oauthor{\bsnm{Kasneci}, \binits{G.}}:
On baselines for local feature attributions.
AAAI 2021 Workshop on Explainable Agency in Artificial Intelligence
(2021).
\arxivurl{2101.00905}
\end{botherref}
\endbibitem

%%% 41
\bibitem{bommert2021stabm}
\begin{barticle}
\bauthor{\bsnm{Bommert}, \binits{A.}},
\bauthor{\bsnm{Lang}, \binits{M.}}:
\batitle{stabm: Stability measures for feature selection}.
\bjtitle{Journal of Open Source Software}
\bvolume{6}(\bissue{59}),
\bfpage{3010}
(\byear{2021})
\end{barticle}
\endbibitem

%%% 42
\bibitem{kalousis2005stability}
\begin{bchapter}
\bauthor{\bsnm{Kalousis}, \binits{A.}},
\bauthor{\bsnm{Prados}, \binits{J.}},
\bauthor{\bsnm{Hilario}, \binits{M.}}:
\bctitle{Stability of feature selection algorithms}.
In: \bbtitle{Fifth {IEEE} International Conference on Data Mining
  ({ICDM}{\textquotesingle}05)}.
\bpublisher{{IEEE}},
\blocation{USA}
(\byear{2005}).
doi:\doiurl{10.1109/icdm.2005.135}
\end{bchapter}
\endbibitem

%%% 43
\bibitem{masud2010classification}
\begin{bchapter}
\bauthor{\bsnm{Masud}, \binits{M.M.}},
\bauthor{\bsnm{Chen}, \binits{Q.}},
\bauthor{\bsnm{Gao}, \binits{J.}},
\bauthor{\bsnm{Khan}, \binits{L.}},
\bauthor{\bsnm{Han}, \binits{J.}},
\bauthor{\bsnm{Thuraisingham}, \binits{B.}}:
\bctitle{Classification and novel class detection of data streams in a dynamic
  feature space}.
In: \bbtitle{Machine Learning and Knowledge Discovery in Databases},
pp. \bfpage{337}--\blpage{352}.
\bpublisher{Springer}, \blocation{???}
(\byear{2010}).
doi:\doiurl{10.1007/978-3-642-15883-4\_22}
\end{bchapter}
\endbibitem

%%% 44
\bibitem{sethi2017reliable}
\begin{barticle}
\bauthor{\bsnm{Sethi}, \binits{T.S.}},
\bauthor{\bsnm{Kantardzic}, \binits{M.}}:
\batitle{On the reliable detection of concept drift from streaming unlabeled
  data}.
\bjtitle{Expert Systems with Applications}
\bvolume{82},
\bfpage{77}--\blpage{99}
(\byear{2017})
\end{barticle}
\endbibitem

%%% 45
\bibitem{manapragada2020eager}
\begin{botherref}
\oauthor{\bsnm{Manapragada}, \binits{C.}},
\oauthor{\bsnm{Gomes}, \binits{H.M.}},
\oauthor{\bsnm{Salehi}, \binits{M.}},
\oauthor{\bsnm{Bifet}, \binits{A.}},
\oauthor{\bsnm{Webb}, \binits{G.I.}}:
An eager splitting strategy for online decision trees.
arXiv preprint arXiv:2010.10935
(2020)
\end{botherref}
\endbibitem

%%% 46
\bibitem{souza2020challenges}
\begin{barticle}
\bauthor{\bsnm{Souza}, \binits{V.M.}},
\bauthor{\bparticle{dos} \bsnm{Reis}, \binits{D.M.}},
\bauthor{\bsnm{Maletzke}, \binits{A.G.}},
\bauthor{\bsnm{Batista}, \binits{G.E.}}:
\batitle{Challenges in benchmarking stream learning algorithms with real-world
  data}.
\bjtitle{Data Mining and Knowledge Discovery}
\bvolume{34}(\bissue{6}),
\bfpage{1805}--\blpage{1858}
(\byear{2020})
\end{barticle}
\endbibitem

%%% 47
\bibitem{kasneci2021tueyeq}
\begin{barticle}
\bauthor{\bsnm{Kasneci}, \binits{E.}},
\bauthor{\bsnm{Kasneci}, \binits{G.}},
\bauthor{\bsnm{Appel}, \binits{T.}},
\bauthor{\bsnm{Haug}, \binits{J.}},
\bauthor{\bsnm{Wortha}, \binits{F.}},
\bauthor{\bsnm{Tibus}, \binits{M.}},
\bauthor{\bsnm{Trautwein}, \binits{U.}},
\bauthor{\bsnm{Gerjets}, \binits{P.}}:
\batitle{T{\"u}eyeq, a rich iq test performance data set with eye movement,
  educational and socio-demographic information}.
\bjtitle{Scientific Data}
\bvolume{8}(\bissue{1}),
\bfpage{1}--\blpage{14}
(\byear{2021})
\end{barticle}
\endbibitem

%%% 48
\bibitem{kasneci2020tueeyeq}
\begin{bchapter}
\bauthor{\bsnm{Kasneci}, \binits{E.}},
\bauthor{\bsnm{Kasneci}, \binits{G.}},
\bauthor{\bsnm{Appel}, \binits{T.}},
\bauthor{\bsnm{Haug}, \binits{J.}},
\bauthor{\bsnm{Wortha}, \binits{F.}},
\bauthor{\bsnm{Tibus}, \binits{M.}},
\bauthor{\bsnm{Trautwein}, \binits{U.}},
\bauthor{\bsnm{Gerjets}, \binits{P.}}:
\bctitle{{TüEyeQ, a rich IQ test performance data set with eye movement,
  educational and socio-demographic information}}.
\bpublisher{Harvard Dataverse},
\blocation{USA}
(\byear{2020}).
doi:\doiurl{10.7910/DVN/JGOCKI}
\end{bchapter}
\endbibitem

%%% 49
\bibitem{bifet2009adaptive}
\begin{bchapter}
\bauthor{\bsnm{Bifet}, \binits{A.}},
\bauthor{\bsnm{Gavalda}, \binits{R.}}:
\bctitle{Adaptive learning from evolving data streams}.
In: \bbtitle{International Symposium on Intelligent Data Analysis},
pp. \bfpage{249}--\blpage{260}
(\byear{2009}).
\bcomment{Springer}
\end{bchapter}
\endbibitem

%%% 50
\bibitem{bifet2007learning}
\begin{bchapter}
\bauthor{\bsnm{Bifet}, \binits{A.}},
\bauthor{\bsnm{Gavalda}, \binits{R.}}:
\bctitle{Learning from time-changing data with adaptive windowing}.
In: \bbtitle{Proceedings of the 2007 SIAM International Conference on Data
  Mining},
pp. \bfpage{443}--\blpage{448}
(\byear{2007}).
\bcomment{SIAM}
\end{bchapter}
\endbibitem

%%% 51
\bibitem{page1954continuous}
\begin{barticle}
\bauthor{\bsnm{Page}, \binits{E.S.}}:
\batitle{Continuous inspection schemes}.
\bjtitle{Biometrika}
\bvolume{41}(\bissue{1/2}),
\bfpage{100}--\blpage{115}
(\byear{1954})
\end{barticle}
\endbibitem

%%% 52
\bibitem{wang2013online}
\begin{barticle}
\bauthor{\bsnm{Wang}, \binits{J.}},
\bauthor{\bsnm{Zhao}, \binits{P.}},
\bauthor{\bsnm{Hoi}, \binits{S.C.}},
\bauthor{\bsnm{Jin}, \binits{R.}}:
\batitle{Online feature selection and its applications}.
\bjtitle{IEEE Transactions on knowledge and data engineering}
\bvolume{26}(\bissue{3}),
\bfpage{698}--\blpage{710}
(\byear{2013})
\end{barticle}
\endbibitem

\end{thebibliography}

\newcommand{\BMCxmlcomment}[1]{}

\BMCxmlcomment{

<refgrp>

<bibl id="B1">
  <title><p>Learning under Concept Drift: A Review</p></title>
  <aug>
    <au><snm>Lu</snm><fnm>J</fnm></au>
    <au><snm>Liu</snm><fnm>A</fnm></au>
    <au><snm>Dong</snm><fnm>F</fnm></au>
    <au><snm>Gu</snm><fnm>F</fnm></au>
    <au><snm>Gama</snm><fnm>J</fnm></au>
    <au><snm>Zhang</snm><fnm>G</fnm></au>
  </aug>
  <source>IEEE Transactions on Knowledge and Data Engineering 31, no. 12
  (2018): 2346-2363</source>
  <pubdate>2018</pubdate>
</bibl>

<bibl id="B2">
  <title><p>Learning in nonstationary environments: A survey</p></title>
  <aug>
    <au><snm>Ditzler</snm><fnm>G</fnm></au>
    <au><snm>Roveri</snm><fnm>M</fnm></au>
    <au><snm>Alippi</snm><fnm>C</fnm></au>
    <au><snm>Polikar</snm><fnm>R</fnm></au>
  </aug>
  <source>IEEE Computational Intelligence Magazine</source>
  <publisher>IEEE</publisher>
  <pubdate>2015</pubdate>
  <volume>10</volume>
  <issue>4</issue>
  <fpage>12</fpage>
  <lpage>-25</lpage>
</bibl>

<bibl id="B3">
  <title><p>A survey on learning from data streams: current and future
  trends</p></title>
  <aug>
    <au><snm>Gama</snm><fnm>J</fnm></au>
  </aug>
  <source>Progress in Artificial Intelligence</source>
  <publisher>Springer</publisher>
  <pubdate>2012</pubdate>
  <volume>1</volume>
  <issue>1</issue>
  <fpage>45</fpage>
  <lpage>-55</lpage>
</bibl>

<bibl id="B4">
  <title><p>Mining high-speed data streams</p></title>
  <aug>
    <au><snm>Domingos</snm><fnm>P</fnm></au>
    <au><snm>Hulten</snm><fnm>G</fnm></au>
  </aug>
  <source>Proceedings of the sixth ACM SIGKDD international conference on
  Knowledge discovery and data mining</source>
  <pubdate>2000</pubdate>
  <fpage>71</fpage>
  <lpage>-80</lpage>
</bibl>

<bibl id="B5">
  <title><p>Stress-testing hoeffding trees</p></title>
  <aug>
    <au><snm>Holmes</snm><fnm>G</fnm></au>
    <au><snm>Kirkby</snm><fnm>R</fnm></au>
    <au><snm>Pfahringer</snm><fnm>B</fnm></au>
  </aug>
  <source>European conference on principles of data mining and knowledge
  discovery</source>
  <pubdate>2005</pubdate>
  <fpage>495</fpage>
  <lpage>-502</lpage>
</bibl>

<bibl id="B6">
  <title><p>Fast perceptron decision tree learning from evolving data
  streams</p></title>
  <aug>
    <au><snm>Bifet</snm><fnm>A</fnm></au>
    <au><snm>Holmes</snm><fnm>G</fnm></au>
    <au><snm>Pfahringer</snm><fnm>B</fnm></au>
    <au><snm>Frank</snm><fnm>E</fnm></au>
  </aug>
  <source>Pacific-Asia conference on knowledge discovery and data
  mining</source>
  <pubdate>2010</pubdate>
  <fpage>299</fpage>
  <lpage>-310</lpage>
</bibl>

<bibl id="B7">
  <title><p>Batch-incremental versus instance-incremental learning in dynamic
  and evolving data</p></title>
  <aug>
    <au><snm>Read</snm><fnm>J</fnm></au>
    <au><snm>Bifet</snm><fnm>A</fnm></au>
    <au><snm>Pfahringer</snm><fnm>B</fnm></au>
    <au><snm>Holmes</snm><fnm>G</fnm></au>
  </aug>
  <source>International symposium on intelligent data analysis</source>
  <pubdate>2012</pubdate>
  <fpage>313</fpage>
  <lpage>-323</lpage>
</bibl>

<bibl id="B8">
  <title><p>Learning regularized hoeffding trees from data streams</p></title>
  <aug>
    <au><snm>Barddal</snm><fnm>JP</fnm></au>
    <au><snm>Enembreck</snm><fnm>F</fnm></au>
  </aug>
  <source>Proceedings of the 34th ACM/SIGAPP Symposium on Applied
  Computing</source>
  <pubdate>2019</pubdate>
  <fpage>574</fpage>
  <lpage>-581</lpage>
</bibl>

<bibl id="B9">
  <title><p>Regularized and incremental decision trees for data
  streams</p></title>
  <aug>
    <au><snm>Barddal</snm><fnm>JP</fnm></au>
    <au><snm>Enembreck</snm><fnm>F</fnm></au>
  </aug>
  <source>Annals of Telecommunications</source>
  <publisher>Springer</publisher>
  <pubdate>2020</pubdate>
  <volume>75</volume>
  <issue>9</issue>
  <fpage>493</fpage>
  <lpage>-503</lpage>
</bibl>

<bibl id="B10">
  <title><p>Issues in Evaluation of Stream Learning Algorithms</p></title>
  <aug>
    <au><snm>Gama</snm><fnm>Ja</fnm></au>
    <au><snm>Sebasti\ {a}o</snm><fnm>R</fnm></au>
    <au><snm>Rodrigues</snm><fnm>PP</fnm></au>
  </aug>
  <publisher>New York, NY, USA: Association for Computing Machinery</publisher>
  <series><title><p>KDD '09</p></title></series>
  <pubdate>2009</pubdate>
  <fpage>329–338</fpage>
  <url>https://doi.org/10.1145/1557019.1557060</url>
</bibl>

<bibl id="B11">
  <title><p>On evaluating stream learning algorithms</p></title>
  <aug>
    <au><snm>Gama</snm><fnm>J</fnm></au>
    <au><snm>Sebasti{\~{a}}o</snm><fnm>R</fnm></au>
    <au><snm>Rodrigues</snm><fnm>PP</fnm></au>
  </aug>
  <source>Machine Learning</source>
  <publisher>Springer Science and Business Media {LLC}</publisher>
  <pubdate>2012</pubdate>
  <volume>90</volume>
  <issue>3</issue>
  <fpage>317</fpage>
  <lpage>-346</lpage>
</bibl>

<bibl id="B12">
  <title><p>{CD}-{MOA}: Change Detection Framework for Massive Online
  Analysis</p></title>
  <aug>
    <au><snm>Bifet</snm><fnm>A</fnm></au>
    <au><snm>Read</snm><fnm>J</fnm></au>
    <au><snm>Pfahringer</snm><fnm>B</fnm></au>
    <au><snm>Holmes</snm><fnm>G</fnm></au>
    <au><snm>{\v{Z}}liobait{\.{e}}</snm><fnm>I</fnm></au>
  </aug>
  <publisher>Springer Berlin Heidelberg</publisher>
  <pubdate>2013</pubdate>
  <fpage>92</fpage>
  <lpage>-103</lpage>
</bibl>

<bibl id="B13">
  <title><p>Ensemble learning for data stream analysis: A survey</p></title>
  <aug>
    <au><snm>Krawczyk</snm><fnm>B</fnm></au>
    <au><snm>Minku</snm><fnm>LL</fnm></au>
    <au><snm>Gama</snm><fnm>J</fnm></au>
    <au><snm>Stefanowski</snm><fnm>J</fnm></au>
    <au><snm>Wo{\'{z}}niak</snm><fnm>M</fnm></au>
  </aug>
  <source>Information Fusion</source>
  <publisher>Elsevier {BV}</publisher>
  <pubdate>2017</pubdate>
  <volume>37</volume>
  <fpage>132</fpage>
  <lpage>-156</lpage>
</bibl>

<bibl id="B14">
  <title><p>Catching up with the Data: Research Issues in Mining Data
  Streams</p></title>
  <aug>
    <au><snm>Domingos</snm><fnm>PM</fnm></au>
    <au><snm>Hulten</snm><fnm>G</fnm></au>
  </aug>
  <source>DMKD</source>
  <pubdate>2001</pubdate>
</bibl>

<bibl id="B15">
  <title><p>Scikit-multiflow: A multi-output streaming framework</p></title>
  <aug>
    <au><snm>Montiel</snm><fnm>J</fnm></au>
    <au><snm>Read</snm><fnm>J</fnm></au>
    <au><snm>Bifet</snm><fnm>A</fnm></au>
    <au><snm>Abdessalem</snm><fnm>T</fnm></au>
  </aug>
  <source>The Journal of Machine Learning Research</source>
  <publisher>JMLR. org</publisher>
  <pubdate>2018</pubdate>
  <volume>19</volume>
  <issue>1</issue>
  <fpage>2915</fpage>
  <lpage>-2914</lpage>
</bibl>

<bibl id="B16">
  <title><p>River: machine learning for streaming data in Python</p></title>
  <aug>
    <au><snm>Montiel</snm><fnm>J</fnm></au>
    <au><snm>Halford</snm><fnm>M</fnm></au>
    <au><snm>Mastelini</snm><fnm>SM</fnm></au>
    <au><snm>Bolmier</snm><fnm>G</fnm></au>
    <au><snm>Sourty</snm><fnm>R</fnm></au>
    <au><snm>Vaysse</snm><fnm>R</fnm></au>
    <au><snm>Zouitine</snm><fnm>A</fnm></au>
    <au><snm>Gomes</snm><fnm>HM</fnm></au>
    <au><snm>Read</snm><fnm>J</fnm></au>
    <au><snm>Abdessalem</snm><fnm>T</fnm></au>
    <au><cnm>others</cnm></au>
  </aug>
  <pubdate>2021</pubdate>
</bibl>

<bibl id="B17">
  <title><p>Characterizing concept drift</p></title>
  <aug>
    <au><snm>Webb</snm><fnm>GI</fnm></au>
    <au><snm>Hyde</snm><fnm>R</fnm></au>
    <au><snm>Cao</snm><fnm>H</fnm></au>
    <au><snm>Nguyen</snm><fnm>HL</fnm></au>
    <au><snm>Petitjean</snm><fnm>F</fnm></au>
  </aug>
  <source>Data Mining and Knowledge Discovery</source>
  <publisher>Springer</publisher>
  <pubdate>2016</pubdate>
  <volume>30</volume>
  <issue>4</issue>
  <fpage>964</fpage>
  <lpage>-994</lpage>
</bibl>

<bibl id="B18">
  <title><p>Efficient Online Evaluation of Big Data Stream
  Classifiers</p></title>
  <aug>
    <au><snm>Bifet</snm><fnm>A</fnm></au>
    <au><snm>Francisci Morales</snm><fnm>G</fnm></au>
    <au><snm>Read</snm><fnm>J</fnm></au>
    <au><snm>Holmes</snm><fnm>G</fnm></au>
    <au><snm>Pfahringer</snm><fnm>B</fnm></au>
  </aug>
  <source>Proceedings of the 21th {ACM} {SIGKDD} International Conference on
  Knowledge Discovery and Data Mining</source>
  <publisher>New York, NY, USA: {ACM}</publisher>
  <pubdate>2015</pubdate>
</bibl>

<bibl id="B19">
  <title><p>A comparative study on concept drift detectors</p></title>
  <aug>
    <au><snm>Gon{\c{c}}alves</snm><fnm>PM</fnm></au>
    <au><snm>Carvalho Santos</snm><fnm>SG</fnm></au>
    <au><snm>Barros</snm><fnm>RS</fnm></au>
    <au><snm>Vieira</snm><fnm>DC</fnm></au>
  </aug>
  <source>Expert Systems with Applications</source>
  <publisher>Elsevier {BV}</publisher>
  <pubdate>2014</pubdate>
  <volume>41</volume>
  <issue>18</issue>
  <fpage>8144</fpage>
  <lpage>-8156</lpage>
</bibl>

<bibl id="B20">
  <title><p>A survey on data preprocessing for data stream mining: Current
  status and future directions</p></title>
  <aug>
    <au><snm>Ram{\'{\i}}rez Gallego</snm><fnm>S</fnm></au>
    <au><snm>Krawczyk</snm><fnm>B</fnm></au>
    <au><snm>Garc{\'{\i}}a</snm><fnm>S</fnm></au>
    <au><snm>Wo{\'{z}}niak</snm><fnm>M</fnm></au>
    <au><snm>Herrera</snm><fnm>F</fnm></au>
  </aug>
  <source>Neurocomputing</source>
  <publisher>Elsevier {BV}</publisher>
  <pubdate>2017</pubdate>
  <volume>239</volume>
  <fpage>39</fpage>
  <lpage>-57</lpage>
</bibl>

<bibl id="B21">
  <title><p>A Survey on Ensemble Learning for Data Stream
  Classification</p></title>
  <aug>
    <au><snm>Gomes</snm><fnm>HM</fnm></au>
    <au><snm>Barddal</snm><fnm>JP</fnm></au>
    <au><snm>Enembreck</snm><fnm>F</fnm></au>
    <au><snm>Bifet</snm><fnm>A</fnm></au>
  </aug>
  <source>{ACM} Computing Surveys</source>
  <publisher>Association for Computing Machinery ({ACM})</publisher>
  <pubdate>2017</pubdate>
  <volume>50</volume>
  <issue>2</issue>
  <fpage>1</fpage>
  <lpage>-36</lpage>
</bibl>

<bibl id="B22">
  <title><p>Big data stream analysis: a systematic literature
  review</p></title>
  <aug>
    <au><snm>Kolajo</snm><fnm>T</fnm></au>
    <au><snm>Daramola</snm><fnm>O</fnm></au>
    <au><snm>Adebiyi</snm><fnm>A</fnm></au>
  </aug>
  <source>Journal of Big Data</source>
  <publisher>Springer Science and Business Media {LLC}</publisher>
  <pubdate>2019</pubdate>
  <volume>6</volume>
  <issue>1</issue>
</bibl>

<bibl id="B23">
  <title><p>Dynamic Model Tree for Interpretable Data Stream
  Learning</p></title>
  <aug>
    <au><snm>Haug</snm><fnm>J</fnm></au>
    <au><snm>Broelemann</snm><fnm>K</fnm></au>
    <au><snm>Kasneci</snm><fnm>G</fnm></au>
  </aug>
  <source>arXiv preprint arXiv:2203.16181</source>
  <pubdate>2022</pubdate>
</bibl>

<bibl id="B24">
  <title><p>On the stability of feature selection algorithms.</p></title>
  <aug>
    <au><snm>Nogueira</snm><fnm>S</fnm></au>
    <au><snm>Sechidis</snm><fnm>K</fnm></au>
    <au><snm>Brown</snm><fnm>G</fnm></au>
  </aug>
  <source>J. Mach. Learn. Res.</source>
  <pubdate>2017</pubdate>
  <volume>18</volume>
  <issue>1</issue>
  <fpage>6345</fpage>
  <lpage>-6398</lpage>
</bibl>

<bibl id="B25">
  <title><p>Leveraging Model Inherent Variable Importance for Stable Online
  Feature Selection</p></title>
  <aug>
    <au><snm>Haug</snm><fnm>J</fnm></au>
    <au><snm>Pawelczyk</snm><fnm>M</fnm></au>
    <au><snm>Broelemann</snm><fnm>K</fnm></au>
    <au><snm>Kasneci</snm><fnm>G</fnm></au>
  </aug>
  <source>Proceedings of the 26th {ACM} {SIGKDD} International Conference on
  Knowledge Discovery {\&} Data Mining</source>
  <publisher>New York, NY, USA: {ACM}</publisher>
  <pubdate>2020</pubdate>
</bibl>

<bibl id="B26">
  <title><p>Prequential {AUC}: properties of the area under the {ROC} curve for
  data streams with concept drift</p></title>
  <aug>
    <au><snm>Brzezinski</snm><fnm>D</fnm></au>
    <au><snm>Stefanowski</snm><fnm>J</fnm></au>
  </aug>
  <source>Knowledge and Information Systems</source>
  <publisher>Springer Science and Business Media {LLC}</publisher>
  <pubdate>2017</pubdate>
  <volume>52</volume>
  <issue>2</issue>
  <fpage>531</fpage>
  <lpage>-562</lpage>
</bibl>

<bibl id="B27">
  <title><p>Learning Parameter Distributions to Detect Concept Drift in Data
  Streams</p></title>
  <aug>
    <au><snm>Haug</snm><fnm>J</fnm></au>
    <au><snm>Kasneci</snm><fnm>G</fnm></au>
  </aug>
  <source>2020 25th International Conference on Pattern Recognition
  ({ICPR})</source>
  <publisher>USA: {IEEE}</publisher>
  <pubdate>2021</pubdate>
</bibl>

<bibl id="B28">
  <title><p>Explanation in artificial intelligence: Insights from the social
  sciences</p></title>
  <aug>
    <au><snm>Miller</snm><fnm>T</fnm></au>
  </aug>
  <source>Artificial Intelligence</source>
  <publisher>Elsevier {BV}</publisher>
  <pubdate>2019</pubdate>
  <volume>267</volume>
  <fpage>1</fpage>
  <lpage>-38</lpage>
</bibl>

<bibl id="B29">
  <title><p>A Survey of Methods for Explaining Black Box Models</p></title>
  <aug>
    <au><snm>Guidotti</snm><fnm>R</fnm></au>
    <au><snm>Monreale</snm><fnm>A</fnm></au>
    <au><snm>Ruggieri</snm><fnm>S</fnm></au>
    <au><snm>Turini</snm><fnm>F</fnm></au>
    <au><snm>Giannotti</snm><fnm>F</fnm></au>
    <au><snm>Pedreschi</snm><fnm>D</fnm></au>
  </aug>
  <source>{ACM} Computing Surveys</source>
  <publisher>Association for Computing Machinery ({ACM})</publisher>
  <pubdate>2019</pubdate>
  <volume>51</volume>
  <issue>5</issue>
  <fpage>1</fpage>
  <lpage>-42</lpage>
</bibl>

<bibl id="B30">
  <title><p>Bias in data-driven artificial intelligence systems—An
  introductory survey</p></title>
  <aug>
    <au><snm>Ntoutsi</snm><fnm>E</fnm></au>
    <au><snm>Fafalios</snm><fnm>P</fnm></au>
    <au><snm>Gadiraju</snm><fnm>U</fnm></au>
    <au><snm>Iosifidis</snm><fnm>V</fnm></au>
    <au><snm>Nejdl</snm><fnm>W</fnm></au>
    <au><snm>Vidal</snm><fnm>ME</fnm></au>
    <au><snm>Ruggieri</snm><fnm>S</fnm></au>
    <au><snm>Turini</snm><fnm>F</fnm></au>
    <au><snm>Papadopoulos</snm><fnm>S</fnm></au>
    <au><snm>Krasanakis</snm><fnm>E</fnm></au>
    <au><cnm>others</cnm></au>
  </aug>
  <source>Wiley Interdisciplinary Reviews: Data Mining and Knowledge
  Discovery</source>
  <publisher>Wiley Online Library</publisher>
  <pubdate>2020</pubdate>
  <volume>10</volume>
  <issue>3</issue>
  <fpage>e1356</fpage>
</bibl>

<bibl id="B31">
  <title><p>CARLA: A Python Library to Benchmark Algorithmic Recourse and
  Counterfactual Explanation Algorithms</p></title>
  <aug>
    <au><snm>Pawelczyk</snm><fnm>M</fnm></au>
    <au><snm>Bielawski</snm><fnm>S</fnm></au>
    <au><snm>Heuvel</snm><fnm>J</fnm></au>
    <au><snm>Richter</snm><fnm>T</fnm></au>
    <au><snm>Kasneci</snm><fnm>G</fnm></au>
  </aug>
  <pubdate>2021</pubdate>
</bibl>

<bibl id="B32">
  <title><p>Machine Learning Interpretability: A Survey on Methods and
  Metrics</p></title>
  <aug>
    <au><snm>Carvalho</snm><fnm>DV</fnm></au>
    <au><snm>Pereira</snm><fnm>EM</fnm></au>
    <au><snm>Cardoso</snm><fnm>JS</fnm></au>
  </aug>
  <source>Electronics</source>
  <publisher>{MDPI} {AG}</publisher>
  <pubdate>2019</pubdate>
  <volume>8</volume>
  <issue>8</issue>
  <fpage>832</fpage>
</bibl>

<bibl id="B33">
  <title><p>Interpretability of Machine Learning Models and Representations: an
  Introduction</p></title>
  <aug>
    <au><snm>Bibal</snm><fnm>A</fnm></au>
    <au><snm>Fr{\'e}nay</snm><fnm>B</fnm></au>
  </aug>
  <source>ESANN</source>
  <pubdate>2016</pubdate>
</bibl>

<bibl id="B34">
  <title><p>Connecting Interpretability and Robustness in Decision Trees
  Through Separation</p></title>
  <aug>
    <au><snm>Moshkovitz</snm><fnm>M</fnm></au>
    <au><snm>Yang</snm><fnm>YY</fnm></au>
    <au><snm>Chaudhuri</snm><fnm>K</fnm></au>
  </aug>
  <source>International Conference on Machine Learning</source>
  <pubdate>2021</pubdate>
  <fpage>7839</fpage>
  <lpage>-7849</lpage>
</bibl>

<bibl id="B35">
  <title><p>Techniques for Interpretable Machine Learning</p></title>
  <aug>
    <au><snm>Du</snm><fnm>M</fnm></au>
    <au><snm>Liu</snm><fnm>N</fnm></au>
    <au><snm>Hu</snm><fnm>X</fnm></au>
  </aug>
  <source>Communications of the {ACM}</source>
  <publisher>Association for Computing Machinery ({ACM})</publisher>
  <pubdate>2019</pubdate>
  <volume>63</volume>
  <issue>1</issue>
  <fpage>68</fpage>
  <lpage>-77</lpage>
</bibl>

<bibl id="B36">
  <title><p>"Why should i trust you?" Explaining the predictions of any
  classifier</p></title>
  <aug>
    <au><snm>Ribeiro</snm><fnm>MT</fnm></au>
    <au><snm>Singh</snm><fnm>S</fnm></au>
    <au><snm>Guestrin</snm><fnm>C</fnm></au>
  </aug>
  <source>Proceedings of the 22nd ACM SIGKDD international conference on
  knowledge discovery and data mining</source>
  <pubdate>2016</pubdate>
  <fpage>1135</fpage>
  <lpage>-1144</lpage>
</bibl>

<bibl id="B37">
  <title><p>A unified approach to interpreting model predictions</p></title>
  <aug>
    <au><snm>Lundberg</snm><fnm>SM</fnm></au>
    <au><snm>Lee</snm><fnm>SI</fnm></au>
  </aug>
  <source>Proceedings of the 31st international conference on neural
  information processing systems</source>
  <pubdate>2017</pubdate>
  <fpage>4768</fpage>
  <lpage>-4777</lpage>
</bibl>

<bibl id="B38">
  <title><p>Licon: A linear weighting scheme for the contribution ofinput
  variables in deep artificial neural networks</p></title>
  <aug>
    <au><snm>Kasneci</snm><fnm>G</fnm></au>
    <au><snm>Gottron</snm><fnm>T</fnm></au>
  </aug>
  <source>Proceedings of the 25th ACM International on Conference on
  Information and Knowledge Management</source>
  <pubdate>2016</pubdate>
  <fpage>45</fpage>
  <lpage>-54</lpage>
</bibl>

<bibl id="B39">
  <title><p>A Benchmark for Interpretability Methods in Deep Neural
  Networks</p></title>
  <aug>
    <au><snm>Hooker</snm><fnm>S</fnm></au>
    <au><snm>Erhan</snm><fnm>D</fnm></au>
    <au><snm>Kindermans</snm><fnm>PJ</fnm></au>
    <au><snm>Kim</snm><fnm>B</fnm></au>
  </aug>
  <source>Advances in Neural Information Processing Systems</source>
  <publisher>Red Hook, NY, USA: Curran Associates, Inc.</publisher>
  <editor>H. Wallach and H. Larochelle and A. Beygelzimer and F.
  d\textquotesingle Alch\'{e}-Buc and E. Fox and R. Garnett</editor>
  <pubdate>2019</pubdate>
  <volume>32</volume>
</bibl>

<bibl id="B40">
  <title><p>On Baselines for Local Feature Attributions</p></title>
  <aug>
    <au><snm>Haug</snm><fnm>J</fnm></au>
    <au><snm>Zürn</snm><fnm>S</fnm></au>
    <au><snm>El Jiz</snm><fnm>P</fnm></au>
    <au><snm>Kasneci</snm><fnm>G</fnm></au>
  </aug>
  <source>AAAI 2021 Workshop on Explainable Agency in Artificial
  Intelligence</source>
  <pubdate>2021</pubdate>
</bibl>

<bibl id="B41">
  <title><p>stabm: Stability Measures for Feature Selection</p></title>
  <aug>
    <au><snm>Bommert</snm><fnm>A</fnm></au>
    <au><snm>Lang</snm><fnm>M</fnm></au>
  </aug>
  <source>Journal of Open Source Software</source>
  <pubdate>2021</pubdate>
  <volume>6</volume>
  <issue>59</issue>
  <fpage>3010</fpage>
</bibl>

<bibl id="B42">
  <title><p>Stability of Feature Selection Algorithms</p></title>
  <aug>
    <au><snm>Kalousis</snm><fnm>A.</fnm></au>
    <au><snm>Prados</snm><fnm>J.</fnm></au>
    <au><snm>Hilario</snm><fnm>M.</fnm></au>
  </aug>
  <source>Fifth {IEEE} International Conference on Data Mining
  ({ICDM}{\textquotesingle}05)</source>
  <publisher>USA: {IEEE}</publisher>
  <pubdate>2005</pubdate>
</bibl>

<bibl id="B43">
  <title><p>Classification and Novel Class Detection of Data Streams in a
  Dynamic Feature Space</p></title>
  <aug>
    <au><snm>Masud</snm><fnm>MM</fnm></au>
    <au><snm>Chen</snm><fnm>Q</fnm></au>
    <au><snm>Gao</snm><fnm>J</fnm></au>
    <au><snm>Khan</snm><fnm>L</fnm></au>
    <au><snm>Han</snm><fnm>J</fnm></au>
    <au><snm>Thuraisingham</snm><fnm>B</fnm></au>
  </aug>
  <source>Machine Learning and Knowledge Discovery in Databases</source>
  <publisher>Springer Berlin Heidelberg</publisher>
  <pubdate>2010</pubdate>
  <fpage>337</fpage>
  <lpage>-352</lpage>
</bibl>

<bibl id="B44">
  <title><p>On the reliable detection of concept drift from streaming unlabeled
  data</p></title>
  <aug>
    <au><snm>Sethi</snm><fnm>TS</fnm></au>
    <au><snm>Kantardzic</snm><fnm>M</fnm></au>
  </aug>
  <source>Expert Systems with Applications</source>
  <publisher>Elsevier</publisher>
  <pubdate>2017</pubdate>
  <volume>82</volume>
  <fpage>77</fpage>
  <lpage>-99</lpage>
</bibl>

<bibl id="B45">
  <title><p>An Eager Splitting Strategy for Online Decision Trees</p></title>
  <aug>
    <au><snm>Manapragada</snm><fnm>C</fnm></au>
    <au><snm>Gomes</snm><fnm>HM</fnm></au>
    <au><snm>Salehi</snm><fnm>M</fnm></au>
    <au><snm>Bifet</snm><fnm>A</fnm></au>
    <au><snm>Webb</snm><fnm>GI</fnm></au>
  </aug>
  <source>arXiv preprint arXiv:2010.10935</source>
  <pubdate>2020</pubdate>
</bibl>

<bibl id="B46">
  <title><p>Challenges in benchmarking stream learning algorithms with
  real-world data</p></title>
  <aug>
    <au><snm>Souza</snm><fnm>VM</fnm></au>
    <au><snm>Reis</snm><fnm>DM</fnm></au>
    <au><snm>Maletzke</snm><fnm>AG</fnm></au>
    <au><snm>Batista</snm><fnm>GE</fnm></au>
  </aug>
  <source>Data Mining and Knowledge Discovery</source>
  <publisher>Springer</publisher>
  <pubdate>2020</pubdate>
  <volume>34</volume>
  <issue>6</issue>
  <fpage>1805</fpage>
  <lpage>-1858</lpage>
</bibl>

<bibl id="B47">
  <title><p>T{\"u}EyeQ, a rich IQ test performance data set with eye movement,
  educational and socio-demographic information</p></title>
  <aug>
    <au><snm>Kasneci</snm><fnm>E</fnm></au>
    <au><snm>Kasneci</snm><fnm>G</fnm></au>
    <au><snm>Appel</snm><fnm>T</fnm></au>
    <au><snm>Haug</snm><fnm>J</fnm></au>
    <au><snm>Wortha</snm><fnm>F</fnm></au>
    <au><snm>Tibus</snm><fnm>M</fnm></au>
    <au><snm>Trautwein</snm><fnm>U</fnm></au>
    <au><snm>Gerjets</snm><fnm>P</fnm></au>
  </aug>
  <source>Scientific Data</source>
  <publisher>Nature Publishing Group</publisher>
  <pubdate>2021</pubdate>
  <volume>8</volume>
  <issue>1</issue>
  <fpage>1</fpage>
  <lpage>-14</lpage>
</bibl>

<bibl id="B48">
  <title><p>{TüEyeQ, a rich IQ test performance data set with eye movement,
  educational and socio-demographic information}</p></title>
  <aug>
    <au><snm>Kasneci</snm><fnm>E</fnm></au>
    <au><snm>Kasneci</snm><fnm>G</fnm></au>
    <au><snm>Appel</snm><fnm>T</fnm></au>
    <au><snm>Haug</snm><fnm>J</fnm></au>
    <au><snm>Wortha</snm><fnm>F</fnm></au>
    <au><snm>Tibus</snm><fnm>M</fnm></au>
    <au><snm>Trautwein</snm><fnm>U</fnm></au>
    <au><snm>Gerjets</snm><fnm>P</fnm></au>
  </aug>
  <publisher>USA: Harvard Dataverse</publisher>
  <pubdate>2020</pubdate>
</bibl>

<bibl id="B49">
  <title><p>Adaptive learning from evolving data streams</p></title>
  <aug>
    <au><snm>Bifet</snm><fnm>A</fnm></au>
    <au><snm>Gavalda</snm><fnm>R</fnm></au>
  </aug>
  <source>International Symposium on Intelligent Data Analysis</source>
  <pubdate>2009</pubdate>
  <fpage>249</fpage>
  <lpage>-260</lpage>
</bibl>

<bibl id="B50">
  <title><p>Learning from time-changing data with adaptive
  windowing</p></title>
  <aug>
    <au><snm>Bifet</snm><fnm>A</fnm></au>
    <au><snm>Gavalda</snm><fnm>R</fnm></au>
  </aug>
  <source>Proceedings of the 2007 SIAM international conference on data
  mining</source>
  <pubdate>2007</pubdate>
  <fpage>443</fpage>
  <lpage>-448</lpage>
</bibl>

<bibl id="B51">
  <title><p>Continuous Inspection Schemes</p></title>
  <aug>
    <au><snm>Page</snm><fnm>E. S.</fnm></au>
  </aug>
  <source>Biometrika</source>
  <publisher>[Oxford University Press, Biometrika Trust]</publisher>
  <pubdate>1954</pubdate>
  <volume>41</volume>
  <issue>1/2</issue>
  <fpage>100</fpage>
  <lpage>-115</lpage>
</bibl>

<bibl id="B52">
  <title><p>Online feature selection and its applications</p></title>
  <aug>
    <au><snm>Wang</snm><fnm>J</fnm></au>
    <au><snm>Zhao</snm><fnm>P</fnm></au>
    <au><snm>Hoi</snm><fnm>SC</fnm></au>
    <au><snm>Jin</snm><fnm>R</fnm></au>
  </aug>
  <source>IEEE Transactions on knowledge and data engineering</source>
  <publisher>IEEE</publisher>
  <pubdate>2013</pubdate>
  <volume>26</volume>
  <issue>3</issue>
  <fpage>698</fpage>
  <lpage>-710</lpage>
</bibl>

</refgrp>
} % end of \BMCxmlcomment
% for author-year bibliography (bmc-mathphys or spbasic)
% a) write to bib file (bmc-mathphys only)
% @settings{label, options="nameyear"}

% or include bibliography directly:

\newcommand{\BMCxmlcomment}[1]{}

\BMCxmlcomment{

<refgrp>

<bibl id="B1">
  <title><p>Learning under Concept Drift: A Review</p></title>
  <aug>
    <au><snm>Lu</snm><fnm>J</fnm></au>
    <au><snm>Liu</snm><fnm>A</fnm></au>
    <au><snm>Dong</snm><fnm>F</fnm></au>
    <au><snm>Gu</snm><fnm>F</fnm></au>
    <au><snm>Gama</snm><fnm>J</fnm></au>
    <au><snm>Zhang</snm><fnm>G</fnm></au>
  </aug>
  <source>IEEE Transactions on Knowledge and Data Engineering 31, no. 12
  (2018): 2346-2363</source>
  <pubdate>2018</pubdate>
</bibl>

<bibl id="B2">
  <title><p>Learning in nonstationary environments: A survey</p></title>
  <aug>
    <au><snm>Ditzler</snm><fnm>G</fnm></au>
    <au><snm>Roveri</snm><fnm>M</fnm></au>
    <au><snm>Alippi</snm><fnm>C</fnm></au>
    <au><snm>Polikar</snm><fnm>R</fnm></au>
  </aug>
  <source>IEEE Computational Intelligence Magazine</source>
  <publisher>IEEE</publisher>
  <pubdate>2015</pubdate>
  <volume>10</volume>
  <issue>4</issue>
  <fpage>12</fpage>
  <lpage>-25</lpage>
</bibl>

<bibl id="B3">
  <title><p>A survey on learning from data streams: current and future
  trends</p></title>
  <aug>
    <au><snm>Gama</snm><fnm>J</fnm></au>
  </aug>
  <source>Progress in Artificial Intelligence</source>
  <publisher>Springer</publisher>
  <pubdate>2012</pubdate>
  <volume>1</volume>
  <issue>1</issue>
  <fpage>45</fpage>
  <lpage>-55</lpage>
</bibl>

<bibl id="B4">
  <title><p>Mining high-speed data streams</p></title>
  <aug>
    <au><snm>Domingos</snm><fnm>P</fnm></au>
    <au><snm>Hulten</snm><fnm>G</fnm></au>
  </aug>
  <source>Proceedings of the sixth ACM SIGKDD international conference on
  Knowledge discovery and data mining</source>
  <pubdate>2000</pubdate>
  <fpage>71</fpage>
  <lpage>-80</lpage>
</bibl>

<bibl id="B5">
  <title><p>Stress-testing hoeffding trees</p></title>
  <aug>
    <au><snm>Holmes</snm><fnm>G</fnm></au>
    <au><snm>Kirkby</snm><fnm>R</fnm></au>
    <au><snm>Pfahringer</snm><fnm>B</fnm></au>
  </aug>
  <source>European conference on principles of data mining and knowledge
  discovery</source>
  <pubdate>2005</pubdate>
  <fpage>495</fpage>
  <lpage>-502</lpage>
</bibl>

<bibl id="B6">
  <title><p>Fast perceptron decision tree learning from evolving data
  streams</p></title>
  <aug>
    <au><snm>Bifet</snm><fnm>A</fnm></au>
    <au><snm>Holmes</snm><fnm>G</fnm></au>
    <au><snm>Pfahringer</snm><fnm>B</fnm></au>
    <au><snm>Frank</snm><fnm>E</fnm></au>
  </aug>
  <source>Pacific-Asia conference on knowledge discovery and data
  mining</source>
  <pubdate>2010</pubdate>
  <fpage>299</fpage>
  <lpage>-310</lpage>
</bibl>

<bibl id="B7">
  <title><p>Batch-incremental versus instance-incremental learning in dynamic
  and evolving data</p></title>
  <aug>
    <au><snm>Read</snm><fnm>J</fnm></au>
    <au><snm>Bifet</snm><fnm>A</fnm></au>
    <au><snm>Pfahringer</snm><fnm>B</fnm></au>
    <au><snm>Holmes</snm><fnm>G</fnm></au>
  </aug>
  <source>International symposium on intelligent data analysis</source>
  <pubdate>2012</pubdate>
  <fpage>313</fpage>
  <lpage>-323</lpage>
</bibl>

<bibl id="B8">
  <title><p>Learning regularized hoeffding trees from data streams</p></title>
  <aug>
    <au><snm>Barddal</snm><fnm>JP</fnm></au>
    <au><snm>Enembreck</snm><fnm>F</fnm></au>
  </aug>
  <source>Proceedings of the 34th ACM/SIGAPP Symposium on Applied
  Computing</source>
  <pubdate>2019</pubdate>
  <fpage>574</fpage>
  <lpage>-581</lpage>
</bibl>

<bibl id="B9">
  <title><p>Regularized and incremental decision trees for data
  streams</p></title>
  <aug>
    <au><snm>Barddal</snm><fnm>JP</fnm></au>
    <au><snm>Enembreck</snm><fnm>F</fnm></au>
  </aug>
  <source>Annals of Telecommunications</source>
  <publisher>Springer</publisher>
  <pubdate>2020</pubdate>
  <volume>75</volume>
  <issue>9</issue>
  <fpage>493</fpage>
  <lpage>-503</lpage>
</bibl>

<bibl id="B10">
  <title><p>Issues in Evaluation of Stream Learning Algorithms</p></title>
  <aug>
    <au><snm>Gama</snm><fnm>Ja</fnm></au>
    <au><snm>Sebasti\ {a}o</snm><fnm>R</fnm></au>
    <au><snm>Rodrigues</snm><fnm>PP</fnm></au>
  </aug>
  <publisher>New York, NY, USA: Association for Computing Machinery</publisher>
  <series><title><p>KDD '09</p></title></series>
  <pubdate>2009</pubdate>
  <fpage>329–338</fpage>
  <url>https://doi.org/10.1145/1557019.1557060</url>
</bibl>

<bibl id="B11">
  <title><p>On evaluating stream learning algorithms</p></title>
  <aug>
    <au><snm>Gama</snm><fnm>J</fnm></au>
    <au><snm>Sebasti{\~{a}}o</snm><fnm>R</fnm></au>
    <au><snm>Rodrigues</snm><fnm>PP</fnm></au>
  </aug>
  <source>Machine Learning</source>
  <publisher>Springer Science and Business Media {LLC}</publisher>
  <pubdate>2012</pubdate>
  <volume>90</volume>
  <issue>3</issue>
  <fpage>317</fpage>
  <lpage>-346</lpage>
</bibl>

<bibl id="B12">
  <title><p>{CD}-{MOA}: Change Detection Framework for Massive Online
  Analysis</p></title>
  <aug>
    <au><snm>Bifet</snm><fnm>A</fnm></au>
    <au><snm>Read</snm><fnm>J</fnm></au>
    <au><snm>Pfahringer</snm><fnm>B</fnm></au>
    <au><snm>Holmes</snm><fnm>G</fnm></au>
    <au><snm>{\v{Z}}liobait{\.{e}}</snm><fnm>I</fnm></au>
  </aug>
  <publisher>Springer Berlin Heidelberg</publisher>
  <pubdate>2013</pubdate>
  <fpage>92</fpage>
  <lpage>-103</lpage>
</bibl>

<bibl id="B13">
  <title><p>Ensemble learning for data stream analysis: A survey</p></title>
  <aug>
    <au><snm>Krawczyk</snm><fnm>B</fnm></au>
    <au><snm>Minku</snm><fnm>LL</fnm></au>
    <au><snm>Gama</snm><fnm>J</fnm></au>
    <au><snm>Stefanowski</snm><fnm>J</fnm></au>
    <au><snm>Wo{\'{z}}niak</snm><fnm>M</fnm></au>
  </aug>
  <source>Information Fusion</source>
  <publisher>Elsevier {BV}</publisher>
  <pubdate>2017</pubdate>
  <volume>37</volume>
  <fpage>132</fpage>
  <lpage>-156</lpage>
</bibl>

<bibl id="B14">
  <title><p>Catching up with the Data: Research Issues in Mining Data
  Streams</p></title>
  <aug>
    <au><snm>Domingos</snm><fnm>PM</fnm></au>
    <au><snm>Hulten</snm><fnm>G</fnm></au>
  </aug>
  <source>DMKD</source>
  <pubdate>2001</pubdate>
</bibl>

<bibl id="B15">
  <title><p>Scikit-multiflow: A multi-output streaming framework</p></title>
  <aug>
    <au><snm>Montiel</snm><fnm>J</fnm></au>
    <au><snm>Read</snm><fnm>J</fnm></au>
    <au><snm>Bifet</snm><fnm>A</fnm></au>
    <au><snm>Abdessalem</snm><fnm>T</fnm></au>
  </aug>
  <source>The Journal of Machine Learning Research</source>
  <publisher>JMLR. org</publisher>
  <pubdate>2018</pubdate>
  <volume>19</volume>
  <issue>1</issue>
  <fpage>2915</fpage>
  <lpage>-2914</lpage>
</bibl>

<bibl id="B16">
  <title><p>River: machine learning for streaming data in Python</p></title>
  <aug>
    <au><snm>Montiel</snm><fnm>J</fnm></au>
    <au><snm>Halford</snm><fnm>M</fnm></au>
    <au><snm>Mastelini</snm><fnm>SM</fnm></au>
    <au><snm>Bolmier</snm><fnm>G</fnm></au>
    <au><snm>Sourty</snm><fnm>R</fnm></au>
    <au><snm>Vaysse</snm><fnm>R</fnm></au>
    <au><snm>Zouitine</snm><fnm>A</fnm></au>
    <au><snm>Gomes</snm><fnm>HM</fnm></au>
    <au><snm>Read</snm><fnm>J</fnm></au>
    <au><snm>Abdessalem</snm><fnm>T</fnm></au>
    <au><cnm>others</cnm></au>
  </aug>
  <pubdate>2021</pubdate>
</bibl>

<bibl id="B17">
  <title><p>Characterizing concept drift</p></title>
  <aug>
    <au><snm>Webb</snm><fnm>GI</fnm></au>
    <au><snm>Hyde</snm><fnm>R</fnm></au>
    <au><snm>Cao</snm><fnm>H</fnm></au>
    <au><snm>Nguyen</snm><fnm>HL</fnm></au>
    <au><snm>Petitjean</snm><fnm>F</fnm></au>
  </aug>
  <source>Data Mining and Knowledge Discovery</source>
  <publisher>Springer</publisher>
  <pubdate>2016</pubdate>
  <volume>30</volume>
  <issue>4</issue>
  <fpage>964</fpage>
  <lpage>-994</lpage>
</bibl>

<bibl id="B18">
  <title><p>Efficient Online Evaluation of Big Data Stream
  Classifiers</p></title>
  <aug>
    <au><snm>Bifet</snm><fnm>A</fnm></au>
    <au><snm>Francisci Morales</snm><fnm>G</fnm></au>
    <au><snm>Read</snm><fnm>J</fnm></au>
    <au><snm>Holmes</snm><fnm>G</fnm></au>
    <au><snm>Pfahringer</snm><fnm>B</fnm></au>
  </aug>
  <source>Proceedings of the 21th {ACM} {SIGKDD} International Conference on
  Knowledge Discovery and Data Mining</source>
  <publisher>New York, NY, USA: {ACM}</publisher>
  <pubdate>2015</pubdate>
</bibl>

<bibl id="B19">
  <title><p>A comparative study on concept drift detectors</p></title>
  <aug>
    <au><snm>Gon{\c{c}}alves</snm><fnm>PM</fnm></au>
    <au><snm>Carvalho Santos</snm><fnm>SG</fnm></au>
    <au><snm>Barros</snm><fnm>RS</fnm></au>
    <au><snm>Vieira</snm><fnm>DC</fnm></au>
  </aug>
  <source>Expert Systems with Applications</source>
  <publisher>Elsevier {BV}</publisher>
  <pubdate>2014</pubdate>
  <volume>41</volume>
  <issue>18</issue>
  <fpage>8144</fpage>
  <lpage>-8156</lpage>
</bibl>

<bibl id="B20">
  <title><p>A survey on data preprocessing for data stream mining: Current
  status and future directions</p></title>
  <aug>
    <au><snm>Ram{\'{\i}}rez Gallego</snm><fnm>S</fnm></au>
    <au><snm>Krawczyk</snm><fnm>B</fnm></au>
    <au><snm>Garc{\'{\i}}a</snm><fnm>S</fnm></au>
    <au><snm>Wo{\'{z}}niak</snm><fnm>M</fnm></au>
    <au><snm>Herrera</snm><fnm>F</fnm></au>
  </aug>
  <source>Neurocomputing</source>
  <publisher>Elsevier {BV}</publisher>
  <pubdate>2017</pubdate>
  <volume>239</volume>
  <fpage>39</fpage>
  <lpage>-57</lpage>
</bibl>

<bibl id="B21">
  <title><p>A Survey on Ensemble Learning for Data Stream
  Classification</p></title>
  <aug>
    <au><snm>Gomes</snm><fnm>HM</fnm></au>
    <au><snm>Barddal</snm><fnm>JP</fnm></au>
    <au><snm>Enembreck</snm><fnm>F</fnm></au>
    <au><snm>Bifet</snm><fnm>A</fnm></au>
  </aug>
  <source>{ACM} Computing Surveys</source>
  <publisher>Association for Computing Machinery ({ACM})</publisher>
  <pubdate>2017</pubdate>
  <volume>50</volume>
  <issue>2</issue>
  <fpage>1</fpage>
  <lpage>-36</lpage>
</bibl>

<bibl id="B22">
  <title><p>Big data stream analysis: a systematic literature
  review</p></title>
  <aug>
    <au><snm>Kolajo</snm><fnm>T</fnm></au>
    <au><snm>Daramola</snm><fnm>O</fnm></au>
    <au><snm>Adebiyi</snm><fnm>A</fnm></au>
  </aug>
  <source>Journal of Big Data</source>
  <publisher>Springer Science and Business Media {LLC}</publisher>
  <pubdate>2019</pubdate>
  <volume>6</volume>
  <issue>1</issue>
</bibl>

<bibl id="B23">
  <title><p>Dynamic Model Tree for Interpretable Data Stream
  Learning</p></title>
  <aug>
    <au><snm>Haug</snm><fnm>J</fnm></au>
    <au><snm>Broelemann</snm><fnm>K</fnm></au>
    <au><snm>Kasneci</snm><fnm>G</fnm></au>
  </aug>
  <source>arXiv preprint arXiv:2203.16181</source>
  <pubdate>2022</pubdate>
</bibl>

<bibl id="B24">
  <title><p>On the stability of feature selection algorithms.</p></title>
  <aug>
    <au><snm>Nogueira</snm><fnm>S</fnm></au>
    <au><snm>Sechidis</snm><fnm>K</fnm></au>
    <au><snm>Brown</snm><fnm>G</fnm></au>
  </aug>
  <source>J. Mach. Learn. Res.</source>
  <pubdate>2017</pubdate>
  <volume>18</volume>
  <issue>1</issue>
  <fpage>6345</fpage>
  <lpage>-6398</lpage>
</bibl>

<bibl id="B25">
  <title><p>Leveraging Model Inherent Variable Importance for Stable Online
  Feature Selection</p></title>
  <aug>
    <au><snm>Haug</snm><fnm>J</fnm></au>
    <au><snm>Pawelczyk</snm><fnm>M</fnm></au>
    <au><snm>Broelemann</snm><fnm>K</fnm></au>
    <au><snm>Kasneci</snm><fnm>G</fnm></au>
  </aug>
  <source>Proceedings of the 26th {ACM} {SIGKDD} International Conference on
  Knowledge Discovery {\&} Data Mining</source>
  <publisher>New York, NY, USA: {ACM}</publisher>
  <pubdate>2020</pubdate>
</bibl>

<bibl id="B26">
  <title><p>Prequential {AUC}: properties of the area under the {ROC} curve for
  data streams with concept drift</p></title>
  <aug>
    <au><snm>Brzezinski</snm><fnm>D</fnm></au>
    <au><snm>Stefanowski</snm><fnm>J</fnm></au>
  </aug>
  <source>Knowledge and Information Systems</source>
  <publisher>Springer Science and Business Media {LLC}</publisher>
  <pubdate>2017</pubdate>
  <volume>52</volume>
  <issue>2</issue>
  <fpage>531</fpage>
  <lpage>-562</lpage>
</bibl>

<bibl id="B27">
  <title><p>Learning Parameter Distributions to Detect Concept Drift in Data
  Streams</p></title>
  <aug>
    <au><snm>Haug</snm><fnm>J</fnm></au>
    <au><snm>Kasneci</snm><fnm>G</fnm></au>
  </aug>
  <source>2020 25th International Conference on Pattern Recognition
  ({ICPR})</source>
  <publisher>USA: {IEEE}</publisher>
  <pubdate>2021</pubdate>
</bibl>

<bibl id="B28">
  <title><p>Explanation in artificial intelligence: Insights from the social
  sciences</p></title>
  <aug>
    <au><snm>Miller</snm><fnm>T</fnm></au>
  </aug>
  <source>Artificial Intelligence</source>
  <publisher>Elsevier {BV}</publisher>
  <pubdate>2019</pubdate>
  <volume>267</volume>
  <fpage>1</fpage>
  <lpage>-38</lpage>
</bibl>

<bibl id="B29">
  <title><p>A Survey of Methods for Explaining Black Box Models</p></title>
  <aug>
    <au><snm>Guidotti</snm><fnm>R</fnm></au>
    <au><snm>Monreale</snm><fnm>A</fnm></au>
    <au><snm>Ruggieri</snm><fnm>S</fnm></au>
    <au><snm>Turini</snm><fnm>F</fnm></au>
    <au><snm>Giannotti</snm><fnm>F</fnm></au>
    <au><snm>Pedreschi</snm><fnm>D</fnm></au>
  </aug>
  <source>{ACM} Computing Surveys</source>
  <publisher>Association for Computing Machinery ({ACM})</publisher>
  <pubdate>2019</pubdate>
  <volume>51</volume>
  <issue>5</issue>
  <fpage>1</fpage>
  <lpage>-42</lpage>
</bibl>

<bibl id="B30">
  <title><p>Bias in data-driven artificial intelligence systems—An
  introductory survey</p></title>
  <aug>
    <au><snm>Ntoutsi</snm><fnm>E</fnm></au>
    <au><snm>Fafalios</snm><fnm>P</fnm></au>
    <au><snm>Gadiraju</snm><fnm>U</fnm></au>
    <au><snm>Iosifidis</snm><fnm>V</fnm></au>
    <au><snm>Nejdl</snm><fnm>W</fnm></au>
    <au><snm>Vidal</snm><fnm>ME</fnm></au>
    <au><snm>Ruggieri</snm><fnm>S</fnm></au>
    <au><snm>Turini</snm><fnm>F</fnm></au>
    <au><snm>Papadopoulos</snm><fnm>S</fnm></au>
    <au><snm>Krasanakis</snm><fnm>E</fnm></au>
    <au><cnm>others</cnm></au>
  </aug>
  <source>Wiley Interdisciplinary Reviews: Data Mining and Knowledge
  Discovery</source>
  <publisher>Wiley Online Library</publisher>
  <pubdate>2020</pubdate>
  <volume>10</volume>
  <issue>3</issue>
  <fpage>e1356</fpage>
</bibl>

<bibl id="B31">
  <title><p>CARLA: A Python Library to Benchmark Algorithmic Recourse and
  Counterfactual Explanation Algorithms</p></title>
  <aug>
    <au><snm>Pawelczyk</snm><fnm>M</fnm></au>
    <au><snm>Bielawski</snm><fnm>S</fnm></au>
    <au><snm>Heuvel</snm><fnm>J</fnm></au>
    <au><snm>Richter</snm><fnm>T</fnm></au>
    <au><snm>Kasneci</snm><fnm>G</fnm></au>
  </aug>
  <pubdate>2021</pubdate>
</bibl>

<bibl id="B32">
  <title><p>Machine Learning Interpretability: A Survey on Methods and
  Metrics</p></title>
  <aug>
    <au><snm>Carvalho</snm><fnm>DV</fnm></au>
    <au><snm>Pereira</snm><fnm>EM</fnm></au>
    <au><snm>Cardoso</snm><fnm>JS</fnm></au>
  </aug>
  <source>Electronics</source>
  <publisher>{MDPI} {AG}</publisher>
  <pubdate>2019</pubdate>
  <volume>8</volume>
  <issue>8</issue>
  <fpage>832</fpage>
</bibl>

<bibl id="B33">
  <title><p>Interpretability of Machine Learning Models and Representations: an
  Introduction</p></title>
  <aug>
    <au><snm>Bibal</snm><fnm>A</fnm></au>
    <au><snm>Fr{\'e}nay</snm><fnm>B</fnm></au>
  </aug>
  <source>ESANN</source>
  <pubdate>2016</pubdate>
</bibl>

<bibl id="B34">
  <title><p>Connecting Interpretability and Robustness in Decision Trees
  Through Separation</p></title>
  <aug>
    <au><snm>Moshkovitz</snm><fnm>M</fnm></au>
    <au><snm>Yang</snm><fnm>YY</fnm></au>
    <au><snm>Chaudhuri</snm><fnm>K</fnm></au>
  </aug>
  <source>International Conference on Machine Learning</source>
  <pubdate>2021</pubdate>
  <fpage>7839</fpage>
  <lpage>-7849</lpage>
</bibl>

<bibl id="B35">
  <title><p>Techniques for Interpretable Machine Learning</p></title>
  <aug>
    <au><snm>Du</snm><fnm>M</fnm></au>
    <au><snm>Liu</snm><fnm>N</fnm></au>
    <au><snm>Hu</snm><fnm>X</fnm></au>
  </aug>
  <source>Communications of the {ACM}</source>
  <publisher>Association for Computing Machinery ({ACM})</publisher>
  <pubdate>2019</pubdate>
  <volume>63</volume>
  <issue>1</issue>
  <fpage>68</fpage>
  <lpage>-77</lpage>
</bibl>

<bibl id="B36">
  <title><p>"Why should i trust you?" Explaining the predictions of any
  classifier</p></title>
  <aug>
    <au><snm>Ribeiro</snm><fnm>MT</fnm></au>
    <au><snm>Singh</snm><fnm>S</fnm></au>
    <au><snm>Guestrin</snm><fnm>C</fnm></au>
  </aug>
  <source>Proceedings of the 22nd ACM SIGKDD international conference on
  knowledge discovery and data mining</source>
  <pubdate>2016</pubdate>
  <fpage>1135</fpage>
  <lpage>-1144</lpage>
</bibl>

<bibl id="B37">
  <title><p>A unified approach to interpreting model predictions</p></title>
  <aug>
    <au><snm>Lundberg</snm><fnm>SM</fnm></au>
    <au><snm>Lee</snm><fnm>SI</fnm></au>
  </aug>
  <source>Proceedings of the 31st international conference on neural
  information processing systems</source>
  <pubdate>2017</pubdate>
  <fpage>4768</fpage>
  <lpage>-4777</lpage>
</bibl>

<bibl id="B38">
  <title><p>Licon: A linear weighting scheme for the contribution ofinput
  variables in deep artificial neural networks</p></title>
  <aug>
    <au><snm>Kasneci</snm><fnm>G</fnm></au>
    <au><snm>Gottron</snm><fnm>T</fnm></au>
  </aug>
  <source>Proceedings of the 25th ACM International on Conference on
  Information and Knowledge Management</source>
  <pubdate>2016</pubdate>
  <fpage>45</fpage>
  <lpage>-54</lpage>
</bibl>

<bibl id="B39">
  <title><p>A Benchmark for Interpretability Methods in Deep Neural
  Networks</p></title>
  <aug>
    <au><snm>Hooker</snm><fnm>S</fnm></au>
    <au><snm>Erhan</snm><fnm>D</fnm></au>
    <au><snm>Kindermans</snm><fnm>PJ</fnm></au>
    <au><snm>Kim</snm><fnm>B</fnm></au>
  </aug>
  <source>Advances in Neural Information Processing Systems</source>
  <publisher>Red Hook, NY, USA: Curran Associates, Inc.</publisher>
  <editor>H. Wallach and H. Larochelle and A. Beygelzimer and F.
  d\textquotesingle Alch\'{e}-Buc and E. Fox and R. Garnett</editor>
  <pubdate>2019</pubdate>
  <volume>32</volume>
</bibl>

<bibl id="B40">
  <title><p>On Baselines for Local Feature Attributions</p></title>
  <aug>
    <au><snm>Haug</snm><fnm>J</fnm></au>
    <au><snm>Zürn</snm><fnm>S</fnm></au>
    <au><snm>El Jiz</snm><fnm>P</fnm></au>
    <au><snm>Kasneci</snm><fnm>G</fnm></au>
  </aug>
  <source>AAAI 2021 Workshop on Explainable Agency in Artificial
  Intelligence</source>
  <pubdate>2021</pubdate>
</bibl>

<bibl id="B41">
  <title><p>stabm: Stability Measures for Feature Selection</p></title>
  <aug>
    <au><snm>Bommert</snm><fnm>A</fnm></au>
    <au><snm>Lang</snm><fnm>M</fnm></au>
  </aug>
  <source>Journal of Open Source Software</source>
  <pubdate>2021</pubdate>
  <volume>6</volume>
  <issue>59</issue>
  <fpage>3010</fpage>
</bibl>

<bibl id="B42">
  <title><p>Stability of Feature Selection Algorithms</p></title>
  <aug>
    <au><snm>Kalousis</snm><fnm>A.</fnm></au>
    <au><snm>Prados</snm><fnm>J.</fnm></au>
    <au><snm>Hilario</snm><fnm>M.</fnm></au>
  </aug>
  <source>Fifth {IEEE} International Conference on Data Mining
  ({ICDM}{\textquotesingle}05)</source>
  <publisher>USA: {IEEE}</publisher>
  <pubdate>2005</pubdate>
</bibl>

<bibl id="B43">
  <title><p>Classification and Novel Class Detection of Data Streams in a
  Dynamic Feature Space</p></title>
  <aug>
    <au><snm>Masud</snm><fnm>MM</fnm></au>
    <au><snm>Chen</snm><fnm>Q</fnm></au>
    <au><snm>Gao</snm><fnm>J</fnm></au>
    <au><snm>Khan</snm><fnm>L</fnm></au>
    <au><snm>Han</snm><fnm>J</fnm></au>
    <au><snm>Thuraisingham</snm><fnm>B</fnm></au>
  </aug>
  <source>Machine Learning and Knowledge Discovery in Databases</source>
  <publisher>Springer Berlin Heidelberg</publisher>
  <pubdate>2010</pubdate>
  <fpage>337</fpage>
  <lpage>-352</lpage>
</bibl>

<bibl id="B44">
  <title><p>On the reliable detection of concept drift from streaming unlabeled
  data</p></title>
  <aug>
    <au><snm>Sethi</snm><fnm>TS</fnm></au>
    <au><snm>Kantardzic</snm><fnm>M</fnm></au>
  </aug>
  <source>Expert Systems with Applications</source>
  <publisher>Elsevier</publisher>
  <pubdate>2017</pubdate>
  <volume>82</volume>
  <fpage>77</fpage>
  <lpage>-99</lpage>
</bibl>

<bibl id="B45">
  <title><p>An Eager Splitting Strategy for Online Decision Trees</p></title>
  <aug>
    <au><snm>Manapragada</snm><fnm>C</fnm></au>
    <au><snm>Gomes</snm><fnm>HM</fnm></au>
    <au><snm>Salehi</snm><fnm>M</fnm></au>
    <au><snm>Bifet</snm><fnm>A</fnm></au>
    <au><snm>Webb</snm><fnm>GI</fnm></au>
  </aug>
  <source>arXiv preprint arXiv:2010.10935</source>
  <pubdate>2020</pubdate>
</bibl>

<bibl id="B46">
  <title><p>Challenges in benchmarking stream learning algorithms with
  real-world data</p></title>
  <aug>
    <au><snm>Souza</snm><fnm>VM</fnm></au>
    <au><snm>Reis</snm><fnm>DM</fnm></au>
    <au><snm>Maletzke</snm><fnm>AG</fnm></au>
    <au><snm>Batista</snm><fnm>GE</fnm></au>
  </aug>
  <source>Data Mining and Knowledge Discovery</source>
  <publisher>Springer</publisher>
  <pubdate>2020</pubdate>
  <volume>34</volume>
  <issue>6</issue>
  <fpage>1805</fpage>
  <lpage>-1858</lpage>
</bibl>

<bibl id="B47">
  <title><p>T{\"u}EyeQ, a rich IQ test performance data set with eye movement,
  educational and socio-demographic information</p></title>
  <aug>
    <au><snm>Kasneci</snm><fnm>E</fnm></au>
    <au><snm>Kasneci</snm><fnm>G</fnm></au>
    <au><snm>Appel</snm><fnm>T</fnm></au>
    <au><snm>Haug</snm><fnm>J</fnm></au>
    <au><snm>Wortha</snm><fnm>F</fnm></au>
    <au><snm>Tibus</snm><fnm>M</fnm></au>
    <au><snm>Trautwein</snm><fnm>U</fnm></au>
    <au><snm>Gerjets</snm><fnm>P</fnm></au>
  </aug>
  <source>Scientific Data</source>
  <publisher>Nature Publishing Group</publisher>
  <pubdate>2021</pubdate>
  <volume>8</volume>
  <issue>1</issue>
  <fpage>1</fpage>
  <lpage>-14</lpage>
</bibl>

<bibl id="B48">
  <title><p>{TüEyeQ, a rich IQ test performance data set with eye movement,
  educational and socio-demographic information}</p></title>
  <aug>
    <au><snm>Kasneci</snm><fnm>E</fnm></au>
    <au><snm>Kasneci</snm><fnm>G</fnm></au>
    <au><snm>Appel</snm><fnm>T</fnm></au>
    <au><snm>Haug</snm><fnm>J</fnm></au>
    <au><snm>Wortha</snm><fnm>F</fnm></au>
    <au><snm>Tibus</snm><fnm>M</fnm></au>
    <au><snm>Trautwein</snm><fnm>U</fnm></au>
    <au><snm>Gerjets</snm><fnm>P</fnm></au>
  </aug>
  <publisher>USA: Harvard Dataverse</publisher>
  <pubdate>2020</pubdate>
</bibl>

<bibl id="B49">
  <title><p>Adaptive learning from evolving data streams</p></title>
  <aug>
    <au><snm>Bifet</snm><fnm>A</fnm></au>
    <au><snm>Gavalda</snm><fnm>R</fnm></au>
  </aug>
  <source>International Symposium on Intelligent Data Analysis</source>
  <pubdate>2009</pubdate>
  <fpage>249</fpage>
  <lpage>-260</lpage>
</bibl>

<bibl id="B50">
  <title><p>Learning from time-changing data with adaptive
  windowing</p></title>
  <aug>
    <au><snm>Bifet</snm><fnm>A</fnm></au>
    <au><snm>Gavalda</snm><fnm>R</fnm></au>
  </aug>
  <source>Proceedings of the 2007 SIAM international conference on data
  mining</source>
  <pubdate>2007</pubdate>
  <fpage>443</fpage>
  <lpage>-448</lpage>
</bibl>

<bibl id="B51">
  <title><p>Continuous Inspection Schemes</p></title>
  <aug>
    <au><snm>Page</snm><fnm>E. S.</fnm></au>
  </aug>
  <source>Biometrika</source>
  <publisher>[Oxford University Press, Biometrika Trust]</publisher>
  <pubdate>1954</pubdate>
  <volume>41</volume>
  <issue>1/2</issue>
  <fpage>100</fpage>
  <lpage>-115</lpage>
</bibl>

<bibl id="B52">
  <title><p>Online feature selection and its applications</p></title>
  <aug>
    <au><snm>Wang</snm><fnm>J</fnm></au>
    <au><snm>Zhao</snm><fnm>P</fnm></au>
    <au><snm>Hoi</snm><fnm>SC</fnm></au>
    <au><snm>Jin</snm><fnm>R</fnm></au>
  </aug>
  <source>IEEE Transactions on knowledge and data engineering</source>
  <publisher>IEEE</publisher>
  <pubdate>2013</pubdate>
  <volume>26</volume>
  <issue>3</issue>
  <fpage>698</fpage>
  <lpage>-710</lpage>
</bibl>

</refgrp>
} % end of \BMCxmlcomment

\end{backmatter}
\end{document}